\documentclass{article} 

\usepackage[utf8]{inputenc} 
\usepackage[T1]{fontenc}    
\usepackage{hyperref}       
\usepackage{url}            
\usepackage{booktabs}       
\usepackage{amsmath,amsfonts,amsthm}       
\usepackage{mathtools}      
\usepackage{nicefrac}       
\usepackage{microtype}      
\usepackage{subcaption}
\usepackage{bbm}
\usepackage{bm}
\usepackage{multirow, tabularx}
\usepackage[inline]{enumitem}
\usepackage{diagbox}
\usepackage{pifont}
\usepackage[capitalise]{cleveref}  
\usepackage{hyperref}
\usepackage{comment}
\usepackage{etoolbox}
\usepackage{xcolor}
\usepackage{graphicx}
\usepackage{tikz}
\usepackage[ruled,vlined,linesnumbered]{algorithm2e}
\usepackage{algorithmic}
\usepackage[font=small]{caption}
\usepackage{wrapfig}
\usepackage{makecell}

\usepackage[numbers,sort]{natbib}
\newlength{\defbaselineskip}
\definecolor{redBrown}{RGB}{241, 90, 36} 

\DeclarePairedDelimiter\floor{\lfloor}{\rfloor}

\title{Noisy Batch Active Learning with Deterministic Annealing}

\usepackage{authblk}
\author[1]{Gaurav Gupta}
\author[2]{Anit Kumar Sahu}
\author[2]{Wan-Yi Lin}
\affil[1]{Ming Hsieh Department of Electrical and Computer Engineering,\\ University of Southern California}
\affil[2]{Bosch Center for Artificial Intelligence, Pittsburgh, PA}
\affil[ ]{{\texttt{ggaurav@usc.edu}, \texttt{anit.sahu@gmail.com}, \texttt{wan-yi.lin@us.bosch.com}}}
\date{}

\begin{document}

\maketitle

\begin{abstract}
We study the problem of training machine learning models incrementally with batches of samples annotated with noisy oracles. We select each batch of samples that are important and also diverse via clustering and importance sampling. More importantly, we incorporate model uncertainty into the sampling probability to compensate poor estimation of the importance scores when the training data is too small to build a meaningful model. Experiments on benchmark image classification datasets (MNIST, SVHN, CIFAR10, and EMNIST) show improvement over existing active learning strategies. We introduce an extra denoising layer to deep networks to make active learning robust to label noises and show significant improvements.
\end{abstract}

\section{Introduction}
\label{sec:intro}
Supervised learning is the most widely used machine learning method, but it requires labelled data for training. It is time-consuming and labor-intensive to annotate a large dataset for complex supervised machine learning models. For example, ImageNet \cite{Imagenet_15} reported the time taken to annotate one object to be roughly 55 seconds. Hence an active learning approach which selects the most relevant samples for annotation to incrementally train machine learning models is a very attractive avenue, especially for training deep networks for newer problems that have little annotated data.

Classical active learning appends the training dataset with a single sample-label pair at a time. Given the increasing complexity of machine learning models, it is natural to expand active learning procedures to append a batch of samples at each iteration instead of just one. Keeping such training overhead in mind, a few batch active learning procedures have been developed in the literature \citep{pmlr-v37-wei15,sener2018active,Samarth_ICCV_19}.

\begin{figure}[t]
	\centering
	\begin{subfigure}[t]{0.33\textwidth}
	\centering
	\includegraphics*[height = 1.5in]{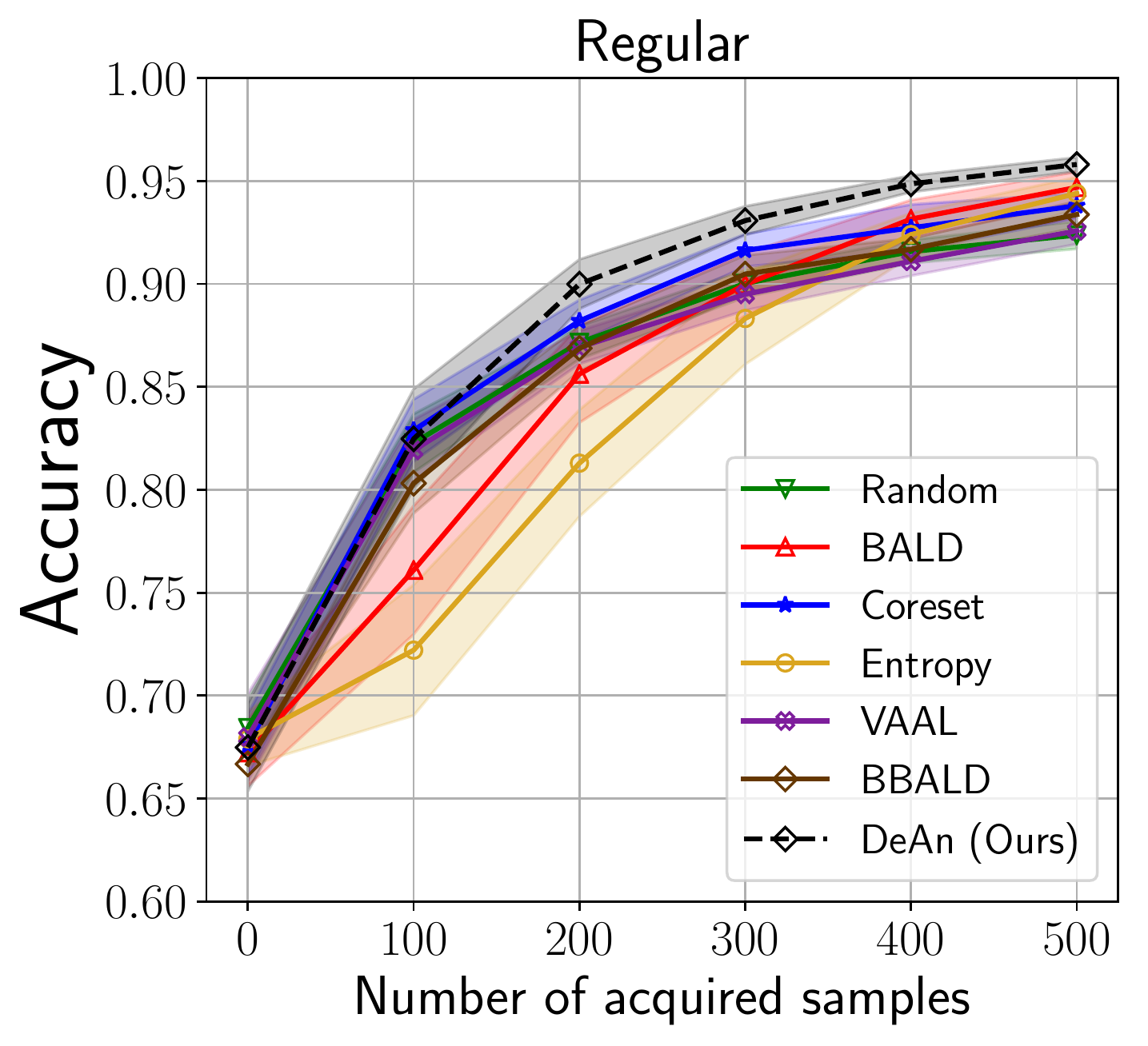}
	\caption{}
	\label{sfig:MNIST_intro_regular}
	\end{subfigure}
	\hspace*{-10pt}
	\begin{subfigure}[t]{0.33\textwidth}
	\centering
	\includegraphics*[height = 1.5in]{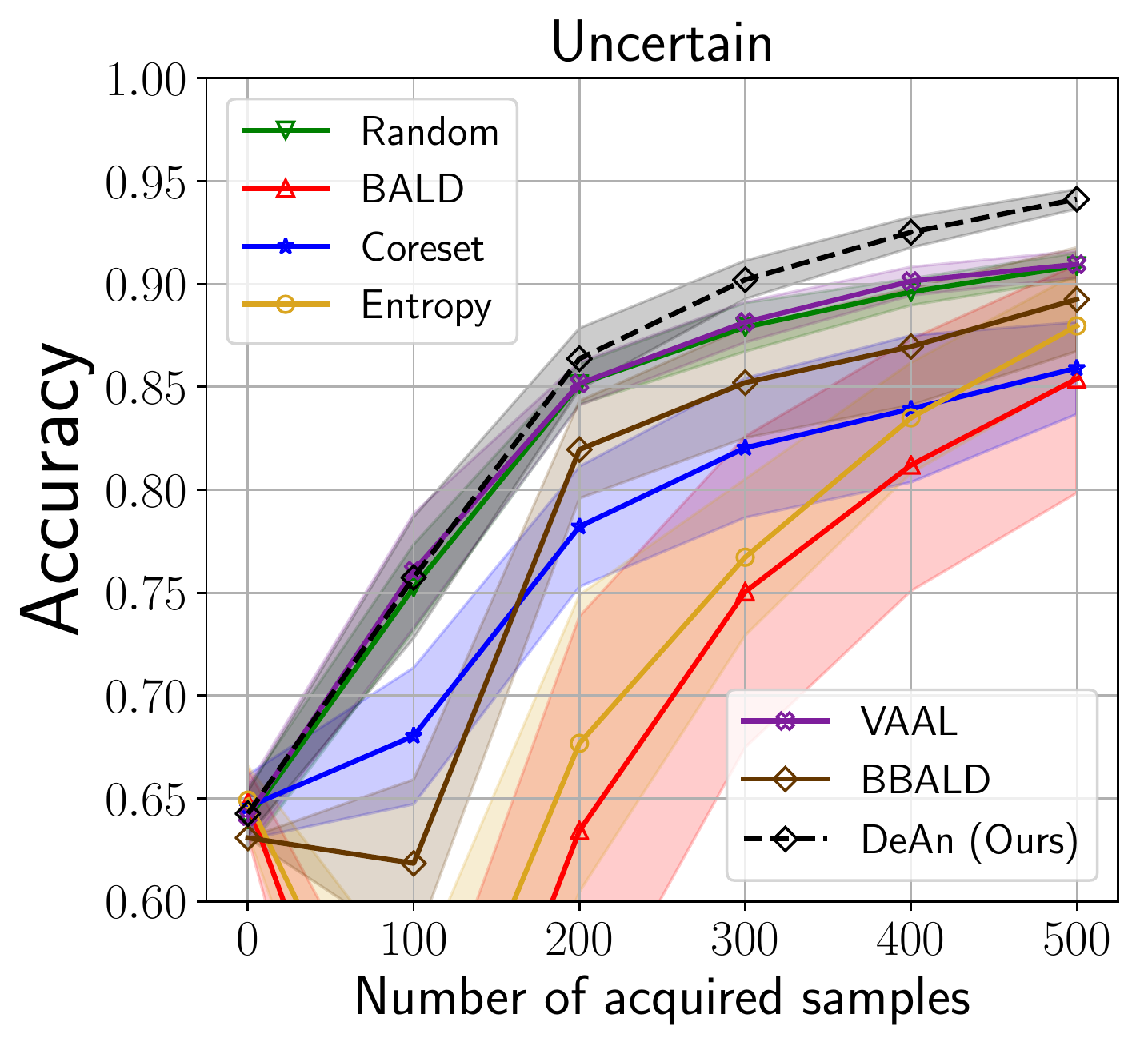}
	\caption{}
	\label{sfig:MNIST_intro_uncertain}
	\end{subfigure}
	\hspace*{-10pt}
	\begin{subfigure}[t]{0.33\textwidth}
	\centering
	\includegraphics*[height = 1.5in]{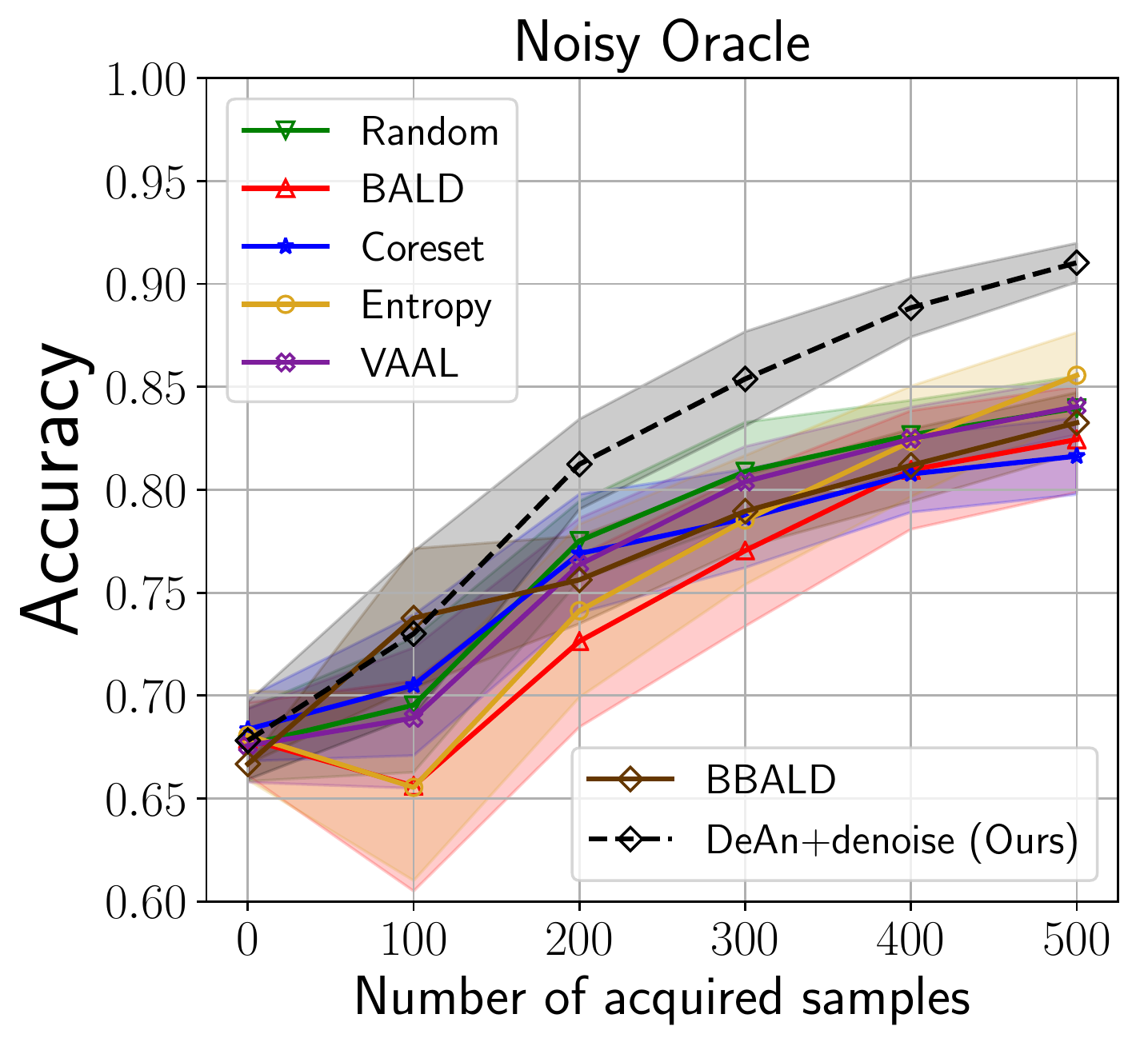}
	\caption{}
	\label{sfig:MNIST_intro_noisy}
	\end{subfigure}
	\caption{Prior active learning methods in MNIST suffers performance degradation with increase in model uncertainty in (\ref{sfig:MNIST_intro_uncertain}), or with oracle noise in (\ref{sfig:MNIST_intro_noisy}). For setup details see Section\,\ref{ssec:ablation}.}
	\vspace*{-7pt}
	\label{fig:MNIST_noise_intro}
\end{figure}

When initializing the model with a very small seed dataset, active learning suffers from the cold-start problem: at the very beginning of active learning procedures, the model is far from being accurate and hence the inferred output of the model is incorrect/uncertain. Since active learning relies on output of the current model to select next samples, a poor initial model provides uncertain estimation of selection criteria which then leads to selection of wrong samples.  Prior art on batch active learning suffers performance degradation due to these issues as shown in Figure\,\ref{sfig:MNIST_intro_uncertain}.

Most active learning procedures assume the oracle to be perfect, i.e., it can always annotate samples correctly. However, in real-world scenarios and given the increasing usage of crowd sourcing, for example Amazon Mechanical Turk (AMT), for labelling data, most oracles are noisy. The noise induced by the oracle in many scenarios is \textit{resolute}. Having multiple annotations on the same sample cannot guarantee noise-free labels due to the presence of systematic bias in the setup and leads to consistent mistakes.  To validate this point, we ran a crowd annotation experiment on ESC50 dataset \citep{ESC50}: each sample is annotated by 5 crowdworkers on AMT and the majority vote of the 5 annotations is considered the label. It turned out for some classes, $10\%$ of the samples are annotated wrong, even with 5 annotators. Details of the experiment is provided in the supplementary materials. Under such noisy oracle scenarios, classical active learning algorithms such as \citep{Chen15b} under-perform as shown in \figurename\,\ref{sfig:MNIST_intro_noisy}. Motivating from these observations, we fashion a batch active learning strategy to be robust to noisy oracles.
The main contributions of this work are as follows: (1) we propose a batch sample selection method based on importance sampling and clustering which caters to drawing a batch which is simultaneously \textbf{diverse} and \textbf{important} to the model; (2) we incorporate model uncertainty into the sampling probability through deterministic annealing to compensate poor estimation of the importance scores when the training data is too small to build a meaningful model; (3) we introduce a denoising layer to deep networks to robustify active learning to noisy oracles. Main results, as shown in Figure\,\ref{fig:mainResults} demonstrate that in noise-free scenario, our method performs as the best over the whole active learning procedure, and in noisy scenario, our method outperforms significantly over state-of-the-art methods.
\section{Related work}
\label{sec:relWork}
\paragraph{Active Learning}
Active learning \citep{Simon2001PhD} is a well-studied problem and has gain interest in deep learning as well. A survey summarizes various existing approaches in \citep{settles.tr09}. In a nutshell, two key and diverse ways to tackle this problem in the literature are \textit{discrimination} and \textit{representation}. The representation line of work focuses on selecting samples that can represent the whole unlabelled training set while the discrimination line of work aims at selecting `tough' examples from the pool set, for example, using information theoretic scores in \citep{Kay1992}, entropy as uncertainty in \citep{Wang2014AL}. Along the lines of ensemble methods we have works, for example, \citep{Beluch_CVPR_18,lakshminarayanan2016simple}.

The work of discrimination-based active learning \citep{Houlsby11bayesianactive} uses mutual information, Bayesian Active Learning by Disagreement (BALD), as discriminating criteria. In \citep{Gal:2017} the authors used dropout approximation to compute the BALD scores for modern Convolutional Neural Networks (CNNs). However, these approaches do not consider batch acquisition and hence lack of diversity in selected batch samples causing performance lag.

\paragraph{Batch Active Learning}
Active learning in the batch acquisition manner has been studied from the perspective of set selection and using submodularity or its variants in a variety of works. The authors in \citep{pmlr-v37-wei15} utilize submodularity for naive Bayes and nearest neighbor. \citep{Chen:2013} considers pool-based Bayesian active learning with a finite set of candidate hypotheses. A pool-based active learning is also discussed in \citep{Ganti2011UPALUP} which considered risk minimization under given hypothesis space. The authors in \citep{sener2018active} use coreset approach to select representative points of the pool set by utilizing network penultimate layer output. The work in \citep{Ash2020Deep} also use penultimate layer output to make batch selection. An adversarial learning of variational auto-encoders is used for batch active learning in \citep{Samarth_ICCV_19}. Recently, a batch version of BALD is proposed in \citep{kirsch2019batchbald}.
\paragraph{Model Uncertainty}
The uncertainty for deep learning models, especially CNNs, was first addressed in \citep{gal16,Gal2016Thesis} using dropout as Bayesian approximation. Model uncertainty approximation using Batch Normalization (BN) has been shown in \citep{Teye2018BayesianUE}. Both of these approaches in some sense exploit the stochastic layers (Dropout, BN) to extract model uncertainty. The importance of model uncertainty is also emphasized in the work of \citep{Kendall:2017}. The work witnesses model as well as label uncertainty which they termed as epistemic and aleatoric uncertainty, respectively. We also address both of these uncertainties in this work.
\paragraph{Noisy Oracle}
The importance of noisy labels from oracle has been realized in the works like \citep{Chen:2015:SSV:2888116.2888204,Chen:2013} which utilized the concept of adaptive submodularity for providing theoretical guarantees. Same problem but with correlated noisy tests is studied in \citep{chen2017near}. Active learning with noisy oracles is also studied in \citep{NaghshvarNoisy} but without deep learning setup. The authors in \citep{khetan2018learning} used a variation of Expectation Maximization algorithm to estimate the correct labels as well as annotating workers quality. 

The closest work to us in the noisy oracle setting for deep learning models are \citep{Jindal2019ANN}. The authors also propose to augment the model with an extra full-connected dense layer. However, the denoising layer does not follow any probability simplex constraint, and they use modified loss function for the noise accountability along with dropout regularization.
\vspace*{-5pt}
\section{Problem Formulation}
\label{sec:probForm}
\vspace*{-5pt}
\paragraph{Notations}
The $i$th ($j$th) row (column) of a matrix ${\bf X}$ is denoted as ${\bf X}_{i,.} ({\bf X}_{.,j})$. $\Delta^{K-1}$ is the probability simplex of dimension $K$, where $\Delta^{K-1} = \{(p_{1},p_{2},\hdots,p_{K})\in\mathbb{R}^{K}\vert \sum\nolimits_{i=1}^{K}p_{i} = 1\wedge \, p_{i}\geq 0\,\,\forall i \}$. The Shannon entropy is defined as: $\mathbb{H}({\bf p}) = -\sum\nolimits_{i=1}^{K}p_{i}\log(p_{i})$, and the Kullback-Leibler (KL) divergence between ${\bf p, q}$ is defined as $KL({\bf p}\vert\vert {\bf q}) = \sum\nolimits_{i=1}^{K}p_{i}\log(p_{i}/q_{i})$. The KL-divergence is always non-negative and is $0$ if and only if ${\bf p}= {\bf q}$. The expectation operator is taken as $\mathbb{E}$. 
We are concerned with a $K$ class classification problem with a sample space $\mathcal{X}$ and label space $\mathcal{Y} = \{1, 2, \hdots, K\}$. The classification model $\mathcal{M}$ is taken to be $g_{\bm \theta}:\mathcal{X}\rightarrow\mathcal{Y}$ parameterized with ${\bm \theta}$. The softmax output of the model is given by ${\bf p}$ =  $\text{softmax}(g_{{\bm \theta}}({\bf x})) \in \Delta^{K-1}$. The batch active learning setup starts with a set of labeled samples $\mathcal{D}_{tr} = \{({\bf x}_{i}, y_{i})\}$ and unlabeled samples $\mathcal{P} = \{({\bf x}_{j})\}$. With a query budget of $b$, we select a batch of unlabeled samples $\mathcal{B}$ as, $\mathcal{B} = \texttt{ALG}(\mathcal{D}_{tr}, \mathcal{M}, b, \mathcal{P}),\,\vert\mathcal{B}\vert\leq b$, where $\texttt{ALG}$ is the selection procedure conditioned on the current state of active learning $(\mathcal{D}_{tr}, \mathcal{M}, b, \mathcal{P})$. $\texttt{ALG}$ is designed with the aim of maximizing the prediction accuracy $\mathbb{E}_{p_{\mathcal{X}\times\mathcal{Y}}}[(h_{{\bm \theta}}({\bf x}) = y)]$. Henceforth, these samples which can potentially maximize the prediction accuracy are termed as \textit{important} samples. After each acquisition iteration, the training dataset is updated as $\mathcal{D}_{tr} = \mathcal{D}_{tr} \cup \{(\mathcal{B}, y_{\mathcal{B}})\}$ where $y_{\mathcal{B}}$ are the labels of $\mathcal{B}$ from an oracle routine.

The oracle takes an input ${\bf x}\in\mathcal{X}$ and outputs the ground truth label $y\in\mathcal{Y}$. This is referred to as `Ideal Oracle' and the mapping from ${\bf x}$ to $y$ is deterministic. A `Noisy Oracle' flips the true output $y$ to $y^{\prime}$ which is what we receive upon querying ${\bf x}$. Similar to \citep{Chen15b}, we assume that the label flipping is independent of the input ${\bf x}$ and thus can be characterized by the conditional probability $p(y^{\prime} = i\vert y = j)$, where $i, j \in \mathcal{Y}$. We also refer this conditional distribution as the noisy-channel, for rest of the paper, the $K$-SC is defined as follows
\begin{equation}
    p(y^{\prime} = i \vert y = j) = \{1-\varepsilon \quad\text{for} \quad i = j,\quad   \varepsilon / (K-1) \quad\text{for} \quad i\neq j \}
\end{equation}
\noindent where $\varepsilon$ is the probability of a label flip, i.e., $p(y^{\prime}\neq y) = \varepsilon$, for Ideal Oracle $\varepsilon=0$. We resort to the usage of $K$-SC because of its simplicity, and in addition, it abstracts the oracle noise strength with a single parameter $\varepsilon$. Therefore, in noisy active learning, after the selection of required subset $\mathcal{B}$, the training dataset (and then the model) is updated as  $\mathcal{D}_{tr} = \mathcal{D}_{tr} \cup \{(\mathcal{B}, y^{\prime}_{\mathcal{B}})\}$. 
\vspace*{-5pt}
\section{Method}
\vspace*{-8pt}
\label{sec:method}
\subsection{Batch Active Learning}
\label{ssec:BAL}
\begin{algorithm}[tb]
	\textbf{Input}: Initial training data $\mathcal{D}_{tr}^{(0)}$, pool of unlabeled samples $\mathcal{P}$,  model architecture $\mathcal{M}^{(0)}$, uncertainty inverse function $f(.)$, batch size $b$, number of AL iterations $T$\\
	\textbf{Output}: Selected batches $\mathcal{B}^{(t)}$, final model $\mathcal{M}^{(T)}$
	\begin{algorithmic}[1] 
		\FOR{$t=1,2,\hdots,T$}
		\STATE Assign importance score to each $x\in \mathcal{P}$ as $\rho_{x} = I({\bm \theta};y\vert {\bf x}, \mathcal{D}_{tr}^{(t-1)})$ \hfill $\triangleright$ Eq.\ref{eqn:baldDef}
		\STATE Perform Agglomerative clustering of the pool samples with $N(b)$ number of clusters using square root of JS-divergence as distance metric to get ${\bf D}$
		\STATE Compute uncertainty estimate $\sigma^{(t-1)}$ of the model $\mathcal{M}^{(t-1)}$, and $\beta^{(t-1)} = f(\sigma^{(t-1)})$
		\FOR{$i = 1, 2, \hdots, b$}
		\STATE Sample cluster centroid $c$ from the categorical distribution $p(c = k) \propto \rho_{k}$
		\STATE Sample $\zeta$ from the Gibbs distribution $p(\zeta = s \vert\mathcal{B}^{(t)}, c, \beta^{(t-1)}, {\bf D})$, and add to $\mathcal{B}^{(t)}$  \hfill $\triangleright$ Eq. \ref{eqn:gibbsDistr}
		\ENDFOR
		\STATE Query oracle for the labels of $\mathcal{B}^{(t)}$ and update $\mathcal{D}_{tr}^{(t)} \leftarrow \mathcal{D}_{tr}^{(t-1)} \cup \{(\mathcal{B}^{(t)}, y)\}$\\
		\STATE Update model as $\mathcal{M}^{(t)}$ using $\mathcal{D}_{tr}^{(t)}$ and set $\mathcal{P} \leftarrow \mathcal{P} \setminus \mathcal{B}^{(t)}$
		\ENDFOR
	\end{algorithmic}
	\caption{Batch Active Learning}
	\label{alg:bal-algorithm}
\end{algorithm}
An ideal batch selection procedure so as to be employed in an active learning setup, must address the following issues, (i) select important samples from the available pool for the current model, and (ii) select a diverse batch to avoid repetitive samples. We note that, at each step, when active learning acquires new samples, both of these issues are addressed by using the currently trained model. However, in the event of an uncertain model, the quantification  of diversity and importance of a batch of samples will also be inaccurate resulting in loss of performance. This is often the case with active learning because we start with less data in hand and consequently an uncertain model. Therefore, we identify the next problem in the active learning as (iii) incorporation of the model uncertainty across active learning iterations.
\paragraph{Batch selection}
The construction of batch active learning algorithm by solving the aforementioned first two problems begins with assignment of an importance score $(\rho)$ to each sample in the pool. Several score functions exist which perform sample wise active learning. To list a few, max-entropy, variation ratios, BALD \citep{Gal:2017}, entropy of the predicted class probabilities \citep{Wang2014AL}. We use BALD as an importance score which quantifies the amount of reduction of uncertainty by incorporating a particular sample for the given model. In principle, we wish to have high BALD score for a sample to be selected. For the sake of completeness, it is defined as follows.
\begin{eqnarray}
\mathbb{I}(y;{\bm \theta}\vert {\bf x}, \mathcal{D}_{tr}) = \mathbb{H}(y\vert {\bf x}, \mathcal{D}_{tr}) - \mathbb{E}_{{\bm \theta}\vert\mathcal{D}_{tr}}\mathbb{H}(y\vert{\bm \theta}, {\bf x}),
\label{eqn:baldDef}
\end{eqnarray}
\noindent where ${\bm \theta}$ are the model parameters. We refer the reader to \citep{Gal:2017} for details regarding the computation of BALD score in (\ref{eqn:baldDef}). To address diversity, we first perform clustering of the pooled samples and then use importance sampling to select cluster centroids. For clustering, the distance metric used is the square root of the Jensen-Shannon (JS) divergence between softmax output of the samples. Formally, for our case, it is defined as $d: \Delta^{K-1}\times \Delta^{K-1}\rightarrow [0,1]$, where $d({\bf p}, {\bf q}) = \sqrt{(KL({\bf p}\vert\vert ({\bf p}+{\bf q})/2) + KL({\bf q}\vert\vert ({\bf p}+{\bf q})/2))/2}$. With little abuse of notation, we interchangeably use $d({\bf p}_{i}, {\bf p}_{j})$ as $d_{i,j}$ where $i, j$ are the sample indices and ${\bf p}_{i}, {\bf p}_{j}$ are corresponding softmax outputs. The advantage of using JS-divergence is two folds; first it captures similarity between probability distributions well, second, unlike KL-divergence it is always bounded between $0$ and $1$. The boundedness helps in incorporating uncertainty which we will discuss shortly. Using the distance metric as $d$ we perform Agglomerative hierarchical clustering \citep{Rokach2005} for a given number of clusters $N$. A cluster centroid is taken as the median score sample of the cluster members. Finally, with all similar samples clustered together, we perform importance sampling of the cluster centroids using their importance score, and a random centroid $c$ is selected as $p(c = k) \propto \rho_{k}$. The clustering and importance sampling together not only take care of selecting important samples but also ensure diversity among the selected samples.

\paragraph{Uncertainty Incorporation}
The discussion we have so far is crucially dependent on the output of the model in hand, i.e., importance score as well as the similarity distance. As noted in our third identified issue with active learning, of model uncertainty, these estimations suffers from inaccuracy in situations involving less training data or uncertain model. The uncertainty of a model, in very general terms, represents the model's confidence of its output. The uncertainty for deep learning models has been approximated in Bayesian settings using dropout in \citep{gal16}, and batch normalization (BN) in \citep{Teye2018BayesianUE}. Both use stochastic layers (dropout, BN) to undergo multiple forward passes and compute the model's confidence in the outputs. For example, confidence could be measured in terms of statistical dispersion of the softmax outputs. In particular, variance of the softmax outputs, variation ratio of the model output decision, etc, are good candidates. We denote the model uncertainty as $\sigma\in[0,1]$, such that $\sigma$ is normalized between $0$ and $1$ with $0$ being complete certainty and $1$ for fully uncertain model. For rest of the work, we compute the uncertainty measure $\sigma$ as variation ratio of the output of model's multiple stochastic forward passes as mentioned in \citep{gal16}.

In the event of an uncertain model ($\sigma\rightarrow 1$), we randomly select samples from the pool initially. However, as the model moves towards being more accurate (low $\sigma$) by acquiring more labeled samples through active learning, the selection of samples should be biased towards importance sampling and clustering. To mathematically model this solution, we use the statistical mechanics approach of deterministic annealing using the Boltzmann-Gibbs distribution \citep{Rose_PhysRevLett.65.945}. In Gibbs distribution $p(i) \propto e^{-\epsilon_{i}/k_{B}T}$, i.e., probability of a system being in an $i$th state is high for low energy $\epsilon_{i}$ states and influenced by the temperature $T$. For example, if $T\rightarrow \infty$, then state energy is irrelevant and all states are equally probable, while if $T\rightarrow 0$, then probability of the system being in the lowest energy state is almost surely $1$. 

We translate this into active learning as follows: For a given cluster centroid $c$, if the model uncertainty is very high $(\sigma\rightarrow 1)$ then all points in the pool (including $c$) should be equally probable to get selected (or uniform random sampling), and if the model is very certain $(\sigma\rightarrow 0)$, then the centroid $c$ itself should be selected. This is achieved by using the state energy analogue as distance $d$ between the cluster centroid $c$ and any sample $x$ in the pool, and temperature analogue as uncertainty estimate $\sigma$ of the model. The distance metric $d$ used by us is always bounded between $0$ and $1$ and it provides nice interpretation for the state energy. Since, in the event of low uncertainty, we wish to perform importance sampling of cluster centroids, and we have $d_{c, c} = 0$ (lowest possible value), therefore by Gibbs distribution, cluster centroid $c$ is selected almost surely. 
\begin{algorithm}[tb]
	\textbf{Input}: Initial training data $\mathcal{D}_{tr}^{(0)}$, pool of unlabeled samples $\mathcal{P}$, $\mathcal{B}^{(0)}=\phi$,  model architecture $\mathcal{M}^{(0)}$, batch size $b$, number of AL iterations $T$, active learning Algorithm $\texttt{ALG}$\\
	\textbf{Output}: Selected batches $\mathcal{B}^{(t)}$, final model $\mathcal{M}^{(T)}$
	\begin{algorithmic}[1] 
		
		\FOR{$t=1,2,\hdots,T$}
		\STATE $\mathcal{B}^{(t)} \leftarrow \texttt{ALG}(\mathcal{D}_{tr}^{(t-1)}, \mathcal{M}^{(t-1)}, b,  \mathcal{P}\setminus \bigcup\nolimits_{t'\leq t-1} \mathcal{B}^{(t')})$
		\STATE Query noisy oracle for the labels of $\mathcal{B}^{(t)}$ and update $\mathcal{D}_{tr}^{(t)} \leftarrow \mathcal{D}_{tr}^{(t-1)} \cup \{(\mathcal{B}^{(t)}, y')\}$\\
		\STATE Get $\mathcal{M}^{\ast\,(t)} \leftarrow \mathcal{M}^{(t)}$ appended with noisy-channel layer at the end
		\STATE Update noisy model as $\mathcal{M}^{\ast\,(t)}$ using $\mathcal{D}_{tr}^{(t)}$
		\STATE Detach required model $\mathcal{M}^{(t)}$ from $\mathcal{M}^{\ast\,(t)}$ by removing the final noisy-channel layer
		\ENDFOR
	\end{algorithmic}
	\caption{Noisy Oracle Active Learning}
	\label{alg:denoisification}
\end{algorithm}

To construct a batch, the samples have to be drawn from the pool using Gibbs distribution without replacement. In the event of samples $s_{1}, \hdots, s_{n}$ already drawn, the probability of drawing a sample $\zeta$ given the cluster centroid $c$, distance matrix ${\bf D} = [d_{i,j}]$ and inverse temperature (or inverse uncertainty) $\beta$ is written as
\vspace*{-6pt}
\begin{align}
\zeta \vert s_{1:n}, c, \beta, {\bf D} &\sim \text{Categorical}\left(\frac{e^{-\beta\,d_{c,1}}}{\sum\limits_{s^{\prime}\in \mathcal{P}^{\prime}}e^{-\beta\,\,d_{c,s^{\prime}}}},\frac{e^{-\beta\,d_{c,2}}}{\sum\limits_{s^{\prime}\in \mathcal{P}^{\prime}}e^{-\beta\,\,d_{c,s^{\prime}}}}, \hdots, \frac{e^{-\beta\,d_{c,\vert P^{\prime}\vert}}}{\sum\limits_{s^{\prime}\in \mathcal{P}^{\prime}}e^{-\beta\,\,d_{c,s^{\prime}}}}\right),
\label{eqn:gibbsDistr}
\end{align}
\noindent where $\mathcal{P}^{\prime} = \mathcal{P}\setminus s_{1:n}$. In theory, the inverse uncertainty $\beta$ can be any $f$ such that $f:[0,1]\rightarrow\mathbb{R}^{+}\cup\{0\}$ and $f(\sigma)\rightarrow\infty$ as $\sigma\rightarrow 0$ and $f(\sigma) = 0$ for $\sigma = 1$. For example, few possible choices for $\beta\, (=f(\sigma))$  are $-\log(\sigma)$, $e^{1/ \sigma}-1$. Different inverse functions will have different growth rate, and the choice of functions is dependent on both the model and the data. Next, since we have drawn the cluster centroid $c$ according to $p(c = k)\propto \rho_{k}$, the probability of drawing a sample $s$ from the pool $\mathcal{P}$ is written as
\begin{equation}
p(\zeta = s \vert s_{1:n},\beta, {\bf D}) = \sum\limits_{c=1}^{N}\frac{\rho_{c}}{\sum\nolimits_{c^{\prime}}\rho_{c^{\prime}}}.\frac{e^{-\beta\,d_{c,s}}}{\sum\nolimits_{s^{\prime}\in \mathcal{P}^{\prime}}e^{-\beta\,\,d_{c,s^{\prime}}}}.
\label{eqn:finalRandomSample}
\end{equation}
We can readily see that upon setting $\beta\rightarrow 0$ in (\ref{eqn:finalRandomSample}), $p(\zeta = s \vert s_{1:n},\beta, {\bf D})$ reduces to $1/\vert\mathcal{P}^{\prime}\vert$ which is nothing but the uniform random distribution in the leftover pool. On setting $\beta\rightarrow\infty$, we have $\zeta = c$ with probability $\rho_{c}/\sum\nolimits_{c^{\prime}}\rho_{c^{\prime}}$ and $\zeta \neq c$ with probability $0$, i.e., selecting cluster centroids from the pool with importance sampling. For all other $0<\beta<\infty$ we have a soft bridge between these two asymptotic cases. The approach of uncertainty based batch active learning is summarized as Algorithm\,\ref{alg:bal-algorithm}. Next, we discuss the solution to address noisy oracles in the context of active learning.

\subsection{Noisy Oracle}
\label{ssec:noisyOracle}
The noisy oracle, as defined in Section\,\ref{sec:probForm}, has non-zero probability for outputting a wrong label when queried with an input sample. To make the model aware of possible noise in the dataset originating from the noisy oracle, we append a denoising layer to the model. The inputs to this denoising layer are the softmax outputs ${\bf p}$ of the original model. For a pictorial representation, we refer the reader to supplementary materials Section\,2. The denoising layer is a fully-connected $K\times K$ dense layer with weights ${\bf W} = [w_{i,j}]$ such that its output ${\bf p}^{\prime} = {\bf W}{\bf p}$. The weights $w_{i,j}$ represent the noisy-channel transition probabilities such that $w_{i,j} = p(y^{\prime} = i\vert y = j)$. Therefore, to be a valid noisy-channel, ${\bf W}$ is constrained as ${\bf W}\in \{{\bf W}\,\vert\, {\bf W}_{.,j}\in \Delta^{K-1}, \,\forall\, 1\leq j \leq K \}$.
While training we use the model upto the denoising layer and train using ${\bf p}^{\prime}$, or label prediction $y^{\prime}$ while for validation/testing we use the model output ${\bf p}$ or label prediction $y$. The active learning algorithm in the presence of noisy oracle is summarized as Algorithm\,\ref{alg:denoisification}. We now proceed to Section\,\ref{sec:experi} for demonstrating the efficacy of our proposed methods across different datasets.

\vspace*{-5pt}
\section{Experiments}
\label{sec:experi}
\subsection{Setup}
\label{ssec:experiSet}
We evaluate the algorithms for training CNNs on four datasets pertaining to image classification; (i) MNIST \citep{Lecun98gradient-basedlearning}, (ii) CIFAR10, \citep{Krizhevsky09learningmultiple}, (iii) SVHN \citep{Netzer2011ReadingDI}, and (iv) Extended MNIST \cite{cohen2017emnist}. All datasets have 10 classes except EMNIST, which has 47 classes. We use the CNN architectures from \citep{fchollet,Gal:2017}. The implementations are done on PyTorch, and we use the Scikit-learn \citep{scikit-learn} package for Agglomerative clustering. For further details, we refer the reader to the supplementary materials.

For training the denoising layer, we initialize it with the identity matrix ${\bf I}_{K}$, i.e., assuming it to be noiseless. The number of clusters $N(b)$ is taken to be as $\floor{5b}$. The uncertainty measure $\sigma$ is computed as the variation ratio of the output prediction across $100$ stochastic forward passes, as coined in \citep{gal16}, through the model using a validation set which is fixed apriori. The inverse uncertainty function $\beta = f(\sigma)$ in Algorithm\,\ref{alg:bal-algorithm} is chosen from $l\,(e^{1/\sigma}-1)$, $-l\log(\sigma)$, where $l$ is a scaling constant fixed using cross-validation. The cross-validation is performed only for the noise-free setting, and all other results with different noise magnitude $\varepsilon$ follow this choice. This is done so as to verify the robustness of the choice of parameters against different noise magnitudes which might not be known apriori.
\begin{figure*}[t]
	\centering
	\begin{subfigure}[b]{\linewidth}
		\includegraphics*[height = 1.5in]{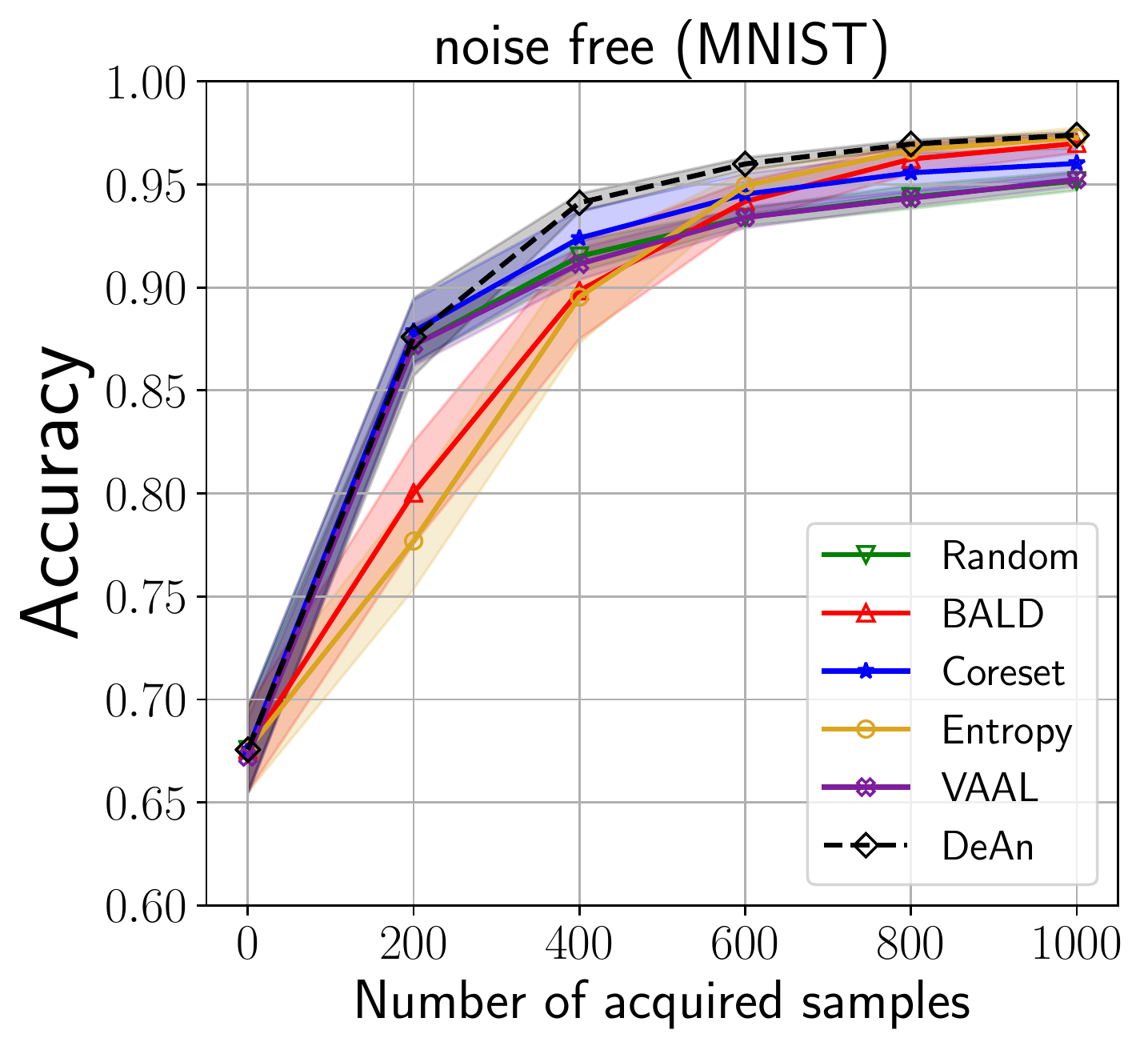}
		\includegraphics*[height = 1.5in]{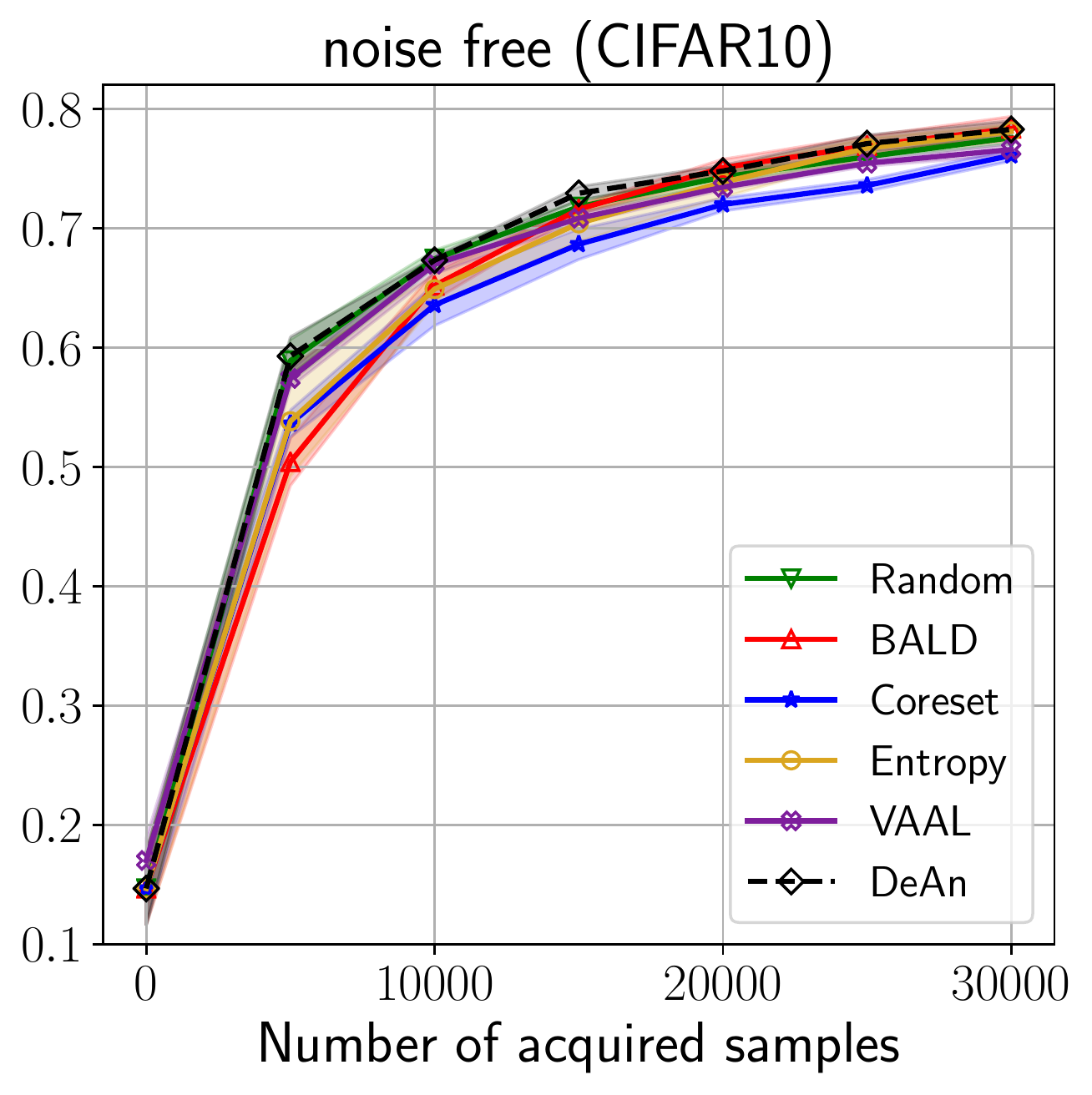}
		\includegraphics*[height = 1.5in]{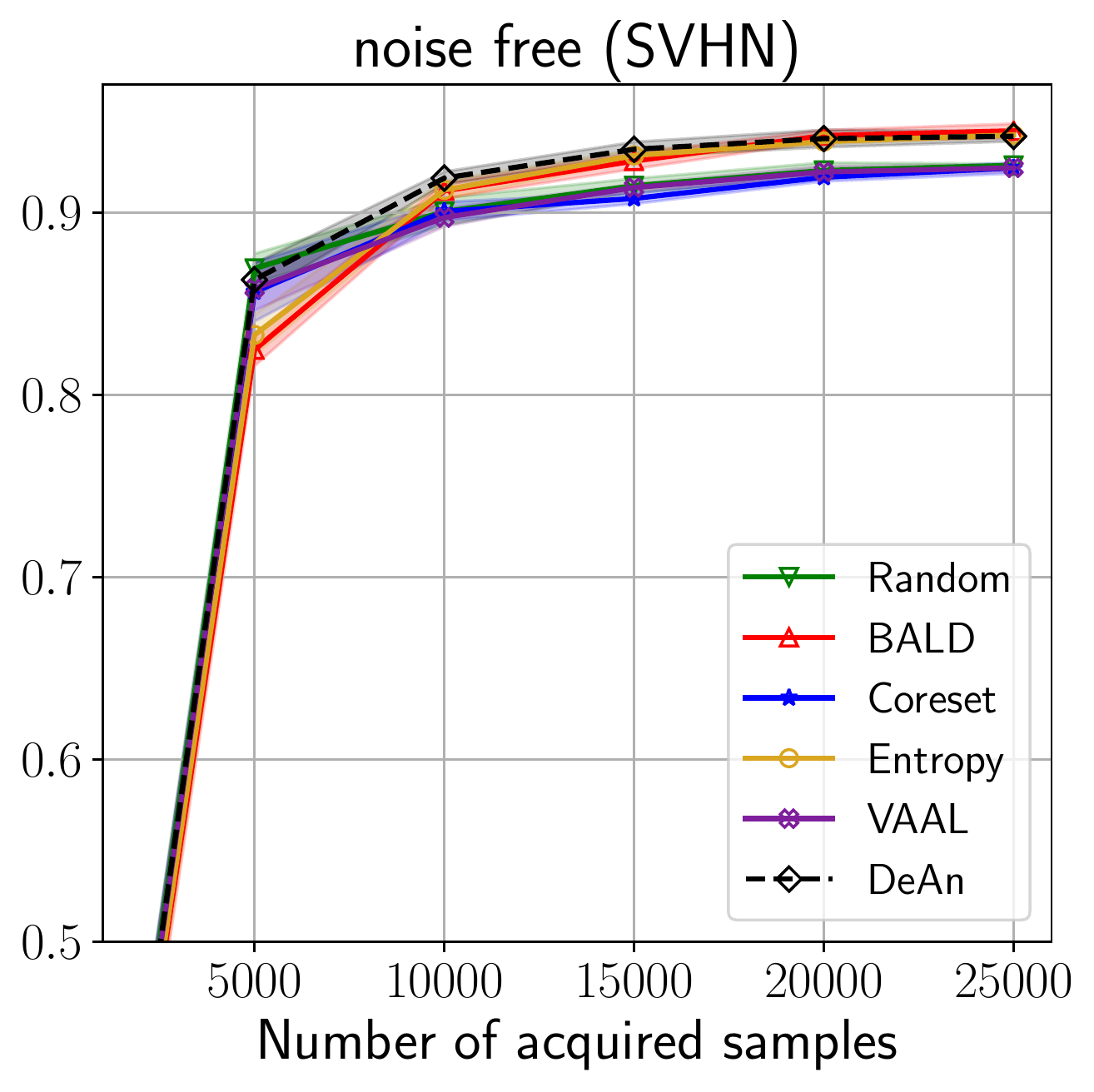}
		\includegraphics*[height = 1.5in]{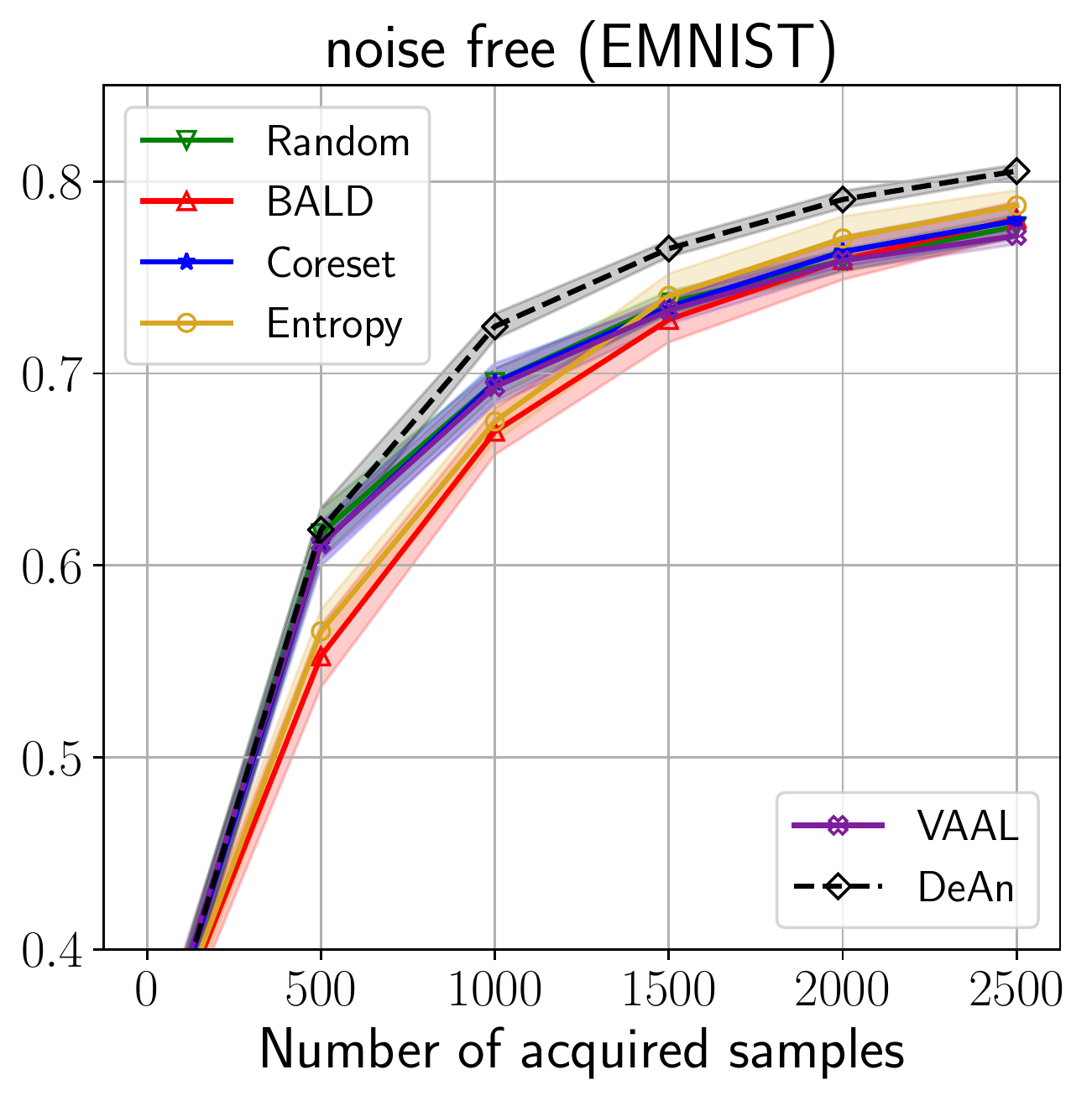}
	\end{subfigure}\\
	\begin{subfigure}[b]{\linewidth}
		\includegraphics*[height = 1.5in]{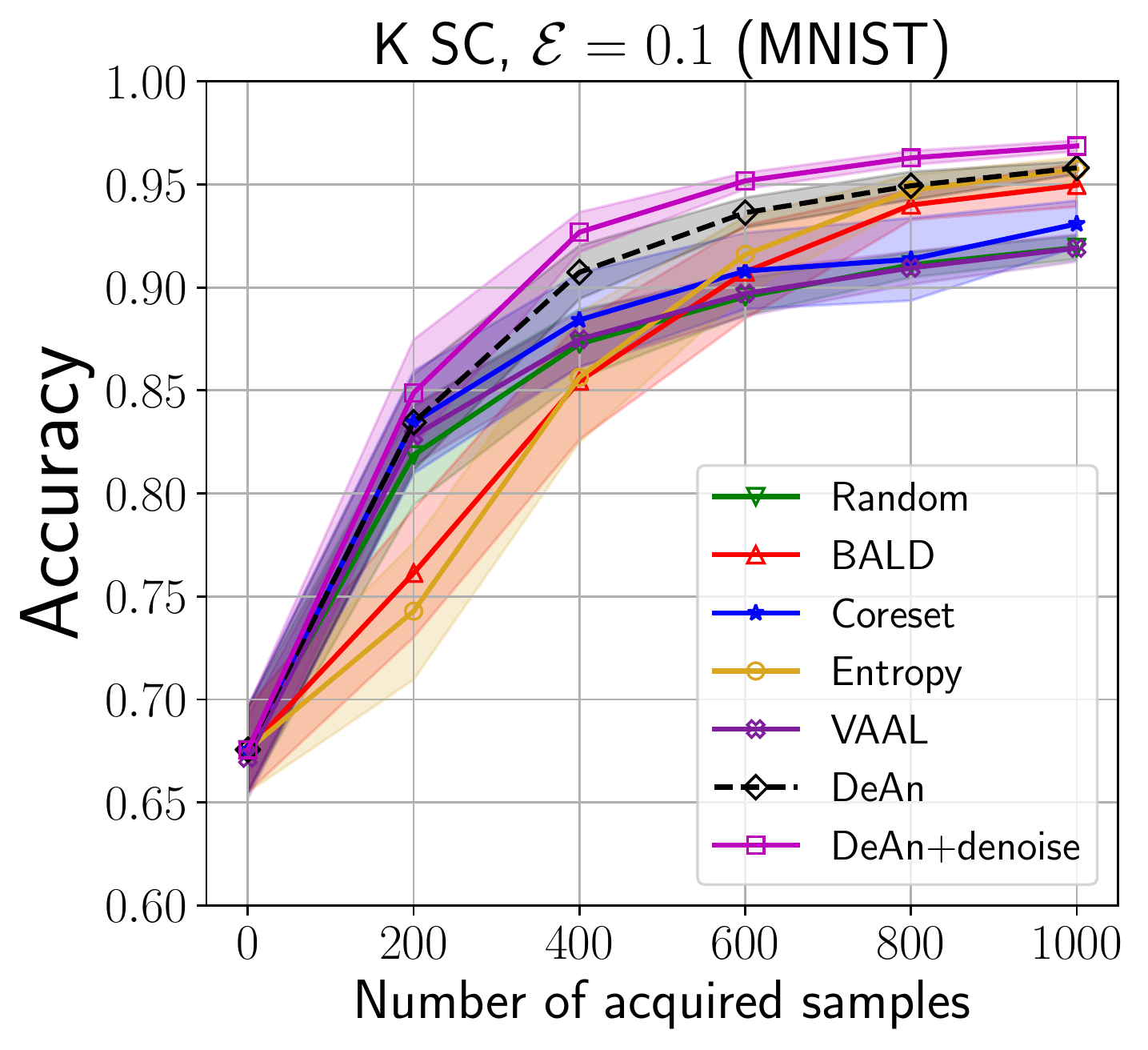}
		\includegraphics*[height = 1.5in]{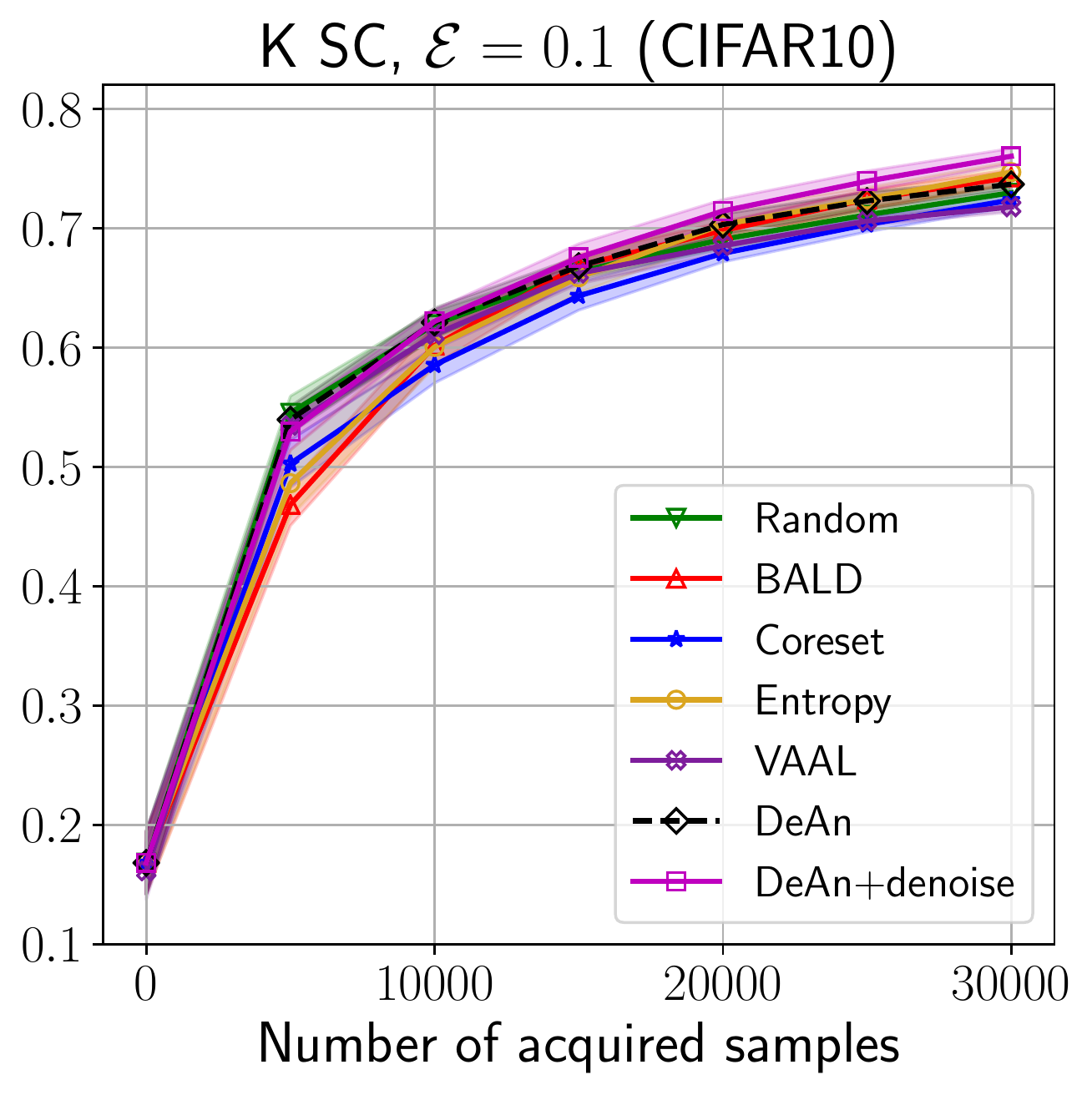}
		\includegraphics*[height = 1.5in]{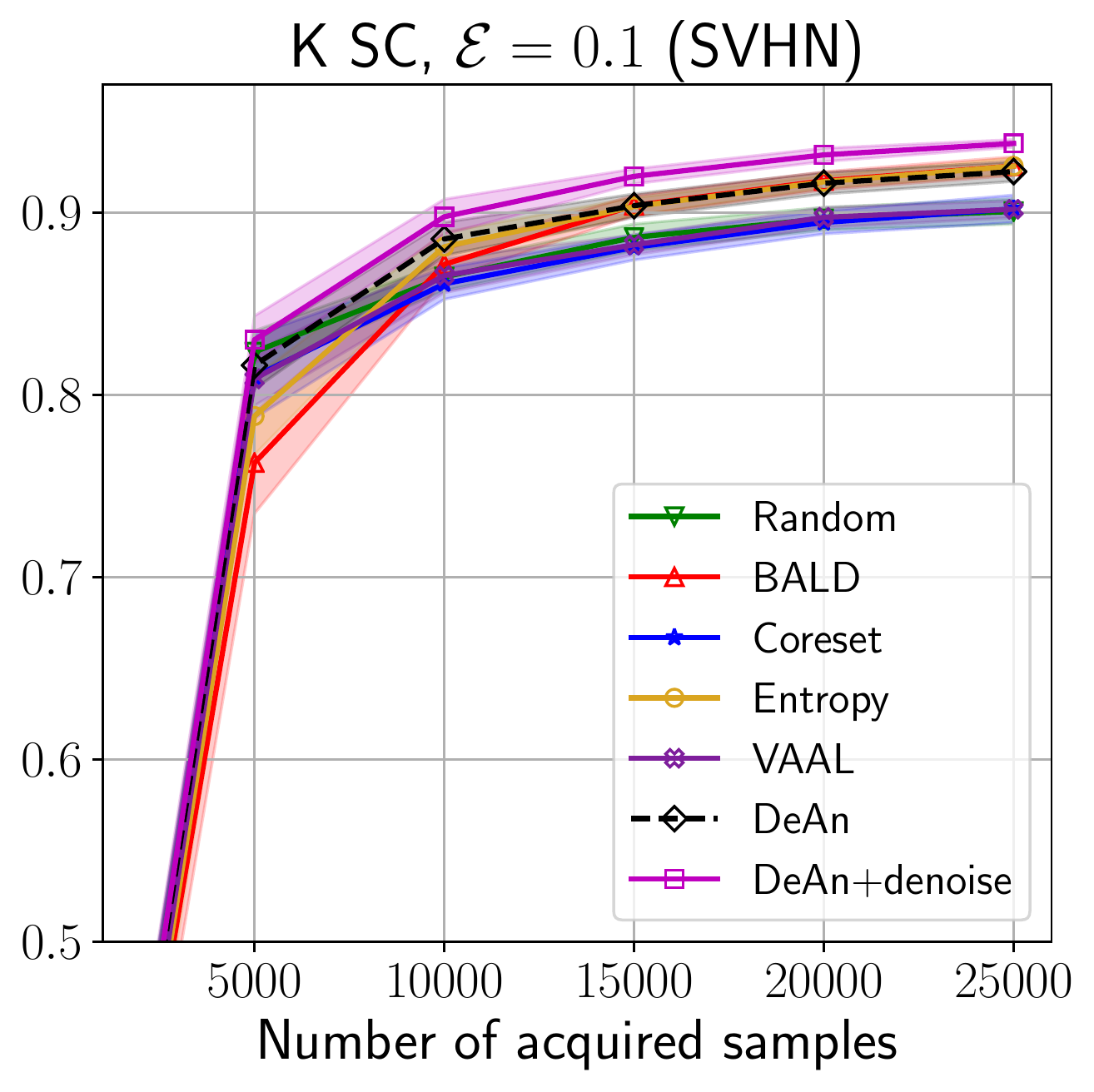}
		\includegraphics*[height = 1.5in]{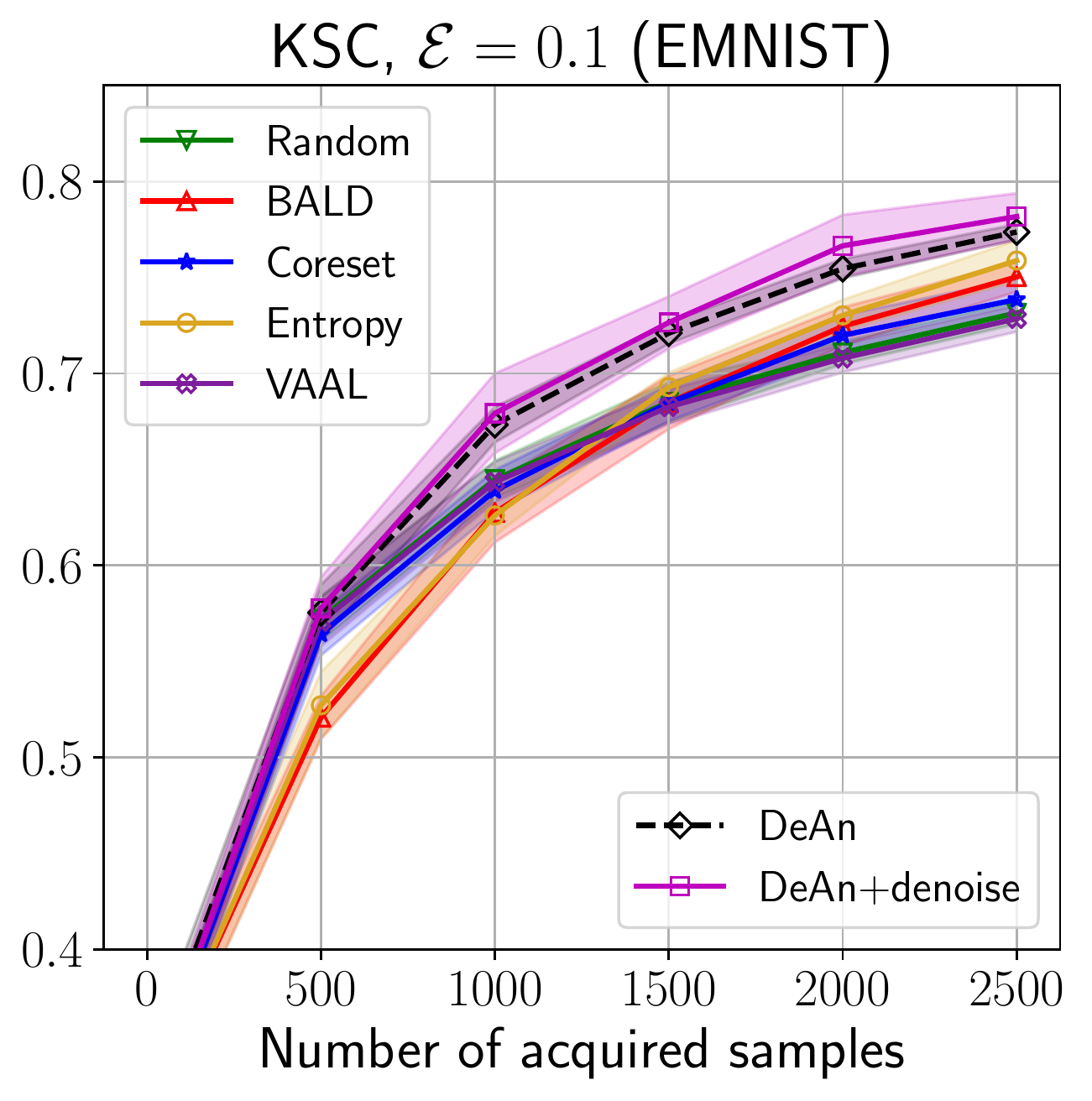}
	\end{subfigure}\\
	\begin{subfigure}[b]{\linewidth}
		\includegraphics*[height = 1.5in]{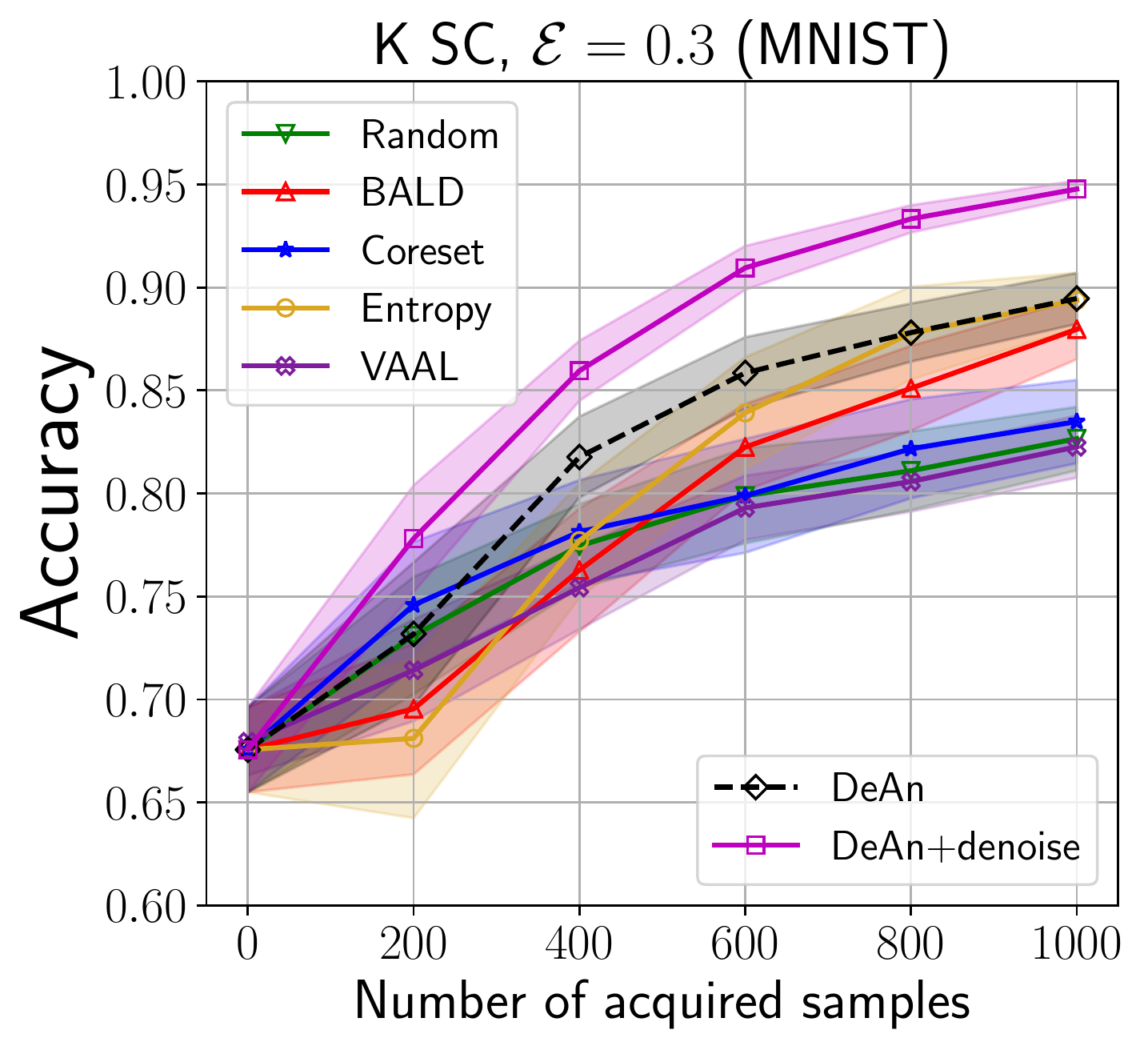}
		\includegraphics*[height = 1.5in]{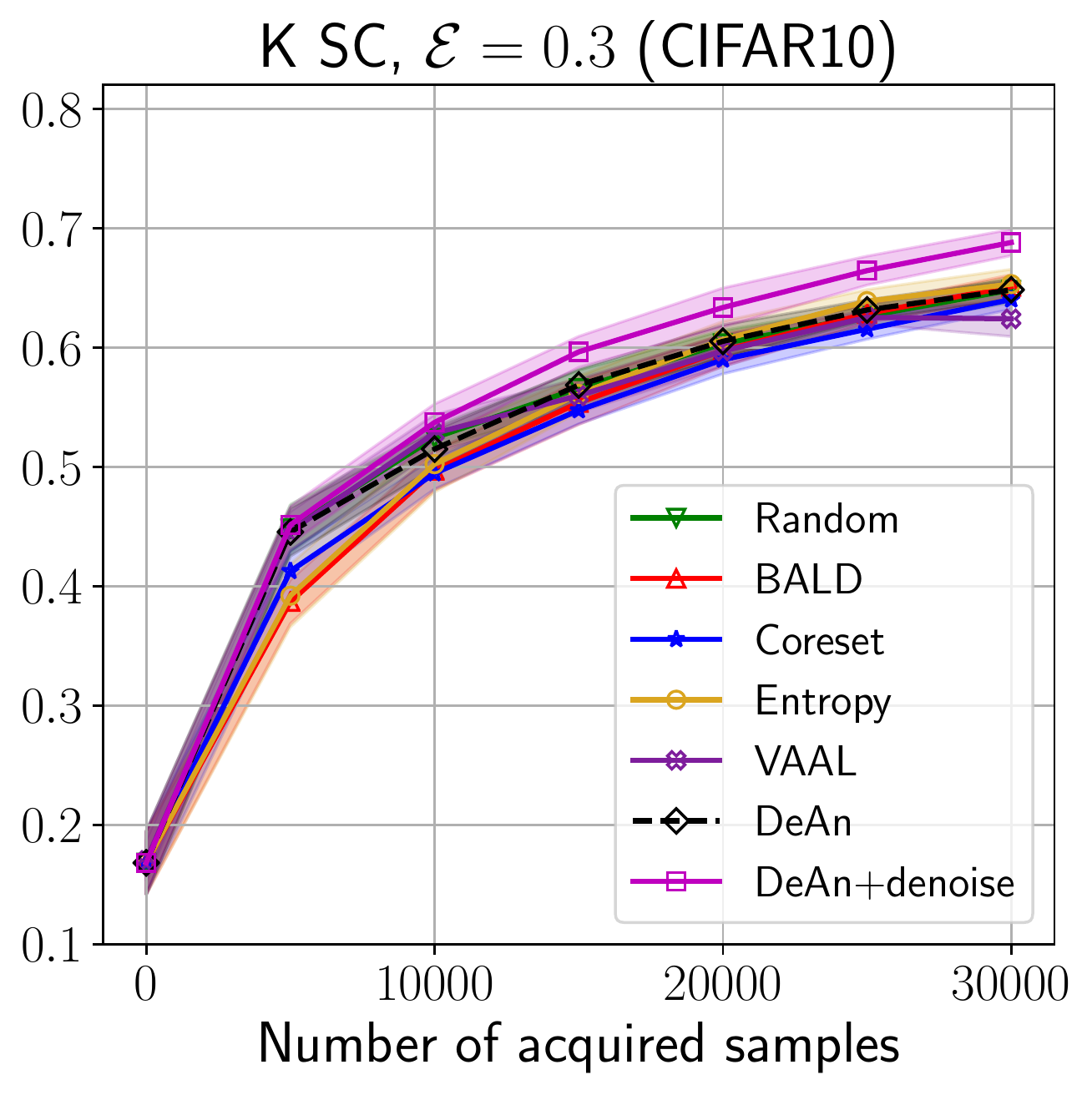}
		\includegraphics*[height = 1.5in]{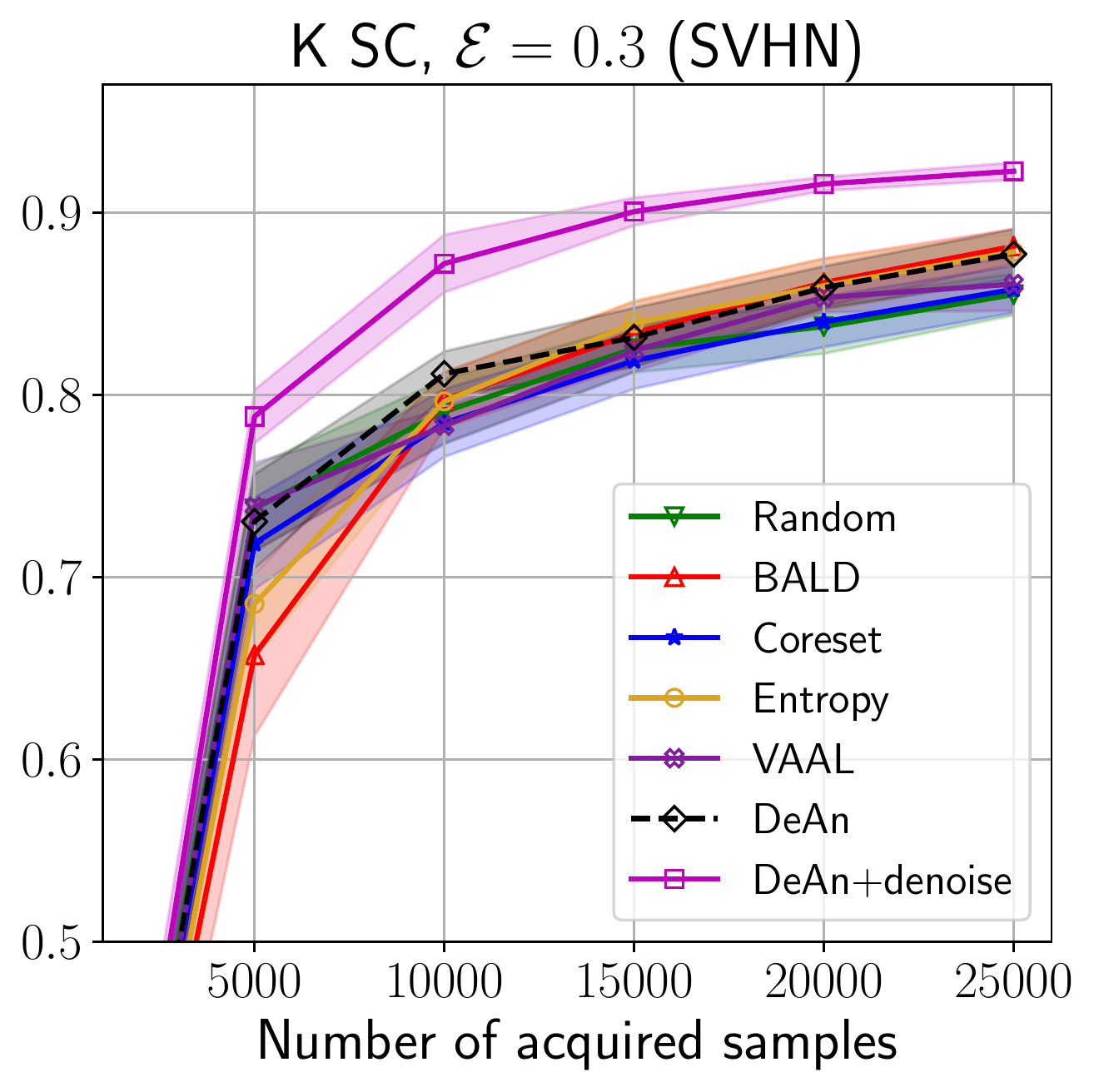}
		\includegraphics*[height = 1.5in]{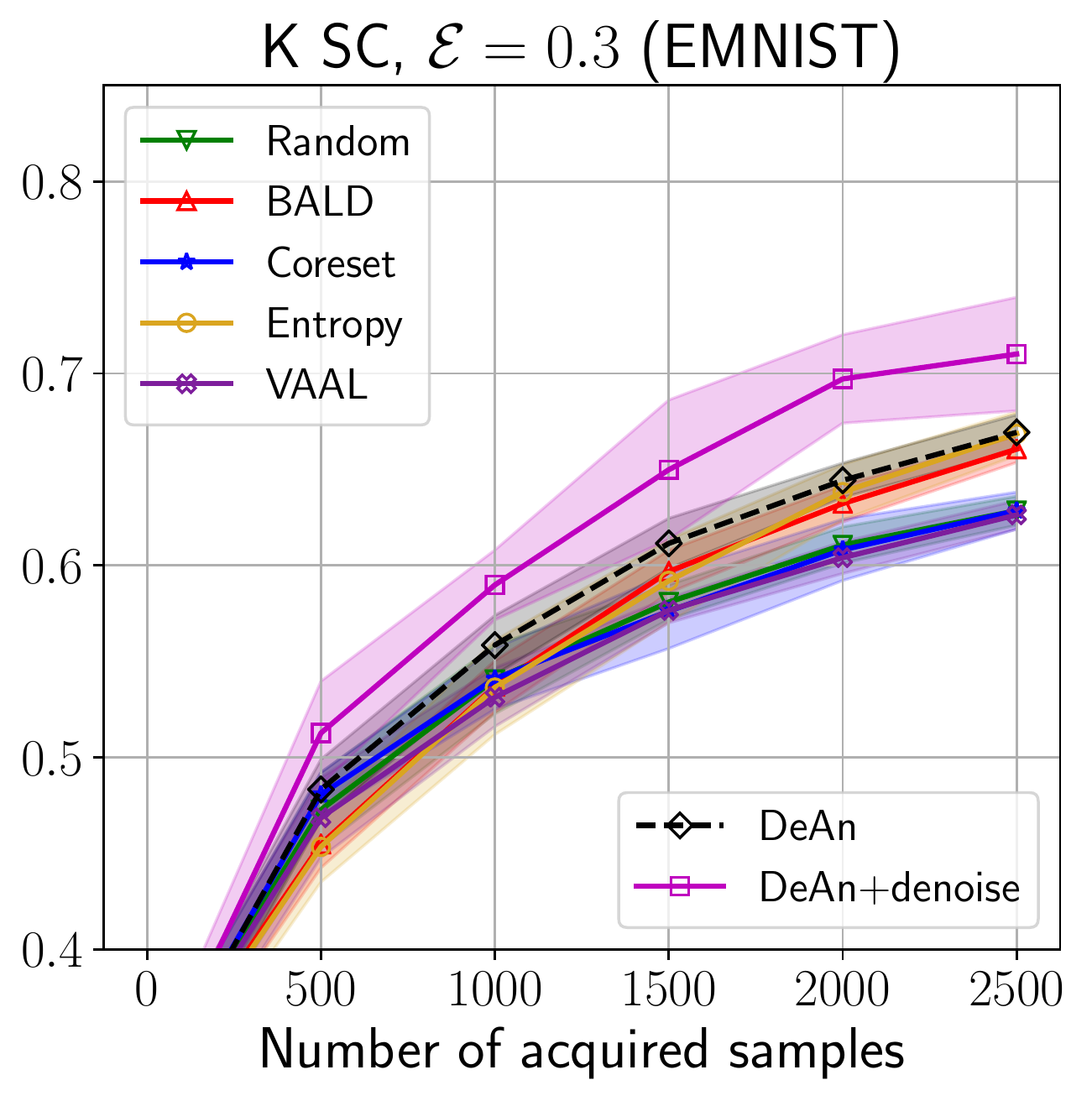}
	\end{subfigure}
	\caption{Active learning results for various algorithms under different levels of noise strength in the oracle decision (noise free, $\varepsilon = 0.1$ and $0.3$) for MNIST, CIFAR10, SVHN, and EMNIST Image datasets.}
	\label{fig:mainResults}
\end{figure*}

\subsection{Results}
\label{ssec:experiResults}
We compare our proposed \textbf{De}terministic \textbf{An}nealing based approach called \textbf{\texttt{DeAn}}\footnote{The code for reproducing all the results is available at \texttt{https://github.com/gaurav71531/DeAn}.} with: (i) \textbf{\texttt{Random}}: A batch is selected by drawing samples from the pool uniform at random without replacement. (ii) \textbf{\texttt{BALD}}: Using model uncertainty and the BALD score, the authors in \citep{Gal:2017} do active learning with single sample acquisition. We use the highest $b$ scoring samples to select a batch. (iii) \textbf{\texttt{Coreset}}: The authors in \citep{sener2018active} proposed a coreset based approach to select the  representative core centroids of the pool set. We use the $2-OPT$ approximation greedy algorithm of the paper. (iv) \textbf{\texttt{Entropy}}: The approach of \citep{Wang2014AL} is implemented via selecting $b$ samples with the highest Shannon entropy $\mathbb{H}({\bf p})$ of the softmax outputs. (v) \textbf{\texttt{VAAL}}: The variational adversarial active learning of \citep{Samarth_ICCV_19}. (vi) \textbf{\texttt{BBALD}}: The batch version of BALD as proposed in \citep{kirsch2019batchbald}. The BBALD has difficulty in acquiring larger batches due to the exponential ($K^{b}$) computations needed or large Monte Carlo samples for sufficient accuracy. Nonetheless, we compare against this approach for acquisition size of 100 (MNIST) which is still double the maximum of what considered in the original work.

In all our experiments, we start with a small number of images $2-5$ per class and retrain the model from scratch after every batch acquisition. In order to make a fair comparison, we provide the same initial point for all active learning algorithms in an experiment. We perform a total of $20$ random initialization and plot the average performance along with the standard deviation vs number of acquired samples by the algorithms.

\figurename\,\ref{fig:mainResults} shows that our proposed algorithm outperforms all the existing algorithms. As an important observation, we note that random selection always works better in the initial stages of all experiments. This observation is explained by the fact that all models suffer from inaccurate predictions at the initial stages. The proposed uncertainty based randomization makes a soft bridge between uniform random sampling and score based importance sampling of the cluster centroids. The proposed approach uses randomness at the initial stages and then learns to switch to weigh the model based inference scores as the model becomes increasingly certain of its output. Therefore, \texttt{DeAn} always envelops the performance of all the other approaches across all four datasets of MNIST, CIFAR10, SVHN, and EMNIST. Results with other parameter combinations and datasets show similar trend, which we have presented in the supplementary materials.

\figurename\,\ref{fig:mainResults} also shows the negative impact of noisy oracle on the active learning performance across all four datasets. The degradation in the performance worsens with increasing oracle noise strength $\varepsilon$. We see that doing denoisification by appending noisy-channel layer helps combating the noisy oracle in \figurename\,\ref{fig:mainResults}. The performance of the proposed noisy oracle active learning is significantly better in all the cases. The prediction accuracy gap between algorithm with/without denoising layer elevates with increase in the noise strength $\varepsilon$.

The baselines like \texttt{VAAL}, \texttt{Coreset} which make representation of the Training + Pool may not always perform well. While coreset uses model output which suffers in the beginning due to model uncertainty, VAAL uses training data only to make representations together with the remaining pool in GAN like setting. The representative of pool points may not always help, especially if there are difficult points to label. In addition to the importance score, the model uncertainty is needed to \textit{assign a confidence to its judgement} which is poor in the beginning and gets strengthened later. The proposed approach works along this direction. Lastly, while robustness against oracle noise is discussed in \citep{Samarth_ICCV_19}, however, we see that incorporating the denoising later implicitly in the model helps better. The intuitive reason being, having noise in the training data changes the discriminative distribution from $p(y\vert {\bf x})$ to $p(y^{\prime}\vert {\bf x})$. Hence, learning $p(y^{\prime}\vert {\bf x})$ from the training data and then recovering $p(y\vert {\bf x})$ makes more sense as discussed in Section\,\ref{ssec:noisyOracle}. The recent batch version of BALD called $\texttt{BBALD}$ which apart from exponential complexity, also suffers from model uncertainty in the beginning as it strictly use model based scores to compute/approximate the joint mutual information as we next in the Section\,\ref{ssec:ablation}.

The uncertainty measure $\sigma$ plays a key role for the proposed algorithm. We have observed that under strong noise influence from the oracle, the model's performance is compromised due to spurious training data as we see in \figurename\ref{fig:mainResults}. This affects the estimation of the uncertainty measure (variation
{\parfillskip0pt\par}
\begin{wraptable}{r}{0.47\textwidth}
	\caption{Uncertainty $\sigma$ across active learning experiment iterations (t) for K-SC ($\varepsilon=0.3$).}
	\label{tab:modelUncertainty}
	\centering
	\begin{tabular}{lllll}
	\toprule
	\multirow{2}*{t} & \multicolumn{2}{c}{MNIST} & \multicolumn{2}{c}{SVHN}\\
	\cmidrule{2-5}
	 & Regular & denoise & Regular & denoise\\
	 \midrule
	1 &0.25 &0.25&0.42&0.42\\
	2 &0.25&0.21&0.28&0.11\\
	3 &0.21&0.15&0.23&0.10\\
	4 &0.20&0.11&0.22&0.08\\
	5 &0.19&0.10&0.21&0.07\\
	\bottomrule
	\end{tabular}
	\vspace*{-10pt}
\end{wraptable}
ratio) as well. We see in Table\,\ref{tab:modelUncertainty} that the model uncertainty does not drop as expected due to the label noise. However, the aid provided by the denoising layer to combat the oracle noise solves this issue. We observe in Table\,\ref{tab:modelUncertainty} that uncertainty drops at a faster rate as the model along with the denoising layer gets access to more training data. Hence, the proposed algorithm along with the denoising layer make better judgment of soft switch between uniform randomness and importance sampling using (\ref{eqn:finalRandomSample}). The availability of better uncertainty estimates for modern deep learning architectures is a promising future research, and the current work will also benefit from it.

\subsection{Ablation Study}
\label{ssec:ablation}
 The two key concepts of the current work are, (i) deterministic annealing based selection, and (ii) denoising layer to tackle oracle noise. We carefully study each of these separately as follows. 

\textbf{Uncertainty Incorporation} Active learning algorithms strictly utilizing model based scores suffer performance degradation if the model is uncertain as shown in the motivation in Introduction. We simulate this situation by reducing the number of training epochs from 50 (Regular) to 20 (Uncertain). The seed dataset, $4$ samples per class, is same for both the cases. From Figure\,\ref{sfig:MNIST_intro_uncertain}, we note that most algorithms, for example, $\texttt{BBALD}, \texttt{Coreset}, \texttt{Entropy}$ perform poorly and even fall below $\texttt{Random}$ selection. $\texttt{BALD}, \texttt{Entropy}$ drops below the initial point due to class imbalance created by inaccurate scores. On the other hand, $\texttt{DeAn}$ is able to cope by carefully utilizing the model uncertainty and make smooth switch between randomness and score based importance sampling.

\begin{figure}[t]
	\centering
	\begin{subfigure}[b]{\linewidth}
		\includegraphics*[height = 1.5in]{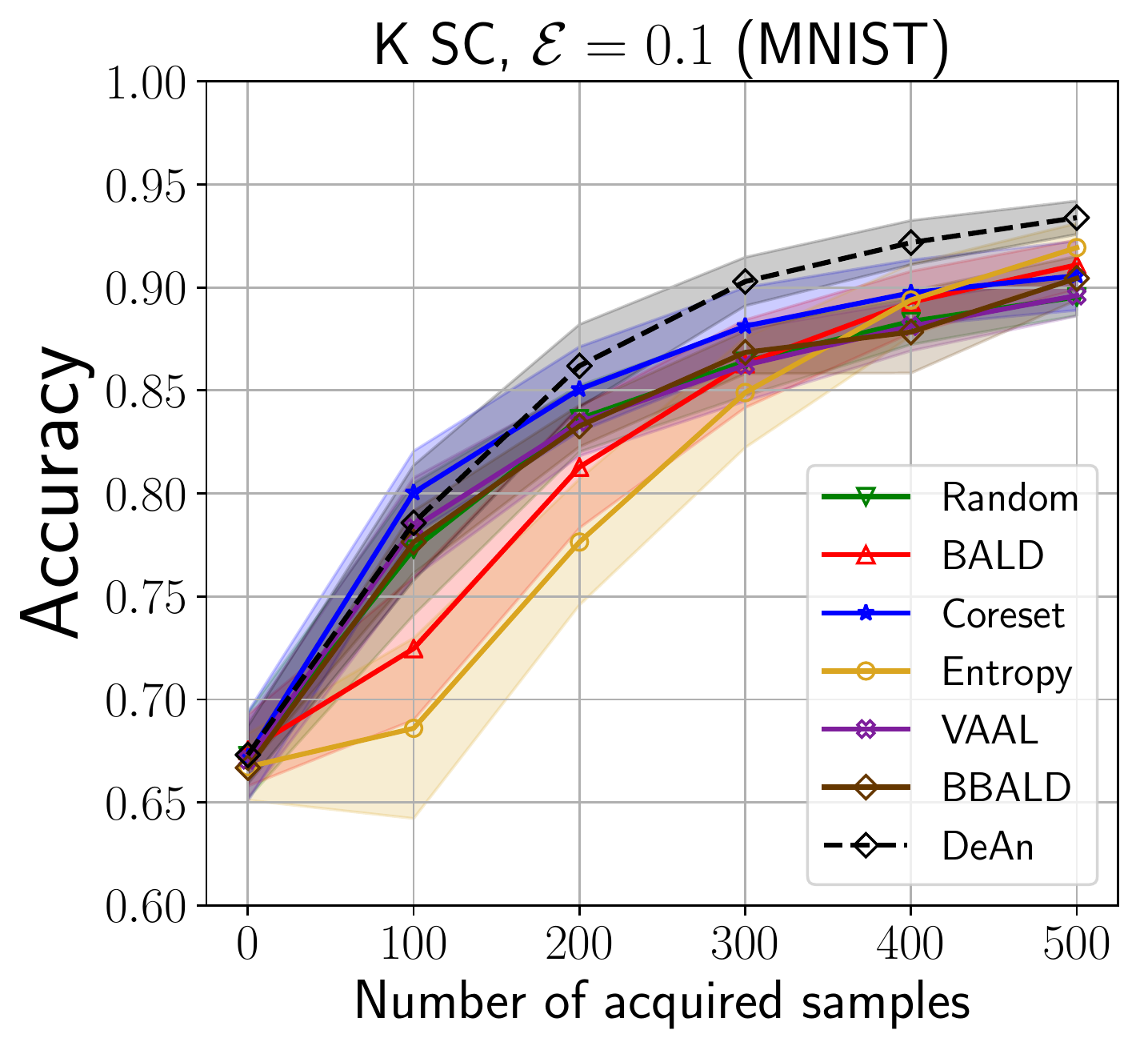}
		\includegraphics*[height = 1.5in]{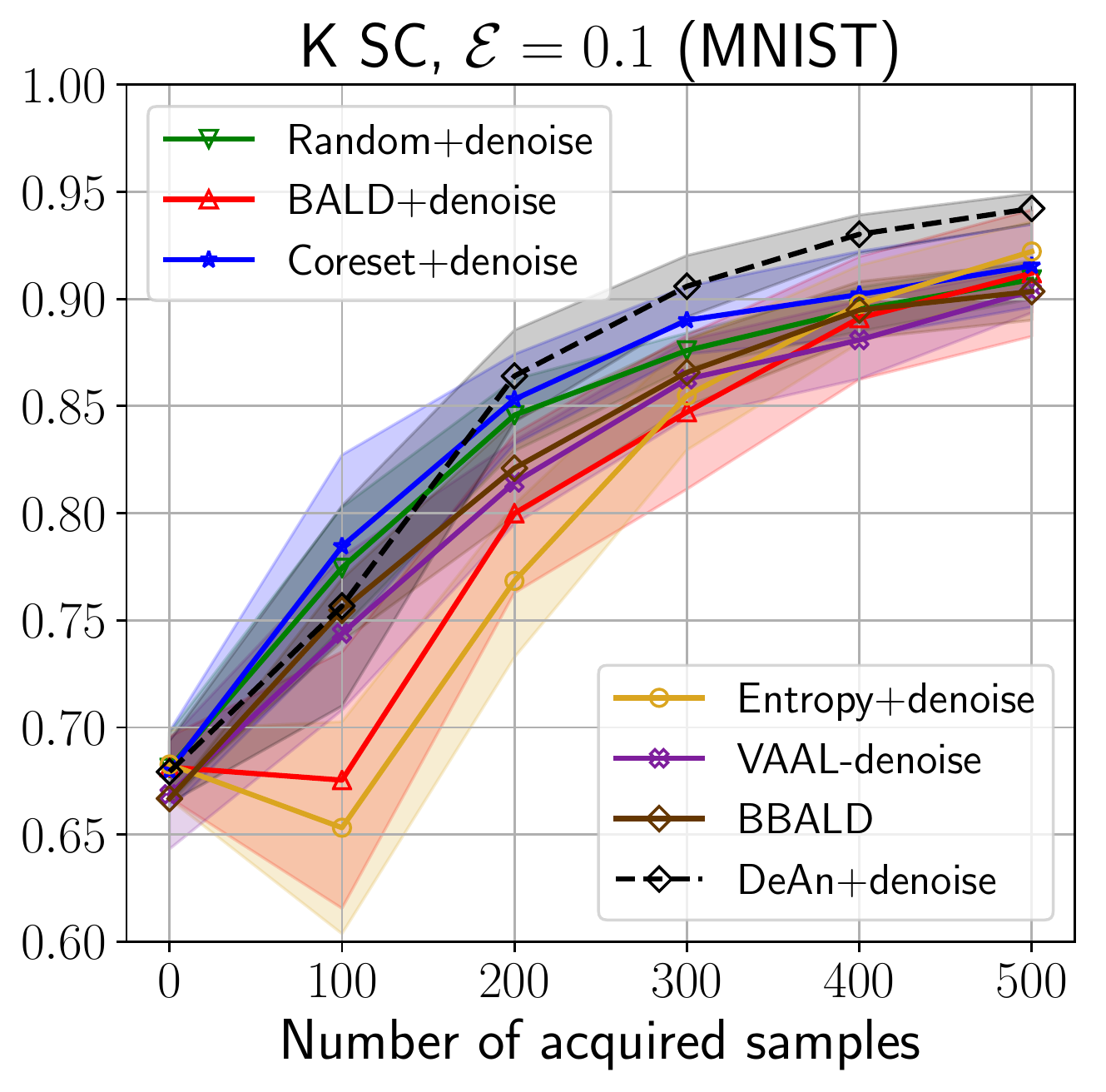}
		\includegraphics*[height = 1.5in]{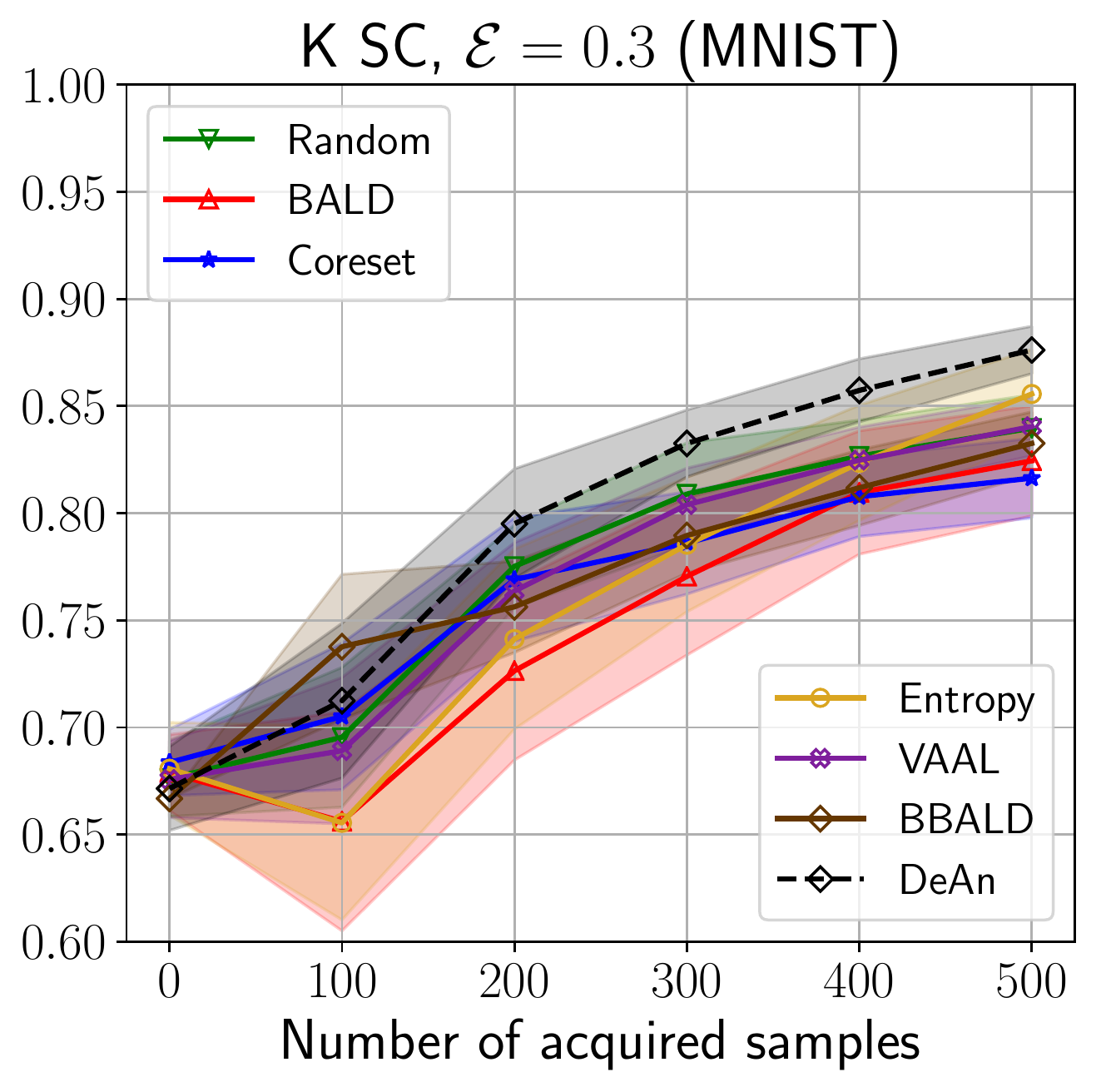}
		\includegraphics*[height = 1.5in]{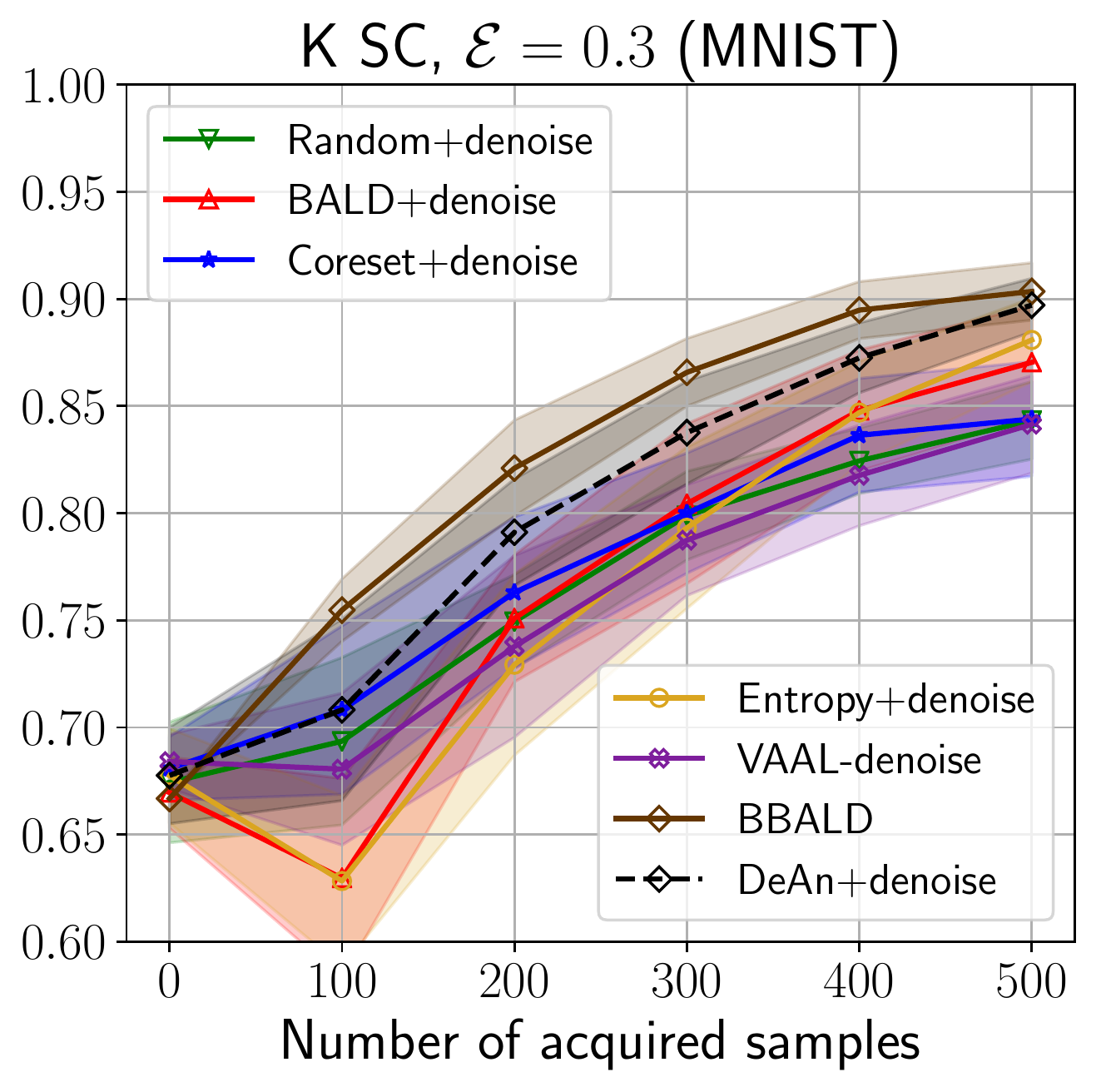}
	\end{subfigure}
	\caption{\textit{Ablation Study with MNIST} using aquisition size of $100$ for two different noise strength, $\varepsilon=0.1, 0.3$. All algorithms show performance improvement in the noisy oracle setup with the proposed denoising layer.}
	\label{fig:ablation}
\end{figure}

\textbf{Denoising} The proposed denoising layer can, in general, be used with any active learning algorithm for combating oracle noise. In Figure\,\ref{fig:ablation} we show that for $\varepsilon = 0.1, 0.3$, the denoising improves performance of all the algorithms in the noisy setup with 50 training epochs for all the cases.

\section{Conclusion}
\label{sec:concl}
In this paper we solve the problem of batch active learning with noisy oracle. We have proposed a batch sample selection mechanism for active learning with access to noisy oracles. We use mutual information as importance score for each sample, and cluster the pool sample space with Jenson-Shannon distance. We point out that active learning algorithms are missing to acknowledge the inaccuracies of the scores in the beginning. Hence, by deterministic annealing, we incorporate model uncertainty/confidence into Gibbs distribution over the clusters and select samples from each cluster with importance sampling. We introduce an additional layer at the output of deep networks to estimate label noise. Experiments on MNIST, SVHN, CIFAR10, and EMNIST show that the proposed method is more robust against noisy labels compared with the state of the art. Even in noise-free scenarios, our method still performs the best/equal than baselines for all four datasets. Our contributions open avenues for exploring applicability of batch active learning in setups involving imperfect data acquisition schemes either by construction or because of resource constraints.

\nocite{GolovinK11, Kay1992, pmlr-v37-wei15, Chen:2013, BUSBY20091183, Martino_IEEE_17}

\clearpage

\bibliographystyle{plainnat}
\bibliography{DeAn}

\begin{thebibliography}{38}
\providecommand{\natexlab}[1]{#1}
\providecommand{\url}[1]{\texttt{#1}}
\expandafter\ifx\csname urlstyle\endcsname\relax
  \providecommand{\doi}[1]{doi: #1}\else
  \providecommand{\doi}{doi: \begingroup \urlstyle{rm}\Url}\fi

\bibitem[Ash et~al.(2020)Ash, Zhang, Krishnamurthy, Langford, and
  Agarwal]{Ash2020Deep}
Jordan~T. Ash, Chicheng Zhang, Akshay Krishnamurthy, John Langford, and Alekh
  Agarwal.
\newblock Deep batch active learning by diverse, uncertain gradient lower
  bounds.
\newblock In \emph{International Conference on Learning Representations}, 2020.
\newblock URL \url{https://openreview.net/forum?id=ryghZJBKPS}.

\bibitem[Beluch et~al.(2018)Beluch, Genewein, Nurnberger, and
  Kohler]{Beluch_CVPR_18}
W.~H. Beluch, T.~Genewein, A.~Nurnberger, and J.~M. Kohler.
\newblock The power of ensembles for active learning in image classification.
\newblock In \emph{CVPR}, pages 9368--9377, jun 2018.

\bibitem[Busby(2009)]{BUSBY20091183}
Daniel Busby.
\newblock Hierarchical adaptive experimental design for gaussian process
  emulators.
\newblock \emph{Reliability Engineering \& System Safety}, 94\penalty0
  (7):\penalty0 1183 -- 1193, 2009.

\bibitem[Chen and Krause(2013)]{Chen:2013}
Yuxin Chen and Andreas Krause.
\newblock Near-optimal batch mode active learning and adaptive submodular
  optimization.
\newblock In \emph{ICML}, volume~28, pages 160--168, 2013.

\bibitem[Chen et~al.(2015{\natexlab{a}})Chen, Hassani, Karbasi, and
  Krause]{Chen15b}
Yuxin Chen, S.~Hamed Hassani, Amin Karbasi, and Andreas Krause.
\newblock Sequential information maximization: When is greedy near-optimal?
\newblock In \emph{CoLT}, pages 338--363, 2015{\natexlab{a}}.

\bibitem[Chen et~al.(2015{\natexlab{b}})Chen, Javdani, Karbasi, Bagnell,
  Srinivasa, and Krause]{Chen:2015:SSV:2888116.2888204}
Yuxin Chen, Shervin Javdani, Amin Karbasi, J.~Andrew Bagnell, Siddhartha
  Srinivasa, and Andreas Krause.
\newblock Submodular surrogates for value of information.
\newblock In \emph{AAAI}, pages 3511--3518, 2015{\natexlab{b}}.

\bibitem[Chen et~al.(2017)Chen, Hassani, Krause, et~al.]{chen2017near}
Yuxin Chen, S~Hamed Hassani, Andreas Krause, et~al.
\newblock Near-optimal bayesian active learning with correlated and noisy
  tests.
\newblock \emph{Electronic Journal of Statistics}, 11\penalty0 (2):\penalty0
  4969--5017, 2017.

\bibitem[Cohen et~al.(2017)Cohen, Afshar, Tapson, and van
  Schaik]{cohen2017emnist}
Gregory Cohen, Saeed Afshar, Jonathan Tapson, and André van Schaik.
\newblock Emnist: an extension of mnist to handwritten letters, 2017.

\bibitem[fchollet(2015)]{fchollet}
fchollet.
\newblock Keras, 2015.
\newblock URL \url{https://github.com/fchollet/keras}.

\bibitem[Gal(2016)]{Gal2016Thesis}
Yarin Gal.
\newblock \emph{Uncertainty in Deep Learning}.
\newblock PhD thesis, University of Cambridge, 2016.

\bibitem[Gal and Ghahramani(2016)]{gal16}
Yarin Gal and Zoubin Ghahramani.
\newblock Dropout as a bayesian approximation: Representing model uncertainty
  in deep learning.
\newblock In \emph{ICML}, volume~48, pages 1050--1059, 2016.

\bibitem[Gal et~al.(2017)Gal, Islam, and Ghahramani]{Gal:2017}
Yarin Gal, Riashat Islam, and Zoubin Ghahramani.
\newblock Deep bayesian active learning with image data.
\newblock In \emph{ICML}, pages 1183--1192, 2017.

\bibitem[Ganti and Gray(2011)]{Ganti2011UPALUP}
Ravi Ganti and Alexander~G. Gray.
\newblock Upal: Unbiased pool based active learning.
\newblock In \emph{AISTATS}, 2011.

\bibitem[Golovin and Krause(2011)]{GolovinK11}
Daniel Golovin and Andreas Krause.
\newblock Adaptive submodularity: Theory and applications in active learning
  and stochastic optimization.
\newblock \emph{J. Artif. Intell. Res.}, 42:\penalty0 427--486, 2011.

\bibitem[Houlsby and Ghahramani(2011)]{Houlsby11bayesianactive}
Neil Houlsby and Zoubin Ghahramani.
\newblock Bayesian active learning for classification and preference learning.
  arxiv:1112.5745, 2011.

\bibitem[Jindal et~al.(2019)Jindal, Nokleby, and Pressel]{Jindal2019ANN}
Ishan Jindal, Matthew~S. Nokleby, and Daniel Pressel.
\newblock A nonlinear, noise-aware, quasi-clustering approach to learning deep
  cnns from noisy labels.
\newblock In \emph{CVPR 2019}, 2019.

\bibitem[Kendall and Gal(2017)]{Kendall:2017}
Alex Kendall and Yarin Gal.
\newblock What uncertainties do we need in bayesian deep learning for computer
  vision?
\newblock In \emph{Proceedings of the 31st International Conference on Neural
  Information Processing Systems}, pages 5580--5590, 2017.

\bibitem[Khetan et~al.(2018)Khetan, Lipton, and Anandkumar]{khetan2018learning}
Ashish Khetan, Zachary~C. Lipton, and Anima Anandkumar.
\newblock Learning from noisy singly-labeled data.
\newblock In \emph{International Conference on Learning Representations}, 2018.
\newblock URL \url{https://openreview.net/forum?id=H1sUHgb0Z}.

\bibitem[Kirsch et~al.(2019)Kirsch, van Amersfoort, and
  Gal]{kirsch2019batchbald}
Andreas Kirsch, Joost van Amersfoort, and Yarin Gal.
\newblock Batchbald: Efficient and diverse batch acquisition for deep bayesian
  active learning, 2019.

\bibitem[Krizhevsky(2009)]{Krizhevsky09learningmultiple}
Alex Krizhevsky.
\newblock Learning multiple layers of features from tiny images.
\newblock Technical report, 2009.

\bibitem[Lakshminarayanan et~al.(2016)Lakshminarayanan, Pritzel, and
  Blundell]{lakshminarayanan2016simple}
Balaji Lakshminarayanan, Alexander Pritzel, and Charles Blundell.
\newblock Simple and scalable predictive uncertainty estimation using deep
  ensembles, 2016.

\bibitem[Lecun et~al.(1998)Lecun, Bottou, Bengio, and
  Haffner]{Lecun98gradient-basedlearning}
Yann Lecun, Léon Bottou, Yoshua Bengio, and Patrick Haffner.
\newblock Gradient-based learning applied to document recognition.
\newblock In \emph{Proceedings of the IEEE}, pages 2278--2324, 1998.

\bibitem[MacKay(1992)]{Kay1992}
David J.~C. MacKay.
\newblock Information-based objective functions for active data selection.
\newblock \emph{Neural Computation}, 4\penalty0 (4):\penalty0 590--604, 1992.

\bibitem[{Martino} et~al.(2017){Martino}, {Vicent}, and
  {Camps-Valls}]{Martino_IEEE_17}
L.~{Martino}, J.~{Vicent}, and G.~{Camps-Valls}.
\newblock Automatic emulator and optimized look-up table generation for
  radiative transfer models.
\newblock In \emph{2017 IEEE International Geoscience and Remote Sensing
  Symposium (IGARSS)}, pages 1457--1460, July 2017.

\bibitem[{Naghshvar} et~al.(2012){Naghshvar}, {Javidi}, and
  {Chaudhuri}]{NaghshvarNoisy}
M.~{Naghshvar}, T.~{Javidi}, and K.~{Chaudhuri}.
\newblock Noisy bayesian active learning.
\newblock In \emph{2012 50th Annual Allerton Conference on Communication,
  Control, and Computing (Allerton)}, pages 1626--1633, Oct 2012.

\bibitem[Netzer et~al.(2011)Netzer, Wang, Coates, Bissacco, Wu, and
  Ng]{Netzer2011ReadingDI}
Yuval Netzer, Tiejie Wang, Adam Coates, Alessandro Bissacco, Baolin Wu, and
  Andrew~Y. Ng.
\newblock Reading digits in natural images with unsupervised feature learning.
\newblock In \emph{NIPS Workshop on Deep Learning and Unsupervised Feature
  Learning}, 2011.

\bibitem[Pedregosa et~al.(2011)Pedregosa, Varoquaux, Gramfort,
  et~al.]{scikit-learn}
F.~Pedregosa, G.~Varoquaux, A.~Gramfort, et~al.
\newblock Scikit-learn: Machine learning in {P}ython.
\newblock \emph{Journal of Machine Learning Research}, 12:\penalty0 2825--2830,
  2011.

\bibitem[Piczak(2015)]{ESC50}
Karol~J. Piczak.
\newblock {ESC}: {Dataset} for {Environmental Sound Classification}.
\newblock In \emph{Proceedings of the 23rd {Annual ACM Conference} on
  {Multimedia}}, pages 1015--1018. {ACM Press}, 2015.
\newblock ISBN 978-1-4503-3459-4.
\newblock \doi{10.1145/2733373.2806390}.
\newblock URL \url{http://dl.acm.org/citation.cfm?doid=2733373.2806390}.

\bibitem[Rokach and Maimon(2005)]{Rokach2005}
Lior Rokach and Oded Maimon.
\newblock \emph{Clustering Methods}, pages 321--352.
\newblock Springer US, Boston, MA, 2005.

\bibitem[Rose et~al.(1990)Rose, Gurewitz, and Fox]{Rose_PhysRevLett.65.945}
Kenneth Rose, Eitan Gurewitz, and Geoffrey~C. Fox.
\newblock Statistical mechanics and phase transitions in clustering.
\newblock \emph{Phys. Rev. Lett.}, 65:\penalty0 945--948, Aug 1990.

\bibitem[Russakovsky et~al.(2015)Russakovsky, Deng, Su, et~al.]{Imagenet_15}
Olga Russakovsky, Jia Deng, Hao Su, et~al.
\newblock Imagenet large scale visual recognition challenge.
\newblock \emph{International Journal of Computer Vision}, 115\penalty0
  (3):\penalty0 211--252, Dec 2015.
\newblock ISSN 1573-1405.
\newblock \doi{10.1007/s11263-015-0816-y}.
\newblock URL \url{https://doi.org/10.1007/s11263-015-0816-y}.

\bibitem[Sener and Savarese(2018)]{sener2018active}
Ozan Sener and Silvio Savarese.
\newblock Active learning for convolutional neural networks: A core-set
  approach.
\newblock In \emph{International Conference on Learning Representations}, 2018.
\newblock URL \url{https://openreview.net/forum?id=H1aIuk-RW}.

\bibitem[Settles(2009)]{settles.tr09}
Burr Settles.
\newblock Active learning literature survey.
\newblock Computer Sciences Technical Report 1648, University of
  Wisconsin--Madison, 2009.

\bibitem[Sinha et~al.(2019)Sinha, Ebrahimi, and Darrell]{Samarth_ICCV_19}
Samarth Sinha, Sayna Ebrahimi, and Trevor Darrell.
\newblock Variational adversarial active learning. arxiv preprint
  arxiv:1904.00370, 2019.

\bibitem[Teye et~al.(2018)Teye, Azizpour, and Smith]{Teye2018BayesianUE}
Mattias Teye, Hossein Azizpour, and Kevin Smith.
\newblock Bayesian uncertainty estimation for batch normalized deep networks.
\newblock In \emph{ICML}, 2018.

\bibitem[Tong(2001)]{Simon2001PhD}
Simon Tong.
\newblock \emph{Active Learning: Theory and Applications}.
\newblock PhD thesis, Stanford University, 2001.

\bibitem[{Wang} and {Shang}(2014)]{Wang2014AL}
D.~{Wang} and Y.~{Shang}.
\newblock A new active labeling method for deep learning.
\newblock In \emph{2014 International Joint Conference on Neural Networks
  (IJCNN)}, pages 112--119, July 2014.

\bibitem[Wei et~al.(2015)Wei, Iyer, and Bilmes]{pmlr-v37-wei15}
Kai Wei, Rishabh Iyer, and Jeff Bilmes.
\newblock Submodularity in data subset selection and active learning.
\newblock In \emph{ICML}, volume~37, pages 1954--1963, 2015.

\end{thebibliography}

\appendix


\section{ESC50 Crowd Labeling Experiment}
\label{sec:esc50_exp}
We selected 10 categories of ESC50 and use Amazon Mechanical Turk for annotation. In each annotation task, the crowd worker is asked to listen to the sound track and pick the class that the sound belongs to, with confidence level. The crowd worker can also pick ``Unsure" if he/she does not think the sound track clearly belongs to one of the 10 categories. For quality control, we embed sound tracks that clearly belong to one class (these are called gold standards) into the set of tasks an annotator will do. If the annotator labels the gold standard sound tracks wrong, then labels from this annotator will be discarded.

\begin{figure}
    \centering
    \includegraphics[height = 2.0in]{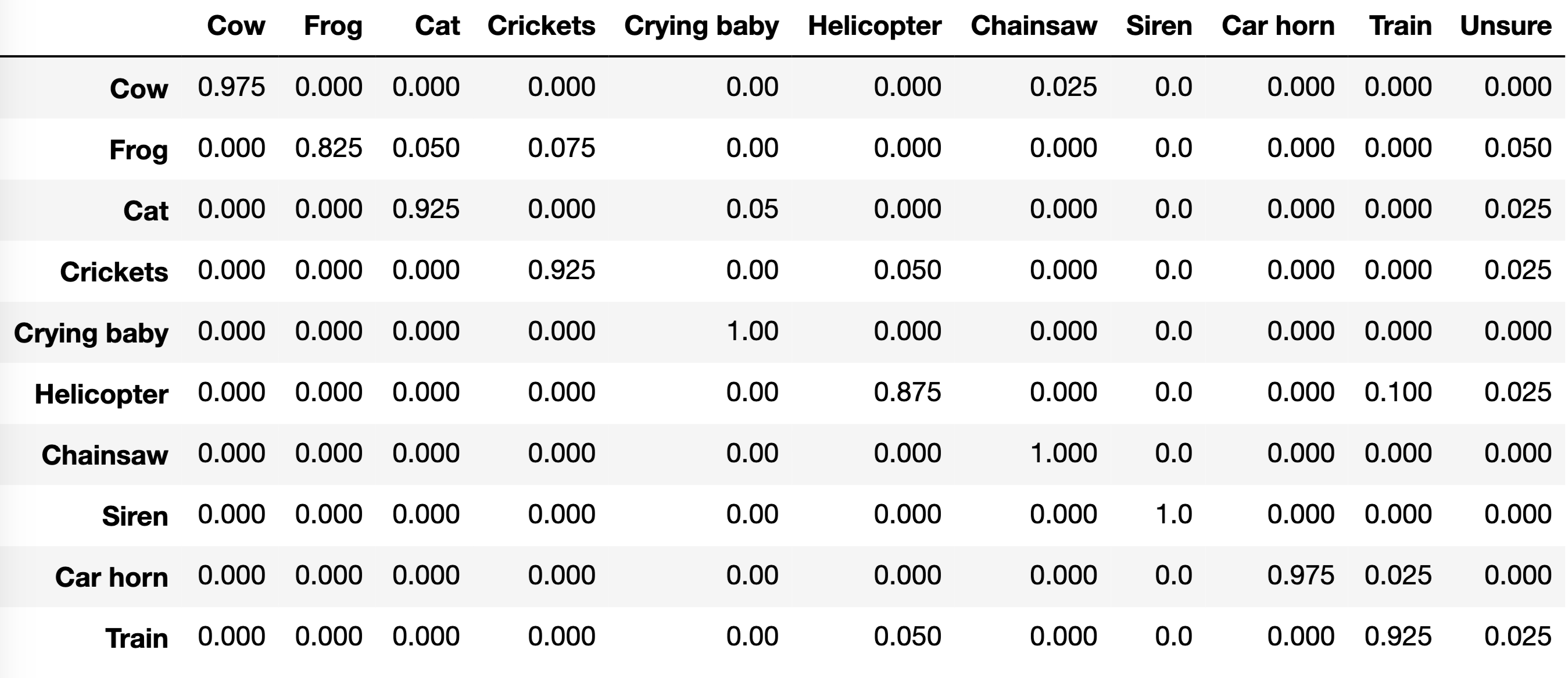}
    \caption{Annotation confusion matrix of 10 classes of ESC50 \label{fig:ESC50}}
\end{figure}

The confusion table of this crowd labeling experiment is shown in \figurename\,\ref{fig:ESC50}: each row corresponds to sound tracks with one ground truth class, and the columns are majority-voted crowd-sourced labels of the sound tracks. We can see that for some classes, such as frog and helicopter, even with 5 crowd workers, the majority vote of their annotation still cannot fully agree with the ground truth class.
\label{appdx:noisy_oracle}

\section{Experiments Details}
\label{sec:exp_det}
\subsection{Models}
\paragraph{MNIST} The model from \citep{fchollet} has one block of Convolution, Convolution, Dropout, MaxPool with 32, 32 4x4 filters. The dropout probability in this block is set to be 0.25. This block is followed by two Dense layers of 128, 10 units with a Dropout layer of probability 0.5 in between them.
\paragraph{EMNIST} The model for EMNIST extends the MNIST model. The first block has Convolution, Convolution, Dropout, and Maxpool with 32, 64 4x4 filters. This is followed by Convolution, Dropout, and Maxpool with 128 4x4 filters. The Dropout probabilities till here are 0.25. Next, two Dense layers follows of 512, 47 hidden units. A Dropout layer of 0.5 probability exists between these two Dense layers.
\paragraph{CIFAR, SVHN} The model has three blocks of [Convolution, Convolution, Dropout, Maxpool] of [32, 32], [64, 64], and [128, 128] 3x3 filters. The Dropout probability is set to 0.25. The convolution blocks are followed by two Dense layers with (i) CIFAR10, SVHN having 128, 10, and (ii) CIFAR100 having 512, 100 hidden units. A Dropout layer of probability 0.5 is put between Dense layers. We note that using this simple model, in particular, for CIFAR10 has $88\%$ accuracy which is only $3\%$ behind using much deeper models like VGG16. Such simplicity is useful in performing active learning experiments over larger number of runs and is sufficient for demonstrating the proof of concept.

\subsection{Hyperparameters}
We use $f(\sigma) = (e^{1/\sigma}-1)/4$ in all the experiments for MNIST, CIFAR10, and SVHN. For EMNIST, CIFAR100 we use $f(\sigma) = (4^{1/\sigma}-1)/4$.

\section{Model training with Noisy Oracle}
\label{sec:denoise_demo}
\begin{figure}
	\centering
	\begin{subfigure}[t]{0.5\linewidth}
	\centering
	\begin{tikzpicture}
	\node[anchor=north west,inner sep=0] at (0,0) {\includegraphics*[trim =5.5cm 8.4cm 11cm 5cm, clip, height = 1.5in]{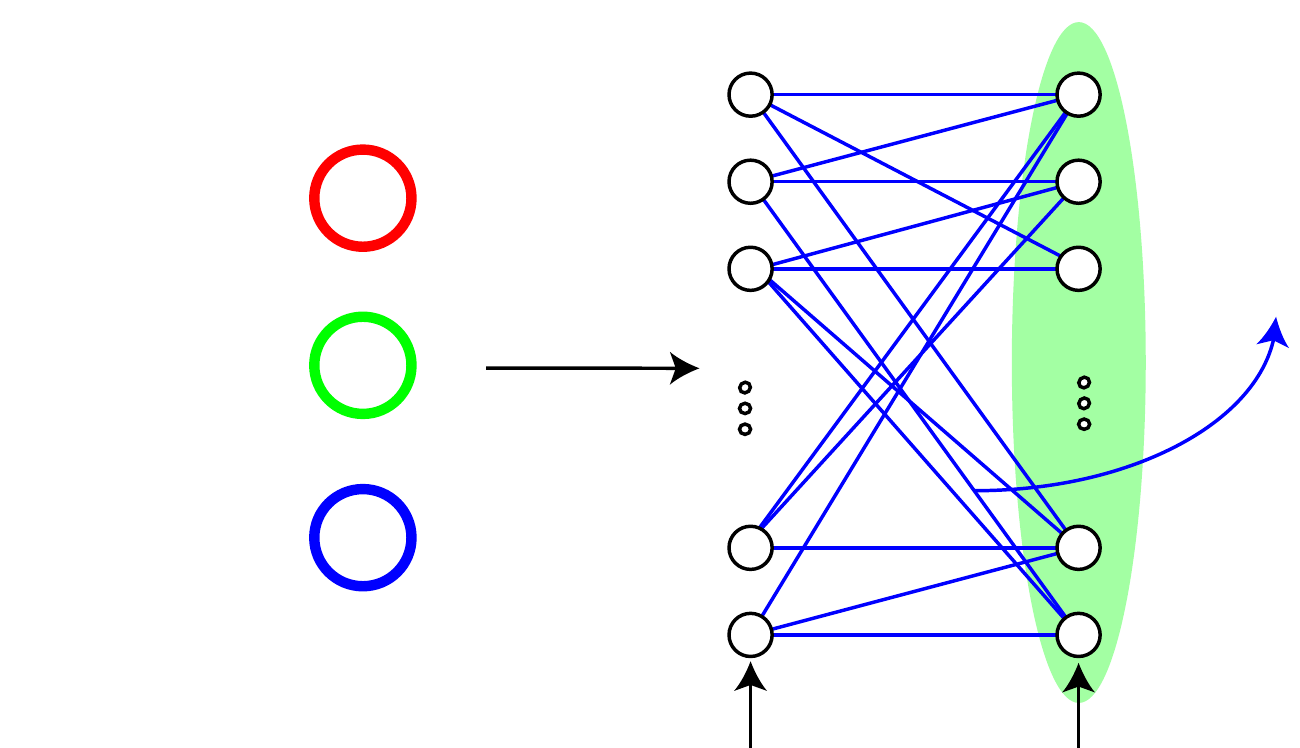}};
	\node[anchor=north] at (1.9,-3.4) {${\bf x}$};
	\node[anchor=north] at (3.9,-3.8) {validate};
	\node[anchor=north] at (5.5,-3.8) {train};
	\node[anchor=north] at (3,-1.3) {$h_{\theta}({\bf x})$};
	\node[anchor=north] at (3.1,-0.1) {$p(y = j)$};
	\node[anchor=north] at (6.4,-0.2) {$p(y^{\prime} = i)$};
	\node[anchor=north] at (7,-1.1) {$p(y^{\prime} = i\vert y = j)$};
	\node[anchor=north] at (2.7,-4.1) {$\underbrace{\hphantom{GauravGuptaGa}}$};
	\node[anchor=north] at (1,-4) {$\mathcal{M}$};
	\node[anchor=north] at (3.6,0.1) {$\overbrace{\hphantom{GauravGuptaGauravGupta}}$};
	\node[anchor=north] at (1,0) {$\mathcal{M}^{\ast}$};
	\end{tikzpicture}
	\caption{denoising layer demonstration}
	\label{sfig:noiseLayer}
	\end{subfigure}
	\hfill
	\begin{subfigure}[t]{0.49\textwidth}
	\centering
	\begin{tikzpicture}
	\node[anchor=north west,inner sep=0] at (0,1.7) {\includegraphics*[trim =12.5cm 9cm 11cm 5cm, clip, height = 1.5in]{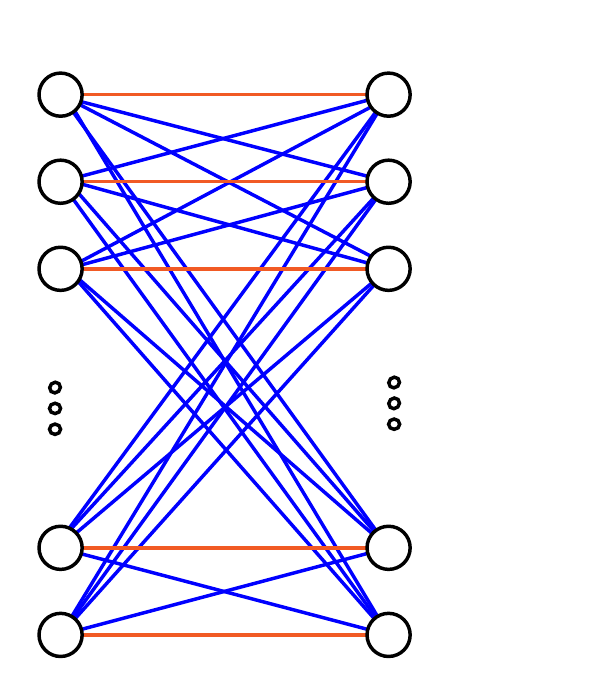}};
	\node[anchor=north] at (1.3,1.7) {{\color{redBrown}$1-\varepsilon$}};
	\node[anchor=north, fill=white] at (1.25,-0.3) {{\color{blue}$\varepsilon/(K-1)$}};
	\node[anchor=north, fill=white] at (1.25,-2.5){};
	\end{tikzpicture}
	\caption{$K$-SC}
	\label{sfig:KSC}
	\end{subfigure}
	\caption{Demonstration of appending the denoising layer to the model $\mathcal{M}$ for getting $\mathcal{M}^{\ast}$ in (a), and $K$-SC channel with probability of error $\varepsilon$ in (b).}
	\label{fig:noiseDemo}
\end{figure}
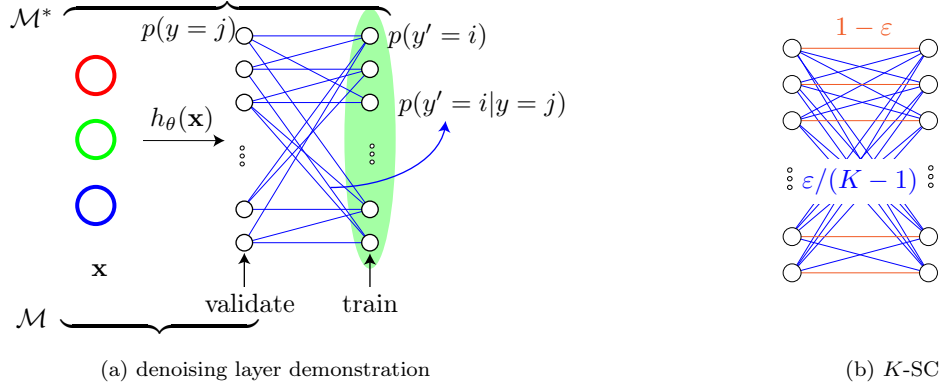

A pictorial description of using the denoising layer along with the existing model is shown in Figure\,\ref{fig:noiseDemo}. The data with labels output from Noisy Oracle is used to train the appended model $\mathcal{M}^{\ast}$. After training with the noisy data, the required model $\mathcal{M}$ is detached from the appended model.

\section{More Experiments}
\label{sec:more_experi}
We present rest of the experimental results supplementary to the ones presented in the main body of Section\,5. 
\subsection{MNIST}
\begin{figure}[ht]
	\centering
	\begin{subfigure}[b]{\linewidth}
	\centering
		\includegraphics*[height = 2.5in]{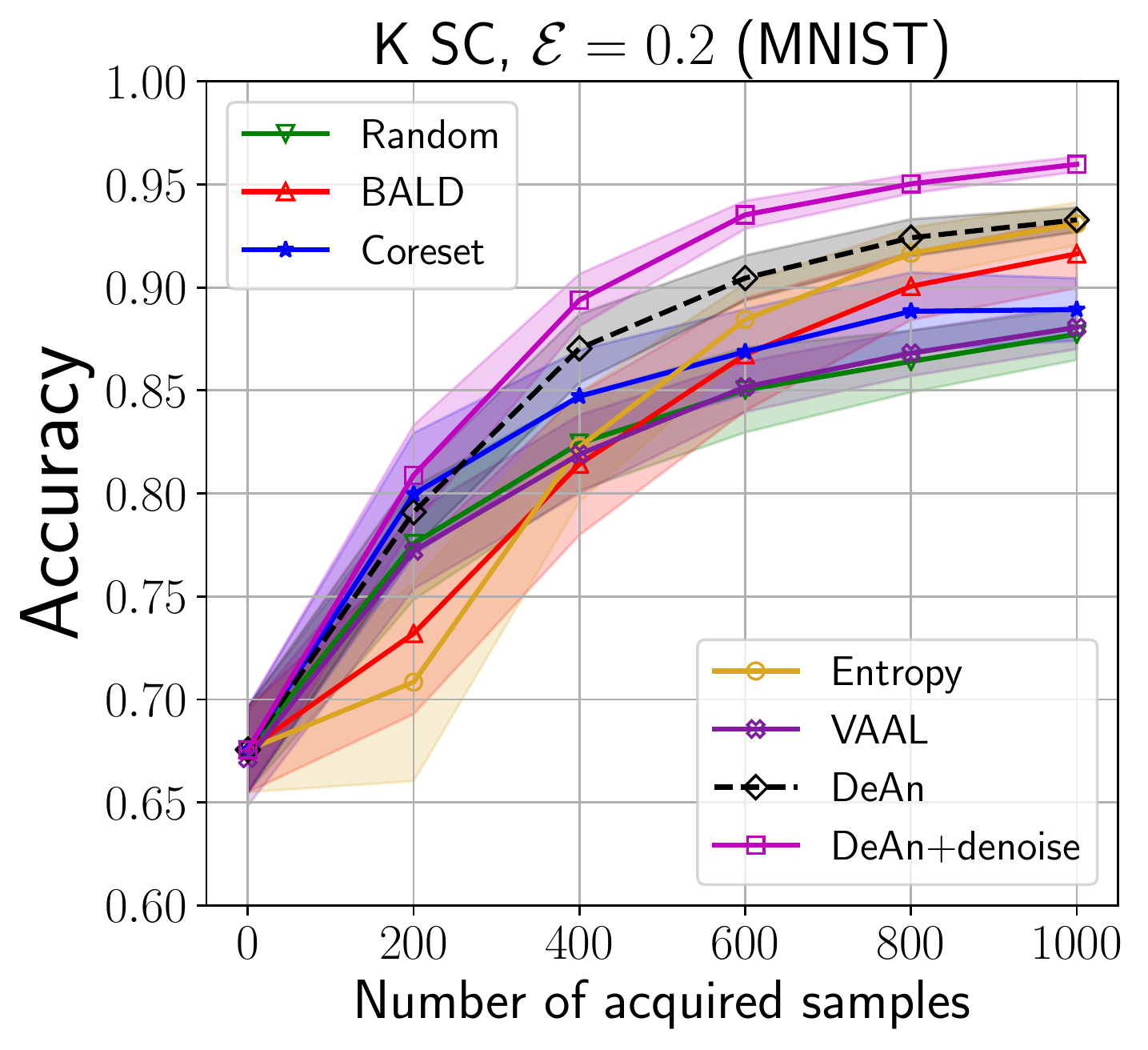}
		\includegraphics*[height = 2.5in]{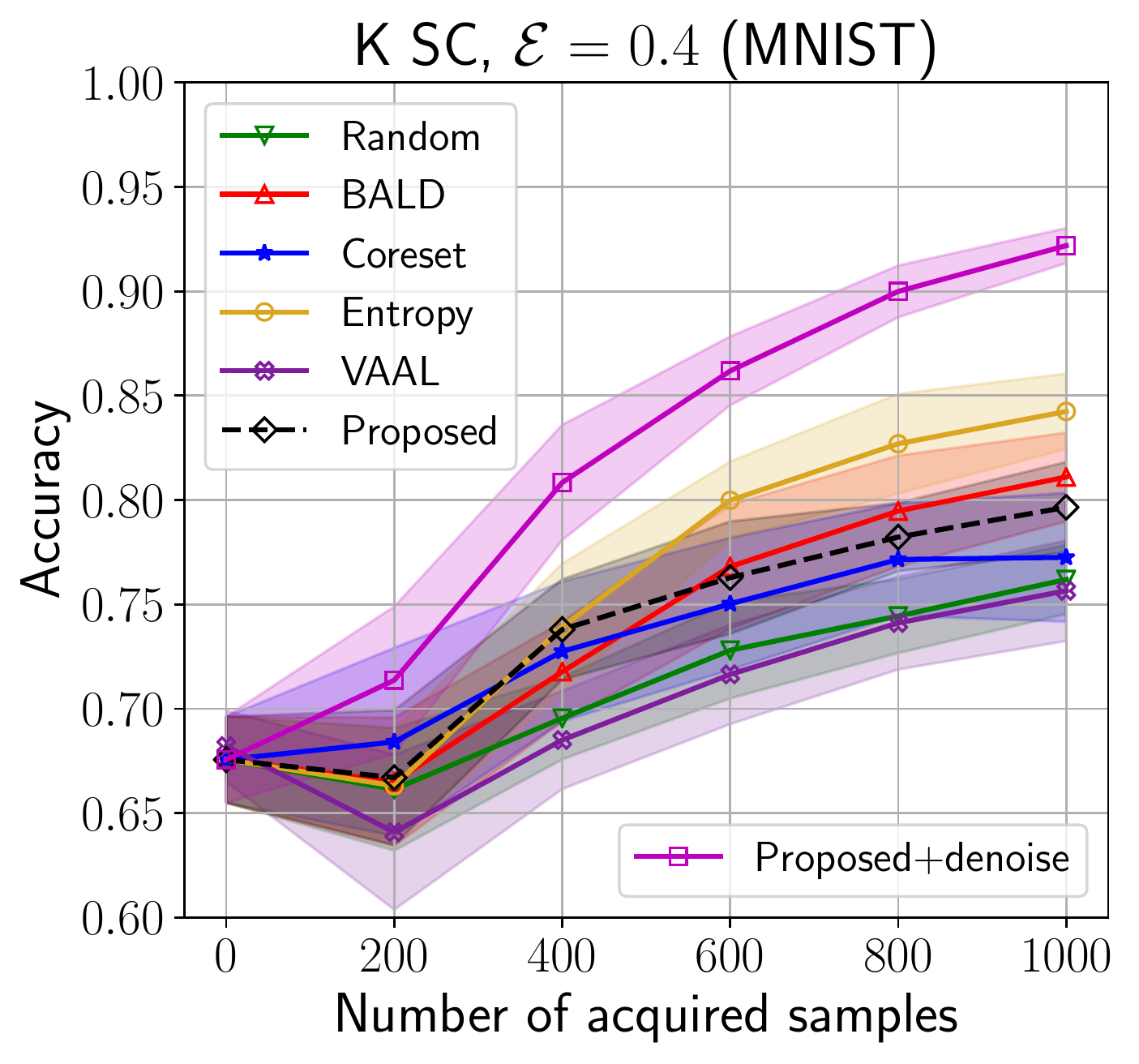}
	\end{subfigure}
	\caption{Active learning results for various algorithms under oracle noise strength $\varepsilon = 0.2, 0.4$ for MNIST Image dataset.}
	\label{fig:mnist_onp_0d2_0d4}
\end{figure}

\begin{figure}[ht]
	\centering
	\begin{subfigure}[b]{\linewidth}
	\centering
	\includegraphics*[height = 2.5in]{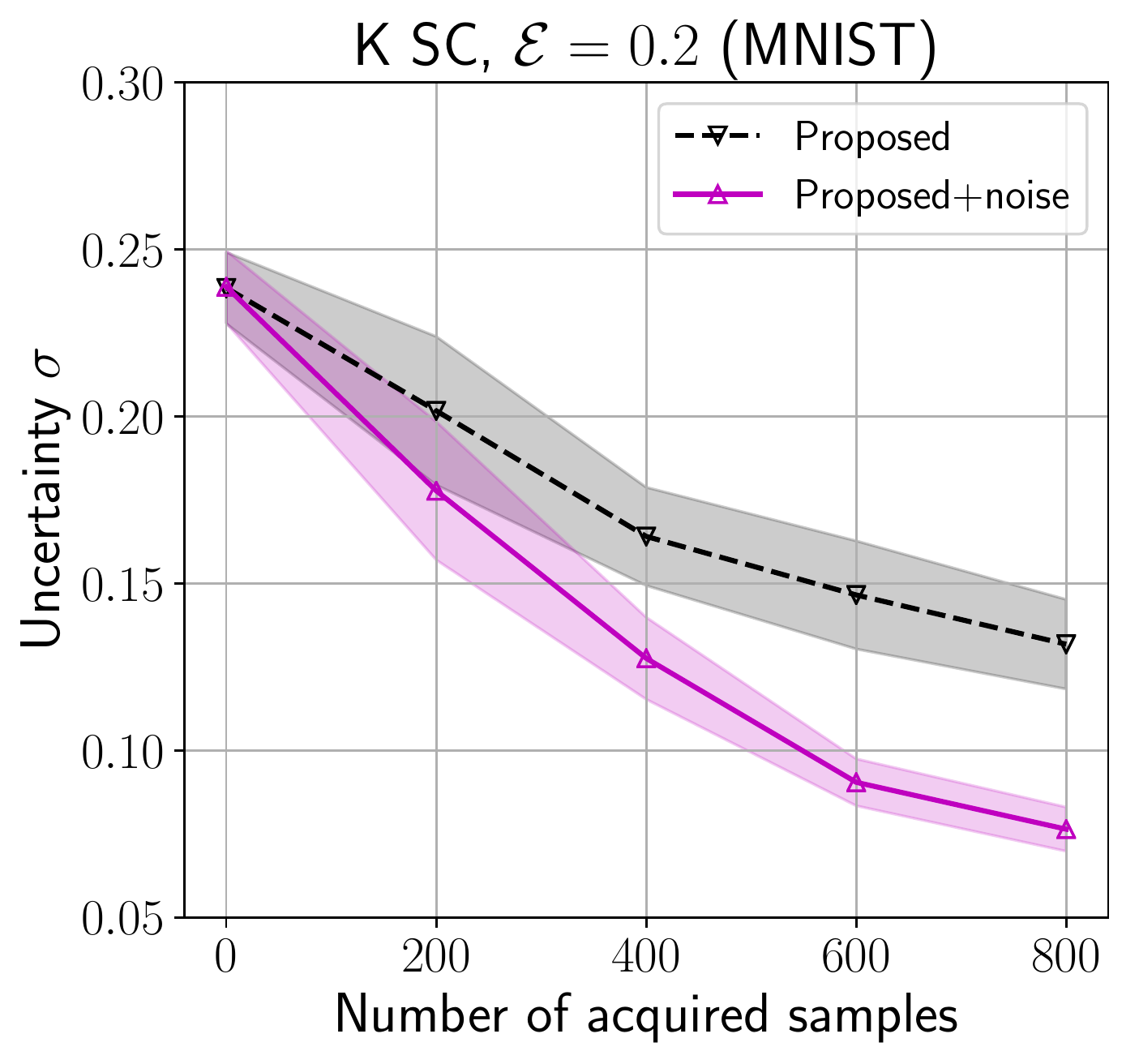}
	\hspace*{-5pt}
	\includegraphics*[height = 2.5in]{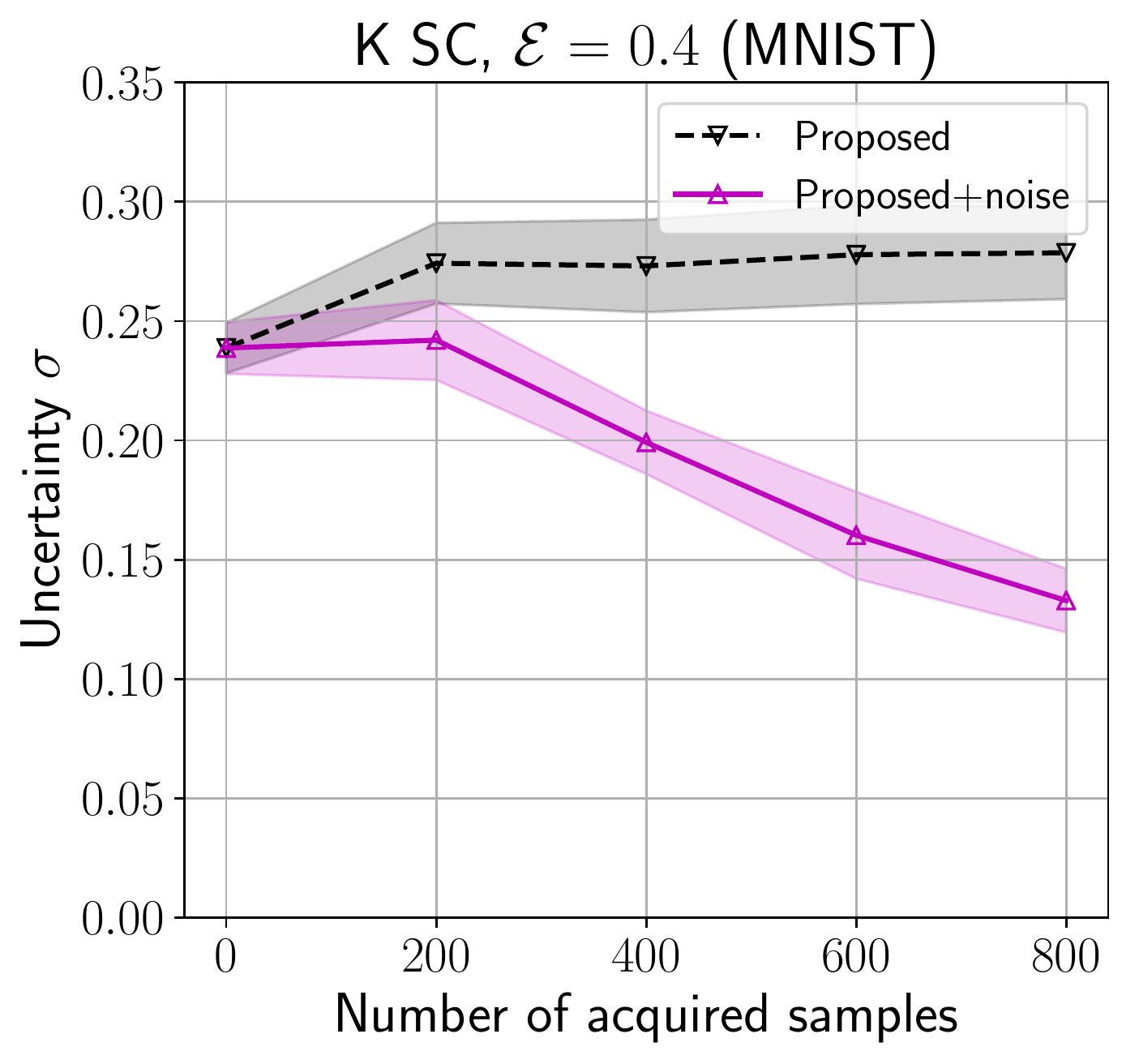}
	\end{subfigure}
\caption{Uncertainty $\sigma$ across active learning experiment for $K$-SC ($\varepsilon=0.2, 0.4$) on MNIST dataset.}
\label{fig:uncertainty_MNIST_0d2_0d4}
\end{figure}

The active learning algorithm performance for oracle noise strength of $\varepsilon = 0.2$ and $\varepsilon = 0.4$ are presented in \figurename\,\ref{fig:mnist_onp_0d2_0d4}. Similarly to what discussed in Section\,5, we observe that the performance of proposed algorithm dominates all other existing works for $\varepsilon = 0.2$. We witnessed that the proposed algorithm performance (without denoising layer) is not able to match other algorithms (BALD and Entropy) when $\varepsilon = 0.4$, even with more training data. The reason for this behavior can be explained using the uncertainty measure $\sigma$ output in the  \figurename\,\ref{fig:uncertainty_MNIST_0d2_0d4}. We see that under strong noise influence from the oracle, the model uncertainty doesn't reduce along the active learning acquisition iterations. Because of this behavior, the proposed uncertainty based algorithm sticks to put more weightage on uniform random sampling, even with more training data. However, we see that using denoising layer, we have better model uncertainty estimates under the influence of noisy oracle. Since the uncertainty estimates improve, as we see in \figurename\,\ref{fig:uncertainty_MNIST_0d2_0d4}, for $\varepsilon = 0.4$, the proposed algorithm along with the denoising layer performs very well and has significant improvement in performance as compared to other approaches.
\subsection{CIFAR10}
\begin{figure}[ht]
	\centering
	\begin{subfigure}[b]{\linewidth}
	\centering
	\includegraphics*[height = 2.4in]{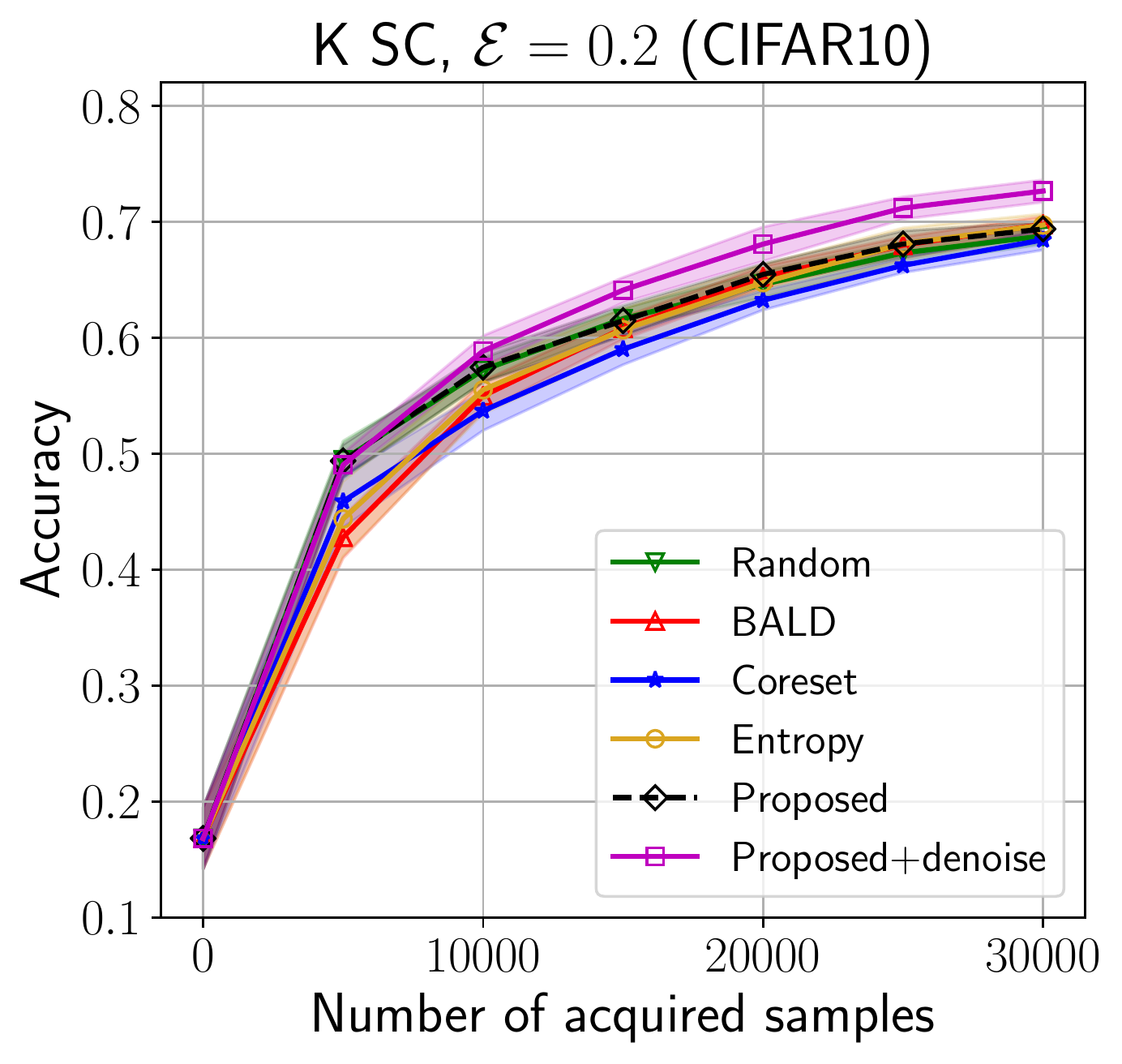}
	\includegraphics*[height = 2.4in]{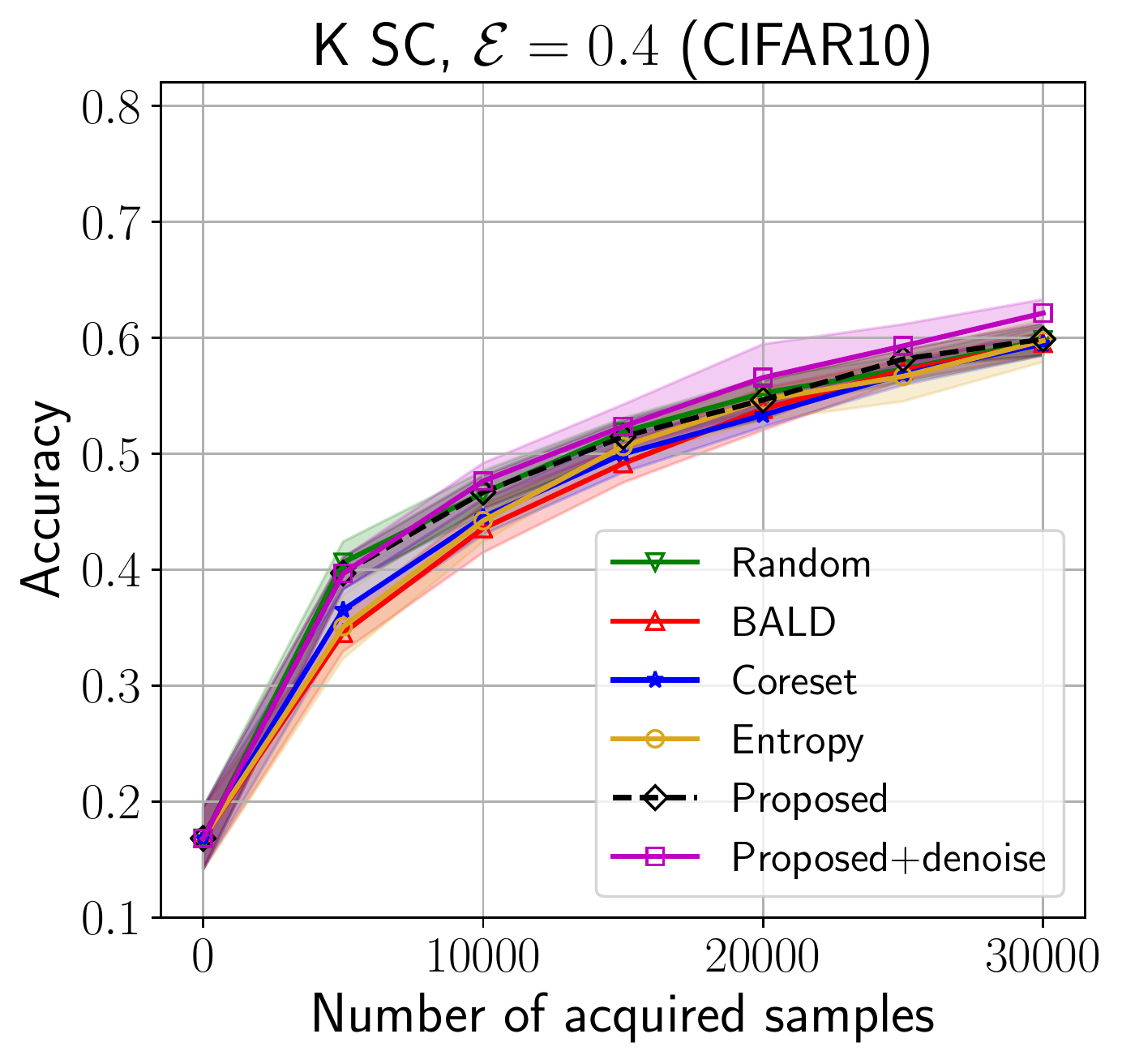}
	\end{subfigure}
	\caption{Active learning results for various algorithms under oracle noise strength $\varepsilon = 0.2, 0.4$ for CIFAR10 Image dataset.}
	\label{fig:cifar10_onp_0d2_0d4}
\end{figure}

The results for CIFAR10 dataset with oracle noise strength of $\varepsilon = 0.2$ and $0.4$ are provided in the \figurename\,\ref{fig:cifar10_onp_0d2_0d4}. We see that the proposed algorithm without/with using the denoising layer outperforms other benchmarks.
\subsection{SVHN}
\begin{figure}[ht]
	\centering
	\begin{subfigure}[b]{\linewidth}
	\centering
		\includegraphics*[height = 2.4in]{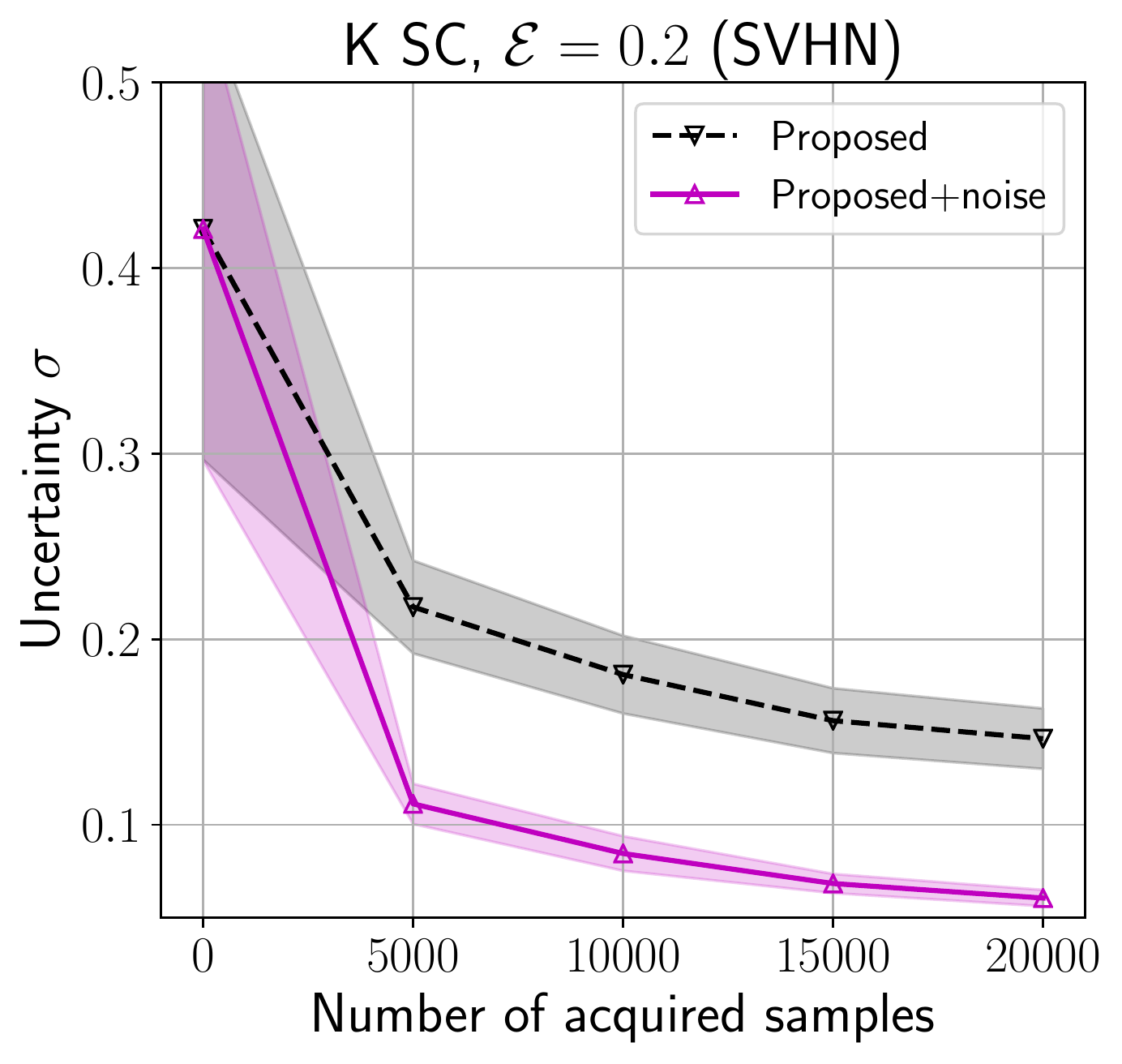}
		\includegraphics*[height = 2.4in]{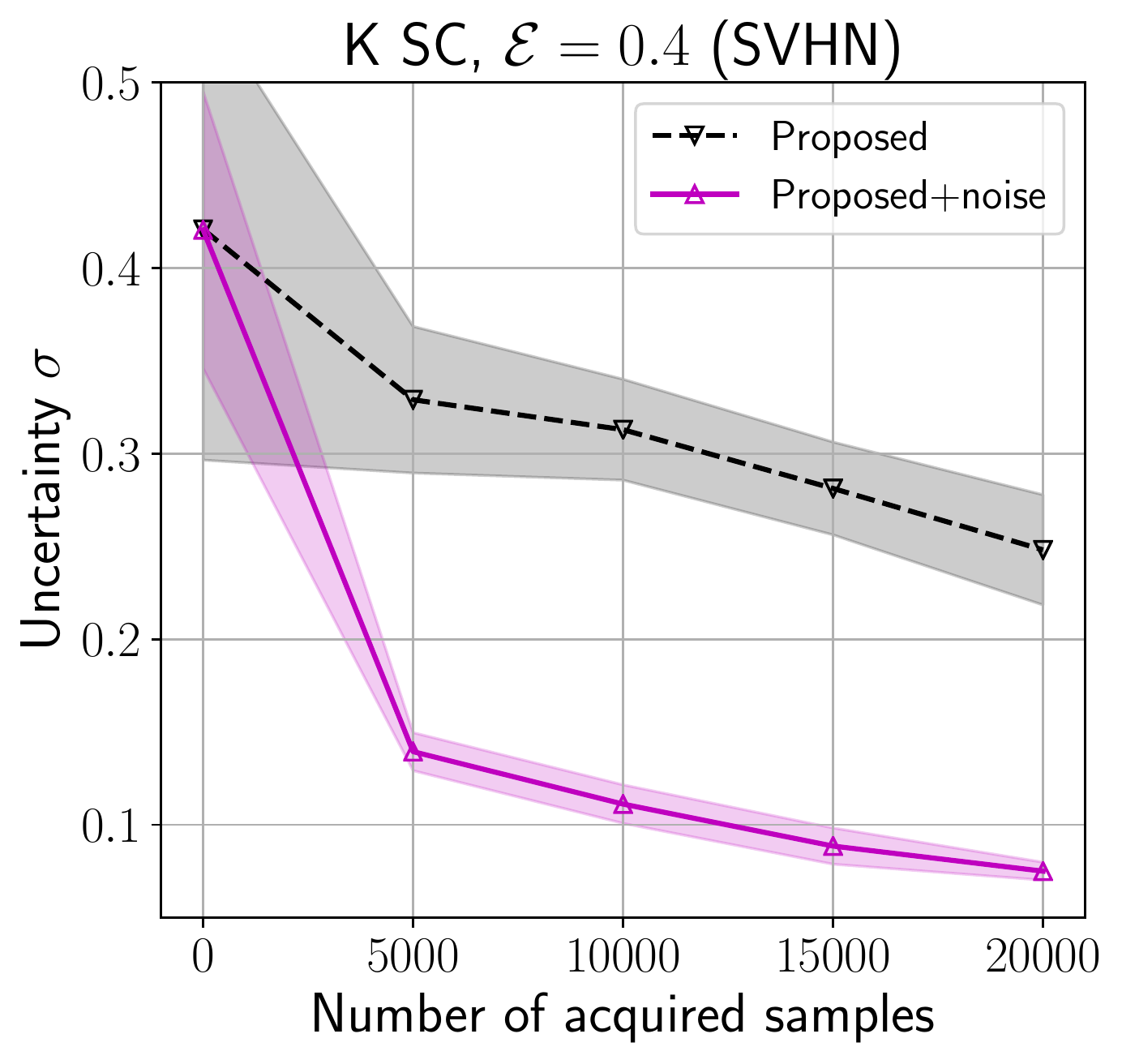}
	\end{subfigure}
\caption{Uncertainty $\sigma$ across active learning experiment for $K$-SC ($\varepsilon=0.2, 0.4$) on SVHN dataset.}
\label{fig:uncertainty_SVHN_0d2_0d4}
\end{figure}

We provide the active learning accuracy results for SVHN dataset with oracle noise strength of $\varepsilon = 0.2$ and $0.4$ in the \figurename\,\ref{fig:cifar10_onp_0d2_0d4}. Similar to other results, we see that the proposed algorithm without/with using the denoising layer outperforms other benchmarks for $\varepsilon = 0.2$. For oracle noise strength of $\varepsilon = 0.4$, we see a similar trend as MNIST regarding performance compromise to the proposed uncertainty based batch selection. The reason is again found in the uncertainty estimates plot in \figurename\,\ref{fig:uncertainty_SVHN_0d2_0d4} for $\varepsilon = 0.4$. With more mislabeled training examples, the model uncertainty estimate doesn't improve with active learning samples acquisition. Hence, the proposed algorithm makes the judgment of staying close to uniform random sampling. However, unlike MNIST in \figurename\,\ref{fig:uncertainty_MNIST_0d2_0d4}, the uncertainty estimate is not that poor for SVHN, i.e., it still decays. Therefore, the performance loss in proposed algorithm is not that significant. While, upon using the denoising layer, the uncertainty estimates improve significantly, and therefore, the proposed algorithm along with the denoising layer outperforms other approaches by big margin.
\begin{figure}[ht]
	\centering
	\begin{subfigure}[b]{\linewidth}
	\centering
		\includegraphics*[height = 2.4in]{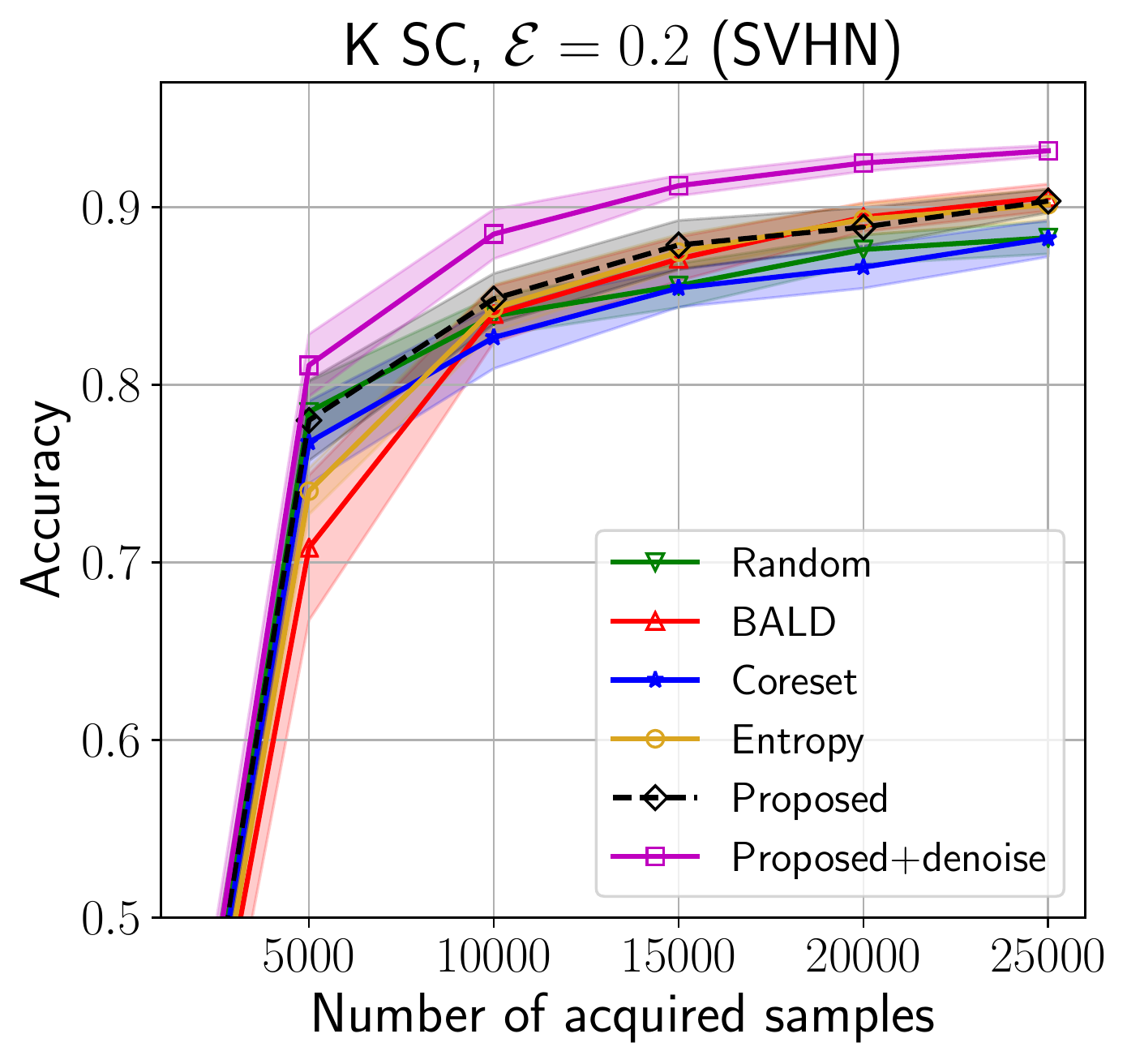}
		\includegraphics*[height = 2.4in]{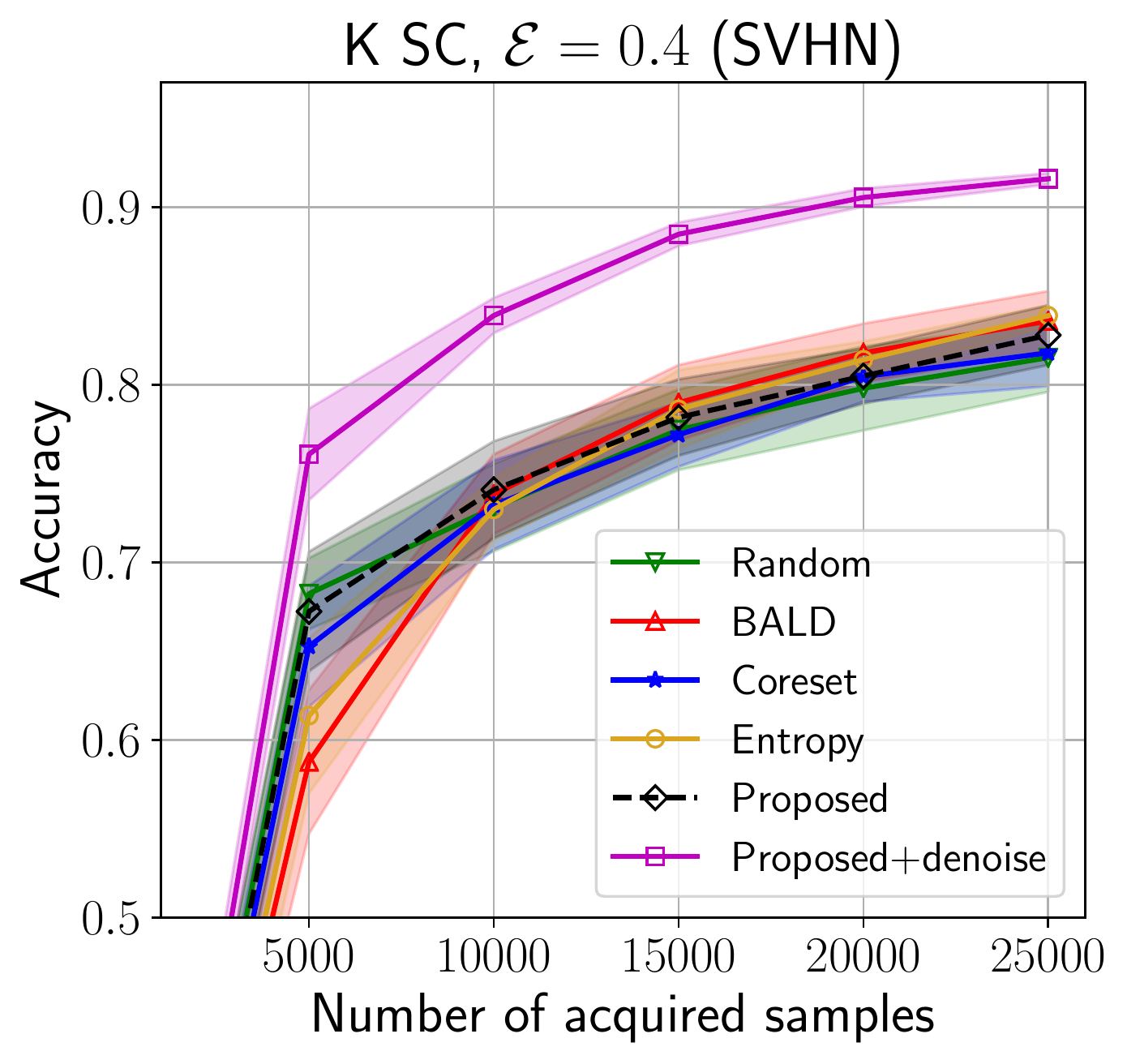}
	\end{subfigure}
	\caption{Active learning results for various algorithms under oracle noise strength $\varepsilon = 0.2, 0.4$ for SVHN Image dataset.}
\label{fig:svhn_onp_0d2_0d4}
\end{figure}
\subsection{EMNIST}
\begin{figure}[ht]
	\centering
	\begin{subfigure}[b]{\linewidth}
	\centering
		\includegraphics*[height = 2.4in]{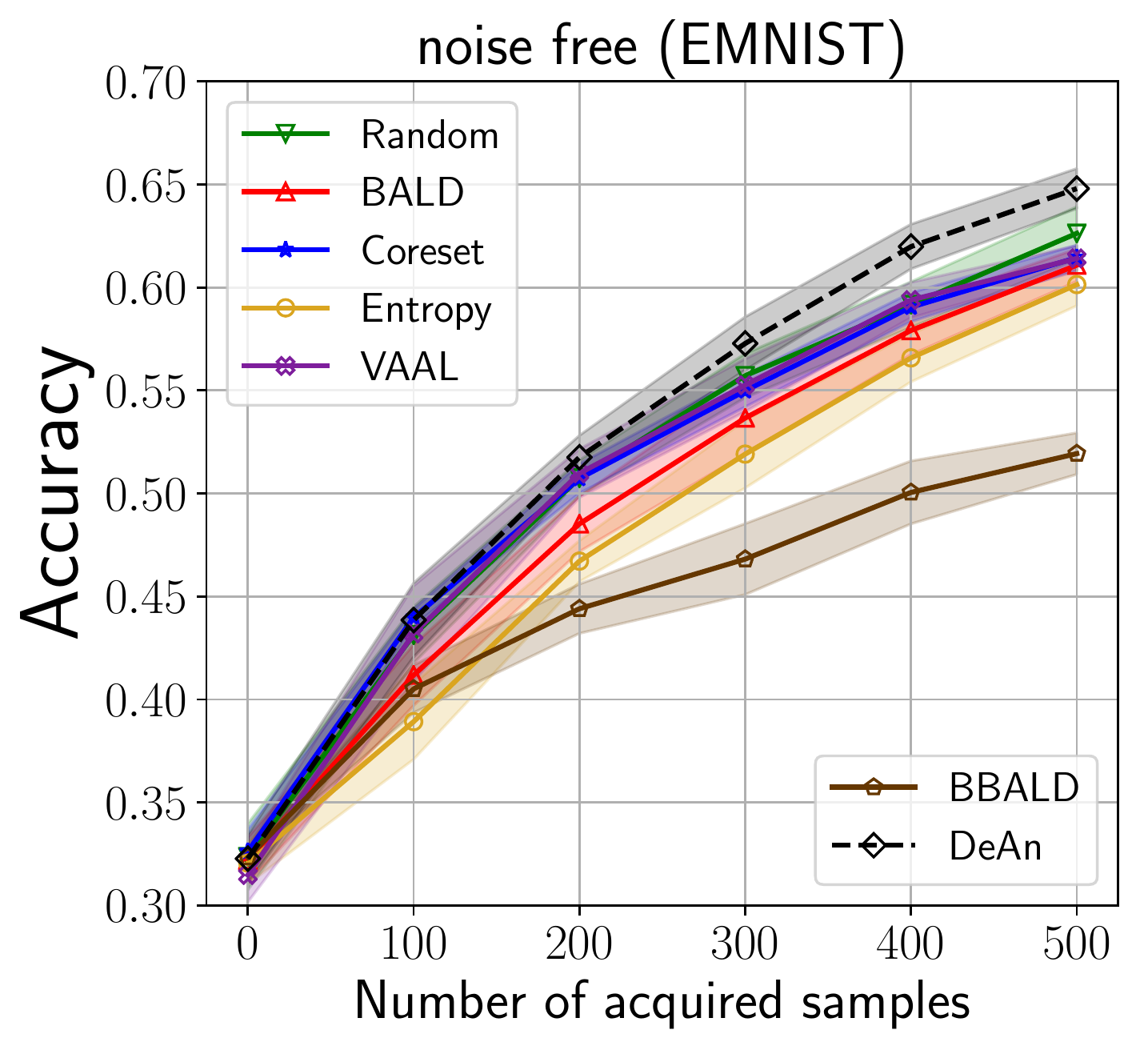}
		\hspace*{-5pt}
		\includegraphics*[height = 2.4in]{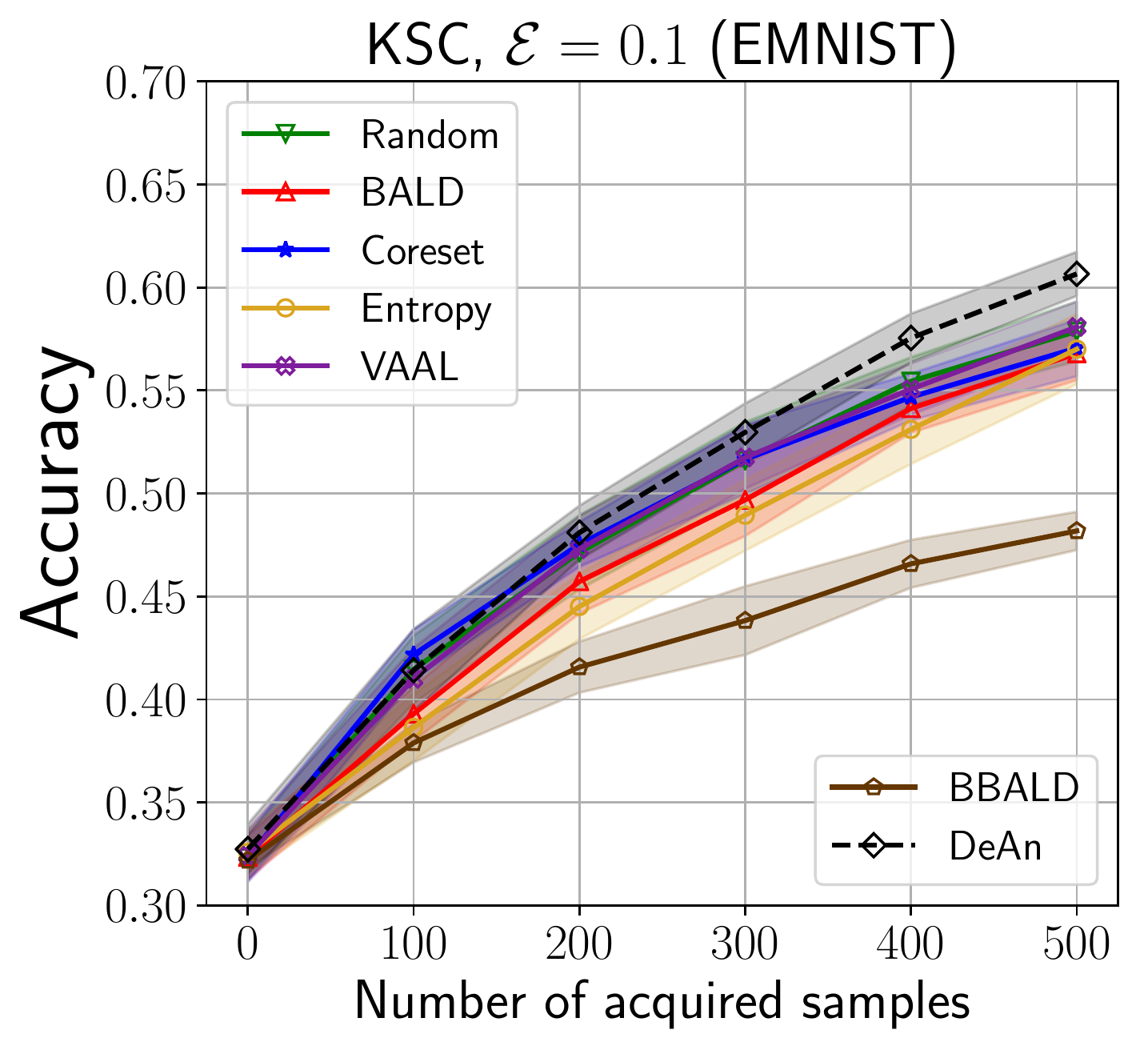}
	\end{subfigure}
	\begin{subfigure}[b]{\linewidth}
	\centering
		\includegraphics*[height = 2.4in]{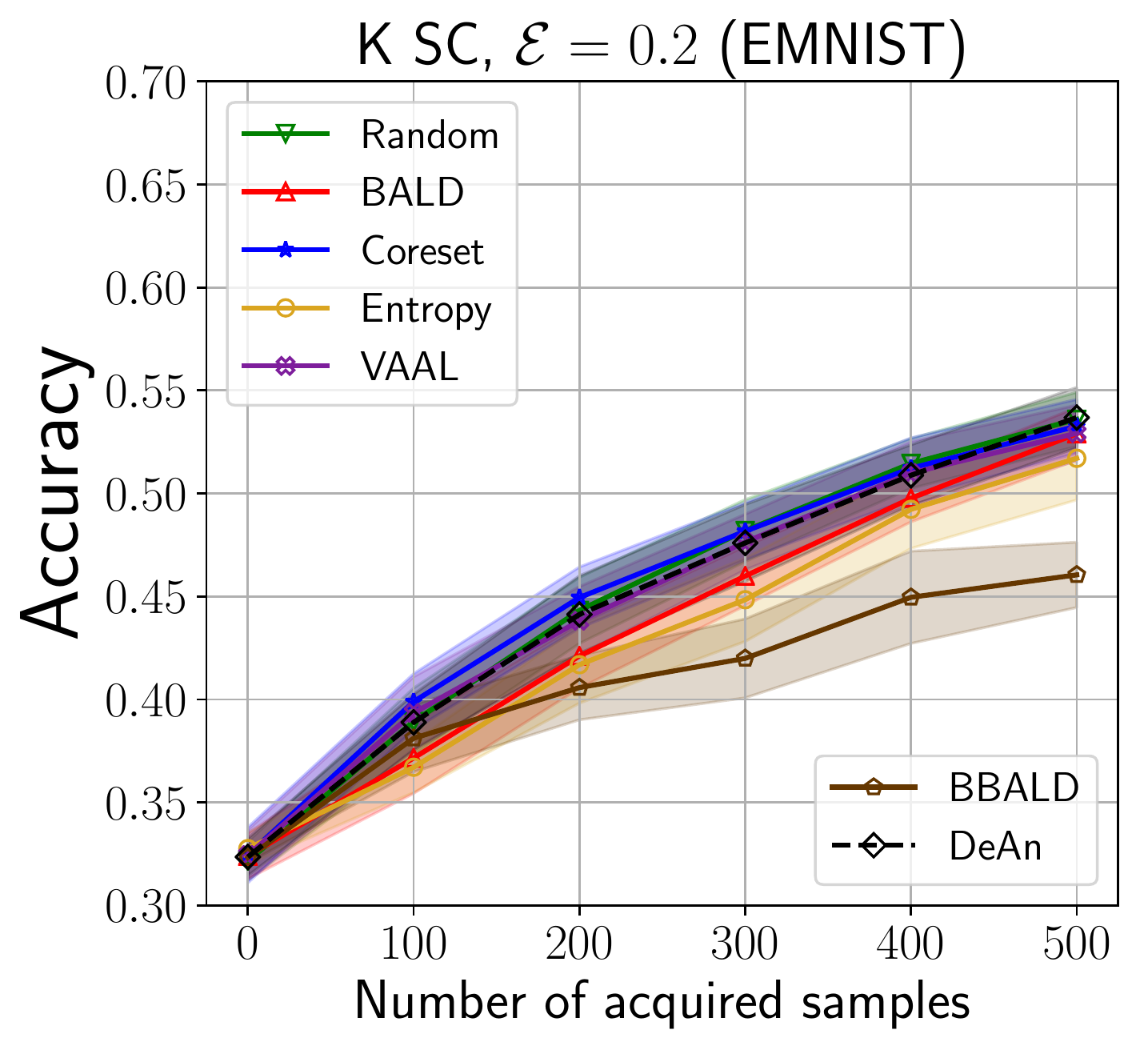}
		\hspace*{-5pt}
		\includegraphics*[height = 2.4in]{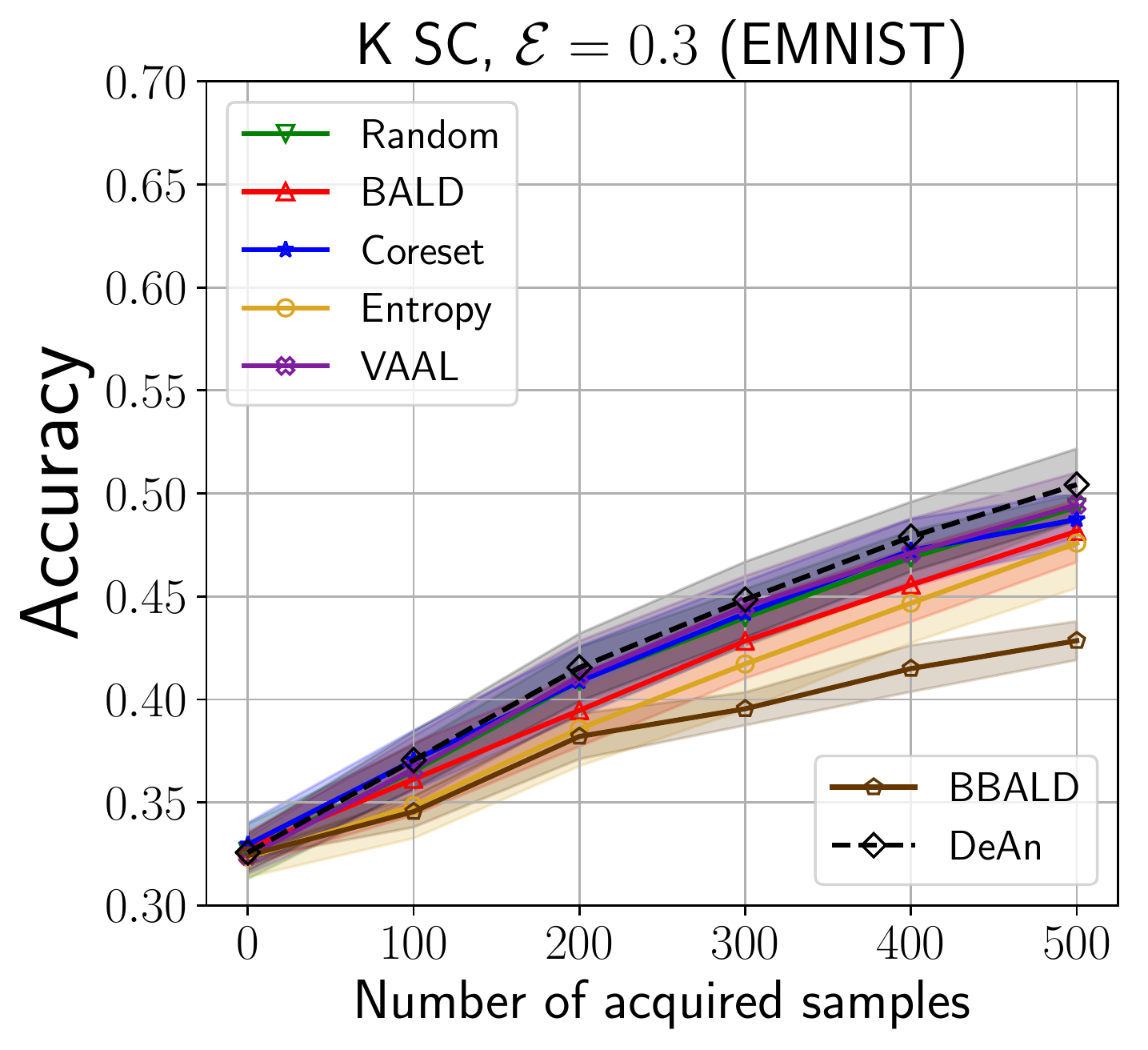}
	\end{subfigure}
	\begin{subfigure}[b]{\linewidth}
	\centering
		\includegraphics*[height = 2.4in]{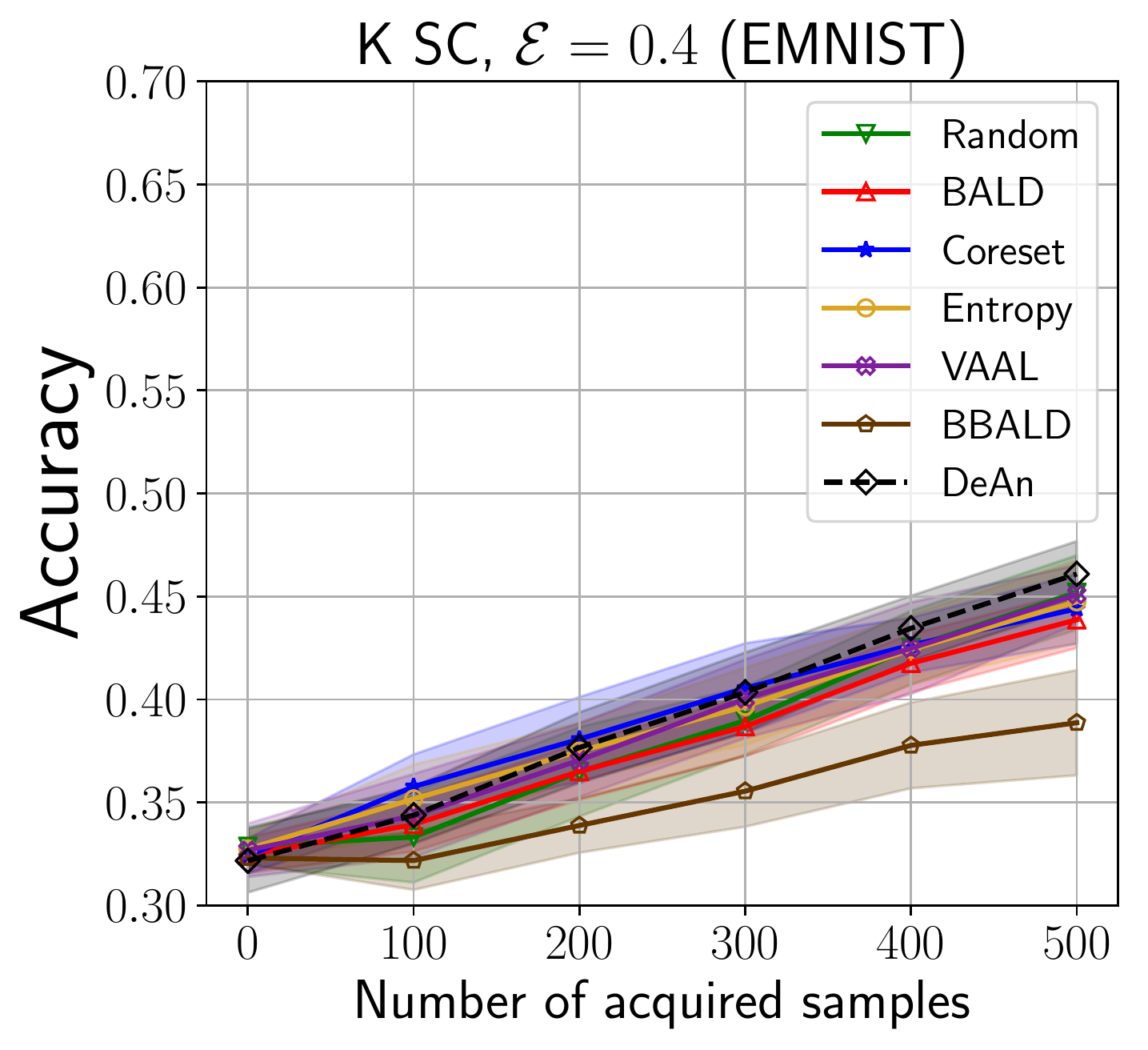}
		\hspace*{-5pt}
	\end{subfigure}
	\caption{Active learning results for various algorithms in noise free setting and under oracle noise strength $\varepsilon = 0.1, 0.3$ for EMNIST Image dataset with $b=100$.}
\label{fig:EMNIST_b_100}
\end{figure}

We perform an additional study with EMNIST along with what already presented in the main version of the paper. The acquisition size is taken to be $100$ to be able to use \texttt{BBALD} as well. We see in the Figure\,\ref{fig:EMNIST_b_100} that the proposed \texttt{DeAn} performs well in different noise strengths. We also observe that the recent BatchBALD perform inferior even to the BALD and Random. The reason being, computation of joint mutual information require $\mathcal{O}(K^{b})$ computations which is exponential. The Monte-Carlo sampling used in the BatchBALD work approximate this term, the error of which grows with increase in $K$ as well as $b$. For EMNIST, $K=47$ is large and with $b=100$ we observe inaccuracies in the selection criteria.

\subsection{CIFAR100}
\begin{figure}[ht]
	\centering
	\begin{subfigure}[b]{\linewidth}
	\centering
		\includegraphics*[height = 2.4in]{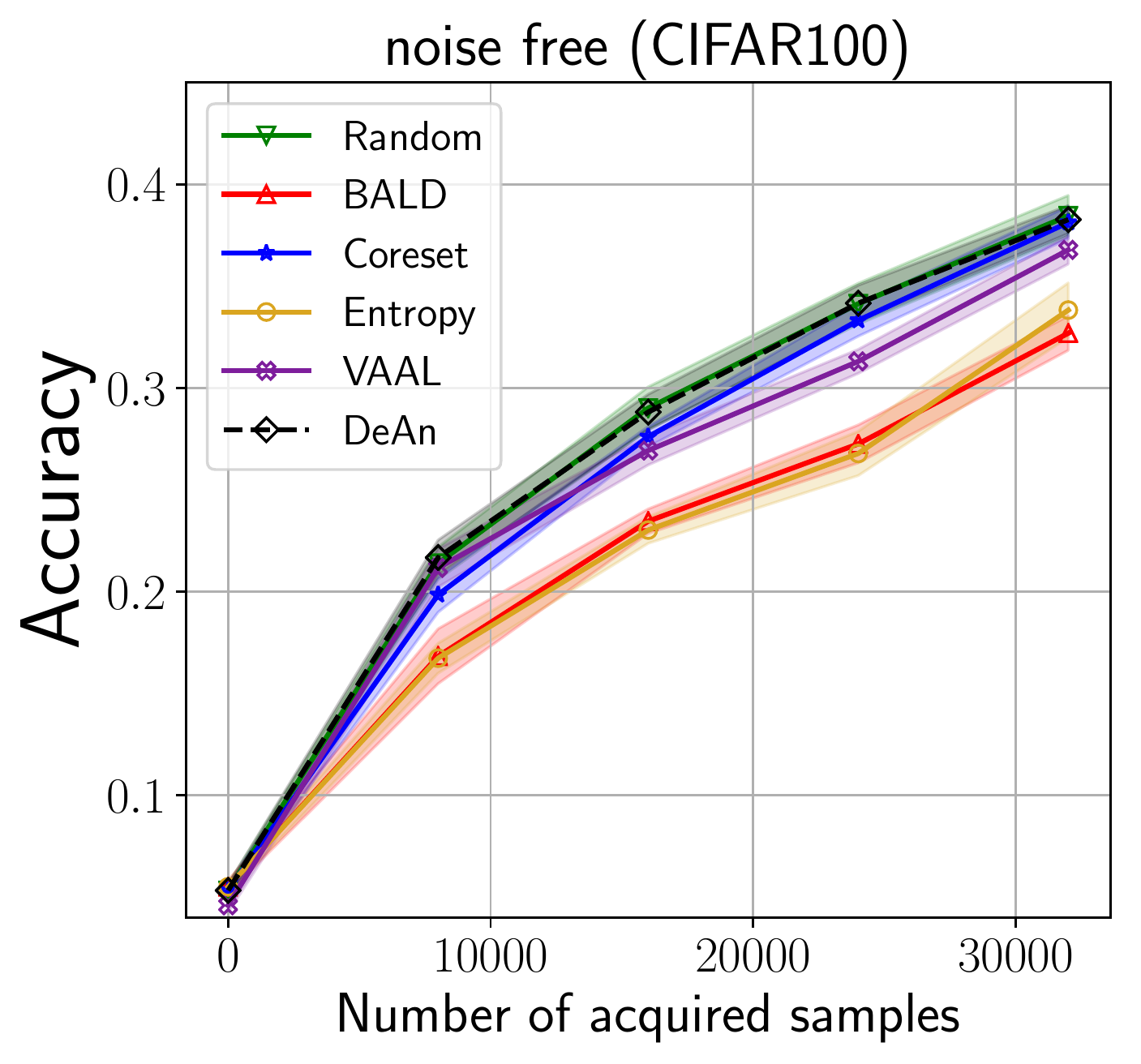}
		\hspace*{-5pt}
		\includegraphics*[height = 2.4in]{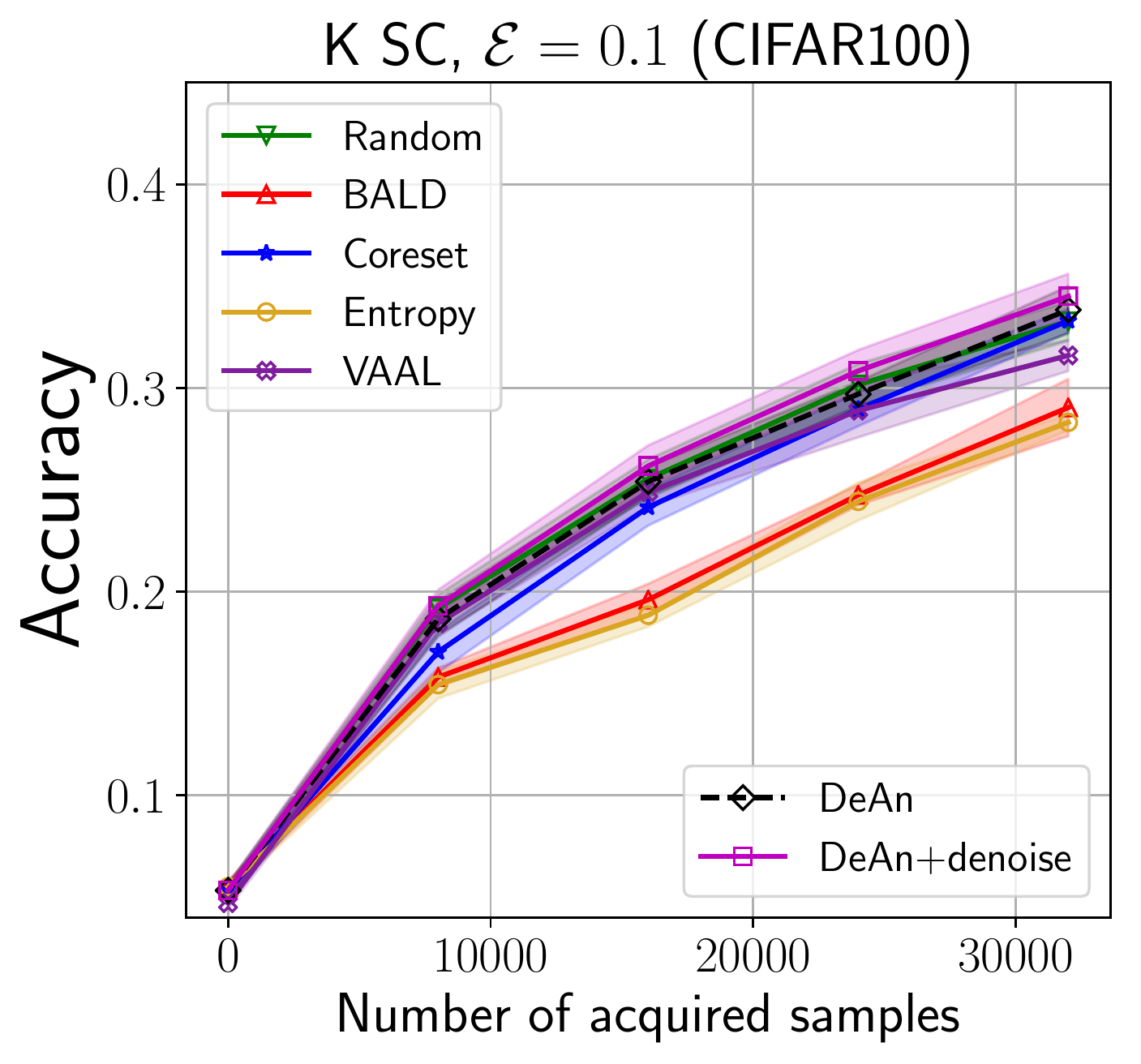}
	\end{subfigure}
	\begin{subfigure}[b]{\linewidth}
	\centering
		\includegraphics*[height = 2.4in]{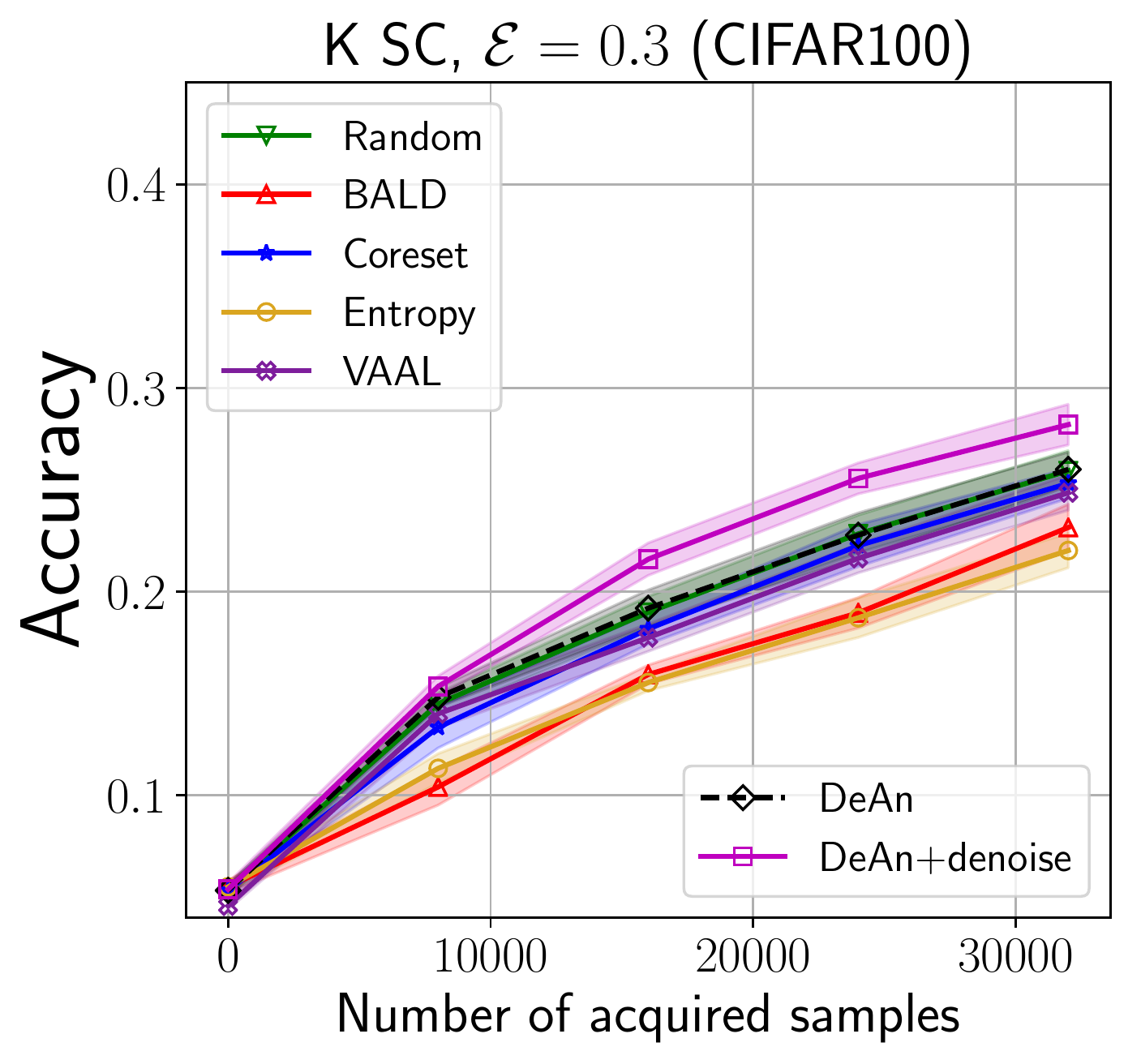}
		\hspace*{-5pt}
	\end{subfigure}
	\caption{Active learning results for various algorithms in noise free setting and under oracle noise strength $\varepsilon = 0.1, 0.3$ for CIFAR100 Image dataset.}
\label{fig:cifar100}
\end{figure}

Using the same setup as explained in Section\,5 of the paper, we evaluate the performance on CIFAR100 \citep{Krizhevsky09learningmultiple} dataset for various active learning algorithms listed in Section\,5.3. We observe in \figurename\,\ref{fig:cifar100} that the proposed uncertainty based algorithm perform similar or better than the baselines. The incorporation of denoising layer helps in countering the affects of noisy oracle as we demonstrate by varying the noise strength $\varepsilon = 0.1, 0.3$.

\section{Ablation Study}
\begin{figure}[ht]
	\centering
	\begin{subfigure}[b]{\linewidth}
	\centering
		\includegraphics*[height = 2.4in]{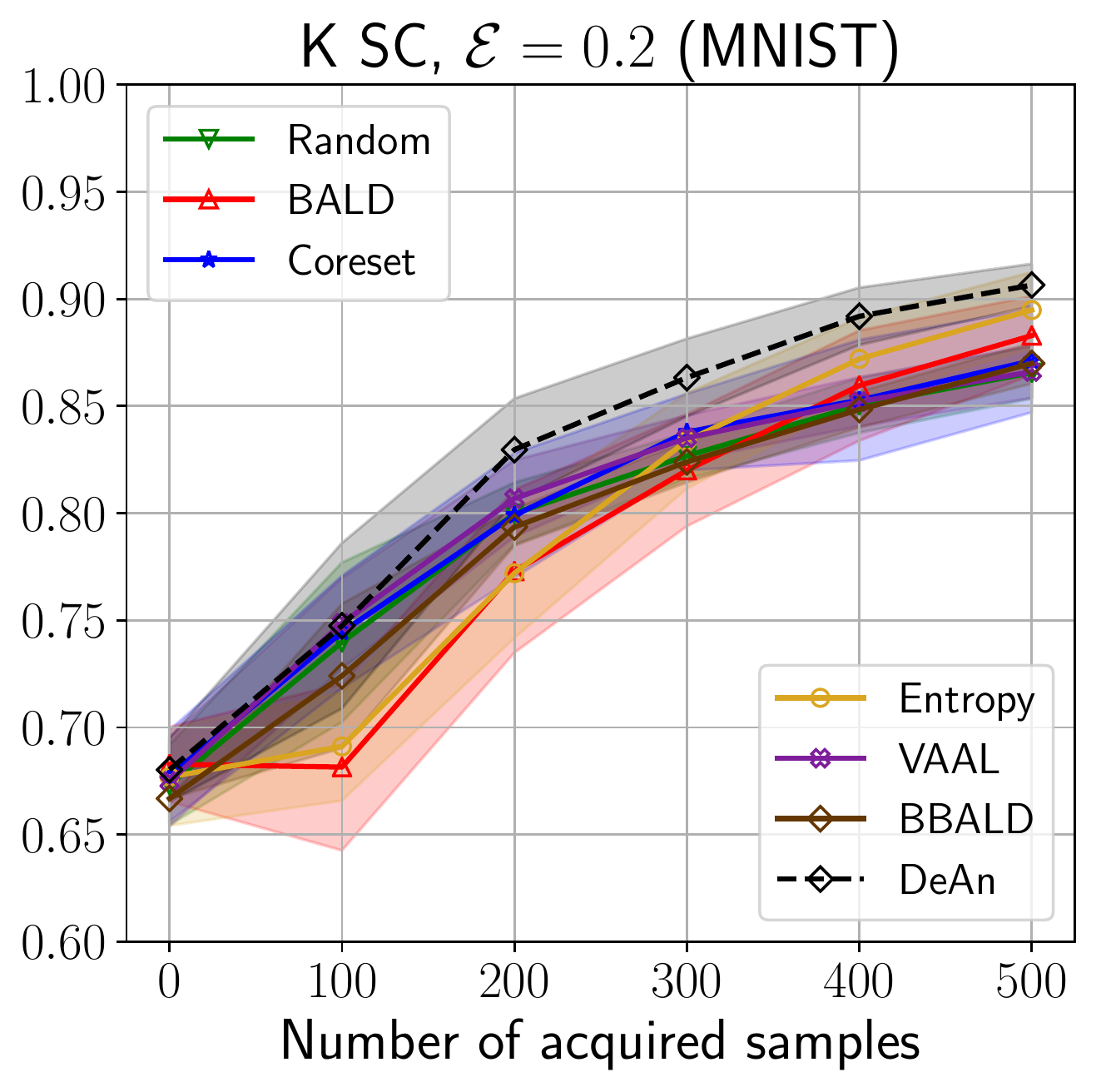}
		\includegraphics*[height = 2.4in]{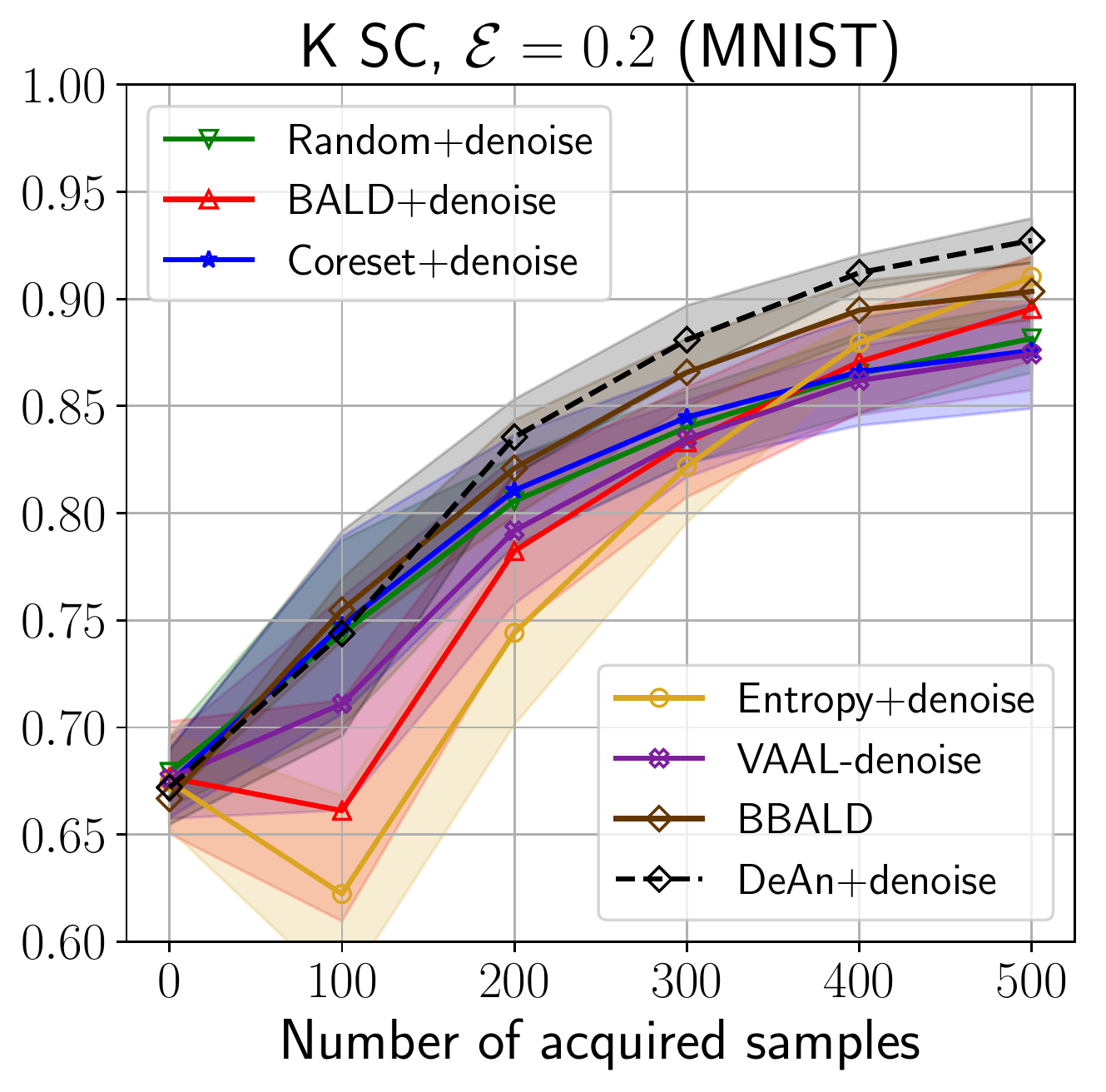}
	\end{subfigure}
	\caption{Active learning results for various algorithms under oracle noise strength $\varepsilon = 0.2$ for MNIST Image dataset.}
	\label{fig:ablation_mnist}
\end{figure}

The remaining of the results of the Ablation study section in the main paper are presented in Figure\,\ref{fig:ablation_mnist}
\begin{figure}[ht]
	\centering
	\begin{subfigure}[b]{\linewidth}
	\centering
		\includegraphics*[height = 2.4in]{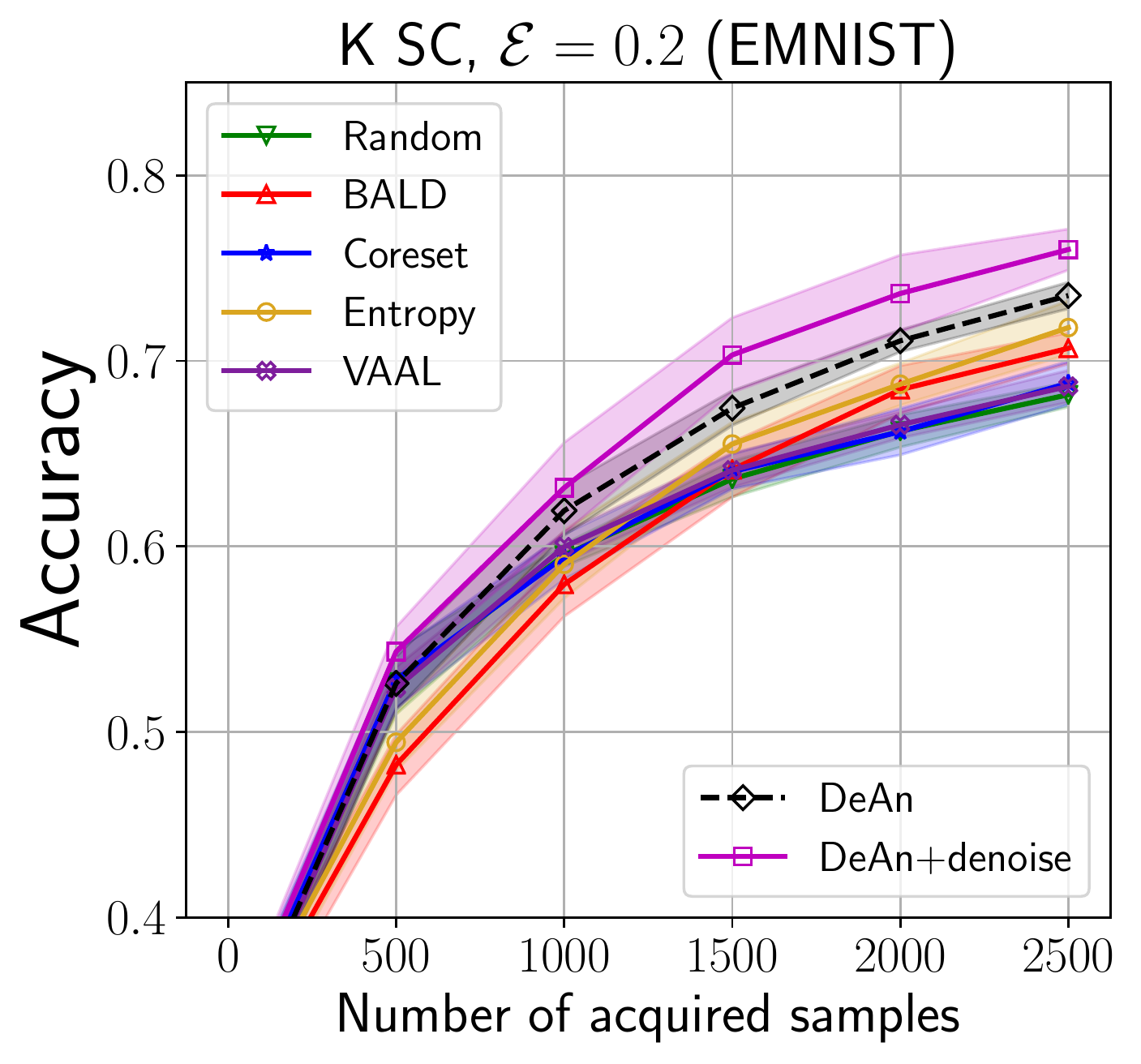}
		\hspace*{-5pt}
		\includegraphics*[height = 2.4in]{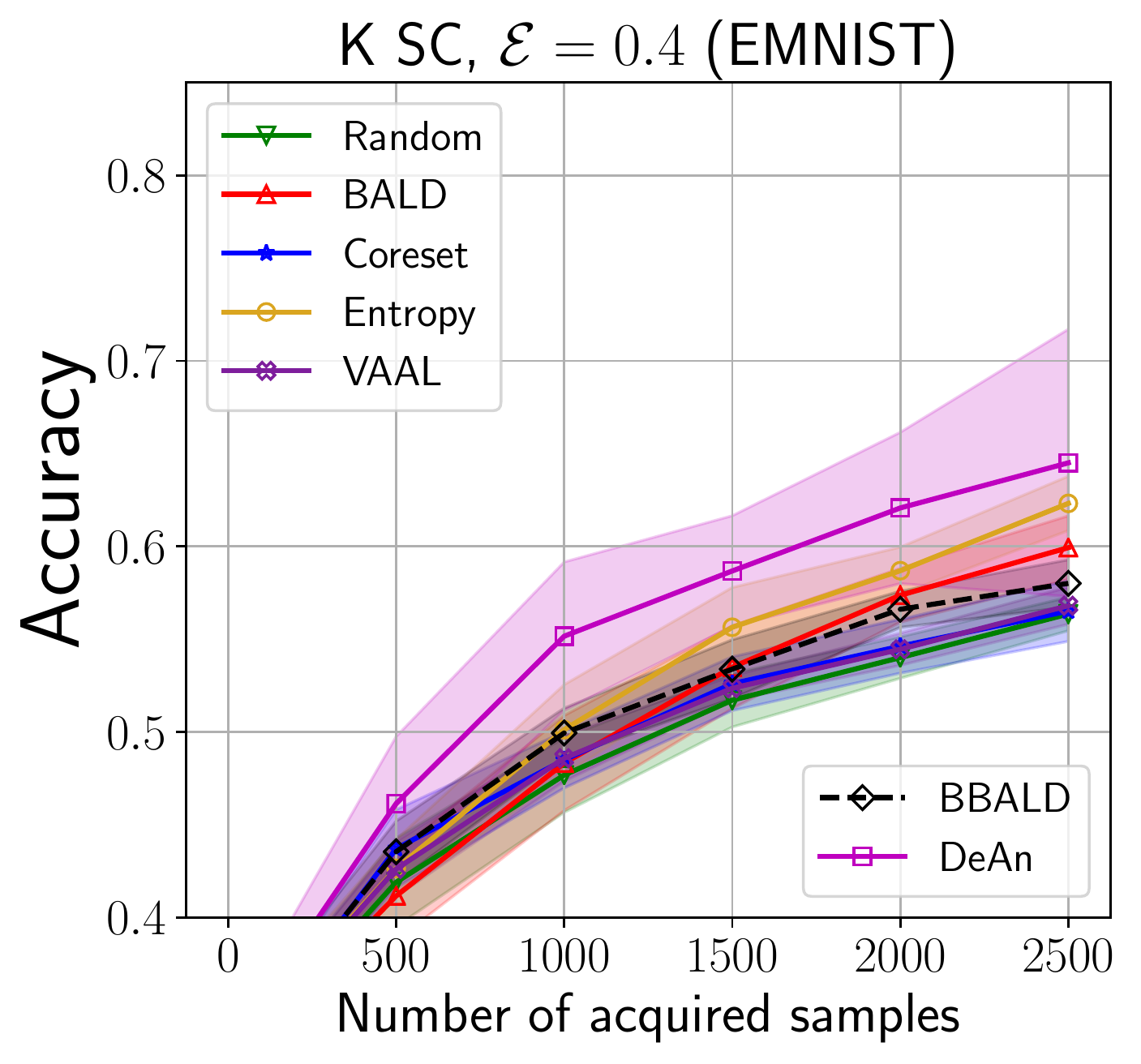}
	\end{subfigure}
	\caption{Active learning results for various algorithms in noise free setting and under oracle noise strength $\varepsilon = 0.2, 0.4$ for EMNIST Image dataset with $b=500$.}
\label{fig:EMNIST_b_500}
\end{figure}

\section{Active learning results}
\label{sec:tab_results}
For a quantitative look at the active learning results, mean and standard deviation of the performance vs. acquisition, in the \figurename\,4 of the paper, we present the results in the tabular format in Table\,\ref{tab:results_MNIST} for MNIST, Table\,\ref{tab:results_CIFAR10} for CIFAR10, Table\,\ref{tab:results_SVHN} for SVHN, Table\,\ref{tab:results_EMNIST_b_500}, \ref{tab:results_EMNIST_b_100} for EMNIST, and Table\,\ref{tab:results_CIFAR100} for CIFAR100, respectively.

\begin{table*}[t!]
    \caption{Active learning results for MNIST dataset.}
	\label{tab:results_MNIST}
	\centering
	\begin{tabular}{llllll}
		\toprule
		\multirow{2}*{Algorithm} & \multicolumn{5}{c}{Number of acquired samples}\\
		\cmidrule{2-6}
		& 200 & 400 & 600 & 800 & 1000 \\
		\midrule
		\multicolumn{6}{c}{noise free}\\
		\midrule
		Random & $87.19 \pm 0.84$ & $91.51 \pm 0.75$ & $93.40 \pm 0.42$ & $94.39 \pm 0.57$ & $95.17 \pm 0.46$ \\
		BALD  & $80.03 \pm 2.47$ & $89.84 \pm 2.35$ & $94.17 \pm 0.98$ & $96.23 \pm 0.72$ & $97.00 \pm 0.47$ \\
		Coreset & $87.87 \pm 1.53$ & $92.40 \pm 1.33$ & $94.53 \pm 1.00$ & $95.57 \pm 0.96$ & $96.02 \pm 0.74$\\
		Entropy & $77.69 \pm 2.34$ & $89.53 \pm 2.24$ & $94.94 \pm 0.69$ & $96.67 \pm 0.35$ & $97.28 \pm 0.48$\\
		VAAL & $87.21 \pm 1.00$ & $91.14 \pm 0.76$ & $93.39 \pm 0.51$ & $94.33 \pm 0.38$ & $95.25 \pm 0.34$ \\
		\textbf{DeAn} & $\bm{87.60 \pm 1.89}$ & $\bm{94.11 \pm 0.44}$ & $\bm{96.00 \pm 0.31}$ & $\bm{96.96 \pm 0.20}$ & $\bm{97.40 \pm 0.15}$\\
		\multicolumn{6}{c}{$\epsilon = 0.1$}\\
		\midrule
		Random & $81.87 \pm 2.42$ & $87.26 \pm 1.69$ & $89.55 \pm 0.89$ & $91.10 \pm 0.61$ & $91.94 \pm 0.56$ \\
		BALD  & $76.13 \pm 3.12$ & $85.45 \pm 2.84$ & $90.75 \pm 2.27$ & $94.01 \pm 0.73$ & $94.96 \pm 1.04$ \\
		Coreset & $83.46 \pm 2.49$ & $88.42 \pm 2.28$ & $90.80 \pm 1.85$ & $91.37 \pm 2.01$ & $93.09 \pm 1.13$\\
		Entropy & $74.28 \pm 3.33$ & $85.64 \pm 3.15$ & $91.59 \pm 1.84$ & $94.72 \pm 0.75$ & $95.73 \pm 0.60$\\
		VAAL & $82.80 \pm 1.43$ & $87.48 \pm 1.35$ & $89.70 \pm 1.10$ & $90.95 \pm 0.80$ & $91.91 \pm 0.66$ \\
		\textbf{DeAn} & $\bm{83.45 \pm 2.33}$ & $\bm{90.74 \pm 1.30}$ & $\bm{93.63 \pm 0.73}$ & $\bm{94.94 \pm 0.68}$ & $\bm{95.81 \pm 0.32}$\\
		\textbf{DeAn}\newline+\textbf{denoise} & $\bm{84.88 \pm 2.59}$ & $\bm{92.68 \pm 0.97}$ & $\bm{95.18 \pm 0.40}$ & $\bm{96.29 \pm 0.34}$ & $\bm{96.87 \pm 0.26}$\\
		\multicolumn{6}{c}{$\epsilon = 0.2$}\\
		\midrule
		Random & $77.57 \pm 2.74$ & $82.44 \pm 2.29$ & $85.03 \pm 2.07$ & $86.41 \pm 1.50$ & $87.73 \pm 1.25$ \\
		BALD  & $73.20 \pm 3.92$ & $81.43 \pm 3.44$ & $86.74 \pm 2.74$ & $90.06 \pm 1.64$ & $91.63 \pm 1.66$ \\
		Coreset & $79.97 \pm 2.96$ & $84.72 \pm 2.25$ & $86.88 \pm 2.07$ & $88.84 \pm 1.89$ & $88.93 \pm 1.52$\\
		Entropy & $70.84 \pm 4.81$ & $82.26 \pm 2.66$ & $88.44 \pm 1.76$ & $91.66 \pm 1.25$ & $93.09 \pm 1.03$\\
		VAAL & $77.21 \pm 1.84$ & $81.90 \pm 1.90$ & $85.17 \pm 1.22$ & $86.81 \pm 1.11$ & $88.06 \pm 1.06$ \\
		\textbf{DeAn} & $\bm{79.09 \pm 1.76}$ & $\bm{87.04 \pm 1.66}$ & $\bm{90.46 \pm 1.10}$ & $\bm{92.41 \pm 0.90}$ & $\bm{93.28 \pm 0.59}$\\
		\textbf{DeAn}\newline+\textbf{denoise} & $\bm{80.88 \pm 2.43}$ & $\bm{89.39 \pm 1.27}$ & $\bm{93.53 \pm 0.69}$ & $\bm{95.03 \pm 0.46}$ & $\bm{95.98 \pm 0.37}$\\
		\multicolumn{6}{c}{$\epsilon = 0.3$}\\
		\midrule
		Random & $73.10 \pm 2.85$ & $77.45 \pm 2.03$ & $79.88 \pm 2.31$ & $81.10 \pm 1.90$ & $82.66 \pm 1.55$ \\
		BALD  & $69.54 \pm 3.18$ & $76.28 \pm 3.00$ & $82.25 \pm 2.09$ & $85.10 \pm 2.06$ & $87.97 \pm 1.46$ \\
		Coreset & $74.58 \pm 3.07$ & $78.14 \pm 2.51$ & $79.89 \pm 2.77$ & $82.17 \pm 2.41$ & $83.49 \pm 2.02$\\
		Entropy & $68.11 \pm 3.87$ & $77.70 \pm 2.77$ & $83.91 \pm 2.70$ & $87.78 \pm 2.25$ & $89.37 \pm 1.34$\\
		VAAL & $71.42 \pm 2.47$ & $75.43 \pm 2.03$ & $79.30 \pm 1.56$ & $80.58 \pm 1.46$ & $82.27 \pm 1.51$ \\
		\textbf{DeAn} & $\bm{73.18 \pm 3.48}$ & $\bm{81.77 \pm 1.97}$ & $\bm{85.85 \pm 1.74}$ & $\bm{87.81 \pm 1.41}$ & $\bm{89.46 \pm 1.24}$\\
		\textbf{DeAn}\newline+\textbf{denoise} & $\bm{77.81 \pm 2.59}$ & $\bm{85.96 \pm 1.46}$ & $\bm{90.95 \pm 1.07}$ & $\bm{93.33 \pm 0.66}$ & $\bm{94.78 \pm 0.41}$\\
		\multicolumn{6}{c}{$\epsilon = 0.4$}\\
		\midrule
		Random & $66.14 \pm 2.93$ & $69.55 \pm 1.97$ & $72.80 \pm 2.30$ & $74.44 \pm 1.77$ & $76.19 \pm 1.66$ \\
		BALD  & $66.50 \pm 3.07$ & $71.77 \pm 2.39$ & $76.81 \pm 2.97$ & $79.47 \pm 2.66$ & $81.10 \pm 2.13$ \\
		Coreset & $68.41 \pm 4.51$ & $72.75 \pm 3.35$ & $75.03 \pm 3.17$ & $77.15 \pm 2.69$ & $77.25 \pm 3.08$\\
		Entropy & $66.31 \pm 2.68$ & $73.92 \pm 3.04$ & $79.99 \pm 1.86$ & $82.70 \pm 2.39$ & $84.25 \pm 1.81$\\
		VAAL & $65.57 \pm 3.35$ & $70.30 \pm 2.39$ & $72.90 \pm 1.83$ & $74.30 \pm 1.61$ & $75.31 \pm 1.51$ \\
		\textbf{DeAn} & $\bm{66.70 \pm 3.23}$ & $\bm{73.80 \pm 2.41}$ & $\bm{76.28 \pm 2.69}$ & $\bm{78.22 \pm 1.64}$ & $\bm{79.65 \pm 2.17}$\\
		\textbf{DeAn}\newline+\textbf{denoise} & $\bm{71.37 \pm 3.53}$ & $\bm{80.84 \pm 2.76}$ & $\bm{86.19 \pm 1.65}$ & $\bm{90.00 \pm 1.24}$ & $\bm{92.19 \pm 0.84}$\\
 		\bottomrule
	\end{tabular}
\end{table*}

\begin{table*}[t!]
		\centering
		\caption{Active learning results for CIFAR10 dataset.}
		\label{tab:results_CIFAR10}
		{\small
		\begin{tabular}{lllllll}
			\toprule
			\multirow{2}*{Algorithm} & \multicolumn{6}{c}{Number of acquired samples}\\
			\cmidrule{2-7}
			& 5000 & 10000 & 15000 & 20000 & 25000 & 30000 \\
			\midrule
			\multicolumn{7}{c}{noise free}\\
			\midrule
			Random &  $58.90 \pm 1.77$  &  $67.44 \pm 0.72$  &  $71.79 \pm 0.49$  &  $74.28 \pm 0.53$  &  $75.99 \pm 0.17$  &  $77.58 \pm 0.50$  \\
			BALD  &  $50.42 \pm 1.99$  &  $65.19 \pm 1.21$  &  $71.58 \pm 0.38$  &  $75.07 \pm 0.71$  &  $76.90 \pm 0.82$  &  $78.35 \pm 1.01$  \\
			Coreset &  $53.62 \pm 1.13$  &  $63.56 \pm 1.74$  &  $68.65 \pm 1.28$  &  $71.99 \pm 0.55$  &  $73.58 \pm 0.49$  &  $76.10 \pm 0.48$  \\
			Entropy &  $53.83 \pm 4.60$  &  $64.89 \pm 1.27$  &  $70.40 \pm 1.53$  &  $73.85 \pm 1.25$  &  $76.71 \pm 0.81$  &  $78.11 \pm 0.57$  \\
			VAAL &  $57.40 \pm 0.72$ & $66.98 \pm 0.81$ & $70.83 \pm 0.42$ & $73.43 \pm 0.23$ & $75.44 \pm 0.27$ & $76.59 \pm 0.46$  \\
			\textbf{DeAn} &  $\bm{59.28 \pm 1.62}$  &  $\bm{67.31 \pm 0.25}$  &  $\bm{72.92 \pm 0.57}$  &  $\bm{74.79 \pm 0.48}$  &  $\bm{77.09 \pm 0.73}$  &  $\bm{78.28 \pm 0.71}$  \\
			\multicolumn{7}{c}{$\epsilon = 0.1$}\\
			\midrule
			Random &  $54.52 \pm 1.41$  &  $61.93 \pm 1.14$  &  $66.37 \pm 0.59$  &  $69.06 \pm 0.78$  &  $71.12 \pm 0.66$  &  $72.97 \pm 0.70$  \\
			BALD  &  $46.83 \pm 1.76$  &  $60.17 \pm 1.61$  &  $66.85 \pm 0.97$  &  $69.81 \pm 0.71$  &  $72.42 \pm 0.79$  &  $74.27 \pm 0.64$  \\
			Coreset &  $50.27 \pm 1.98$  &  $58.53 \pm 1.51$  &  $64.33 \pm 1.20$  &  $67.89 \pm 0.73$  &  $70.31 \pm 0.62$  &  $72.36 \pm 0.58$  \\
			Entropy &  $48.63 \pm 2.78$  &  $60.05 \pm 1.70$  &  $65.89 \pm 1.33$  &  $70.31 \pm 1.28$  &  $72.43 \pm 0.99$  &  $74.69 \pm 0.89$  \\
			VAAL &  $53.47 \pm 1.17$ & $61.08 \pm 1.21$ & $66.22 \pm 0.89$ & $68.54 \pm 0.64$ & $70.60 \pm 0.36$ & $71.78 \pm 0.43$  \\
			\textbf{ DeAn} &  $\bm{53.96 \pm 1.11}$  &  $\bm{62.10 \pm 1.21}$  &  $\bm{66.81 \pm 0.80}$  &  $\bm{70.30 \pm 0.87}$  &  $\bm{72.28 \pm 0.76}$  &  $\bm{73.68 \pm 0.44}$  \\
			\makecell[l]{\textbf{DeAn}\\+\textbf{denoise}} &  $\bm{52.95 \pm 1.54}$  &  $\bm{62.20 \pm 0.97}$  &  $\bm{67.54 \pm 1.15}$  &  $\bm{71.44 \pm 0.96}$  &  $\bm{73.94 \pm 0.85}$  &  $\bm{76.03 \pm 0.66}$  \\
			\multicolumn{7}{c}{$\epsilon = 0.2$}\\
			\midrule
			Random &  $49.46 \pm 1.63$  &  $57.21 \pm 1.04$  &  $61.64 \pm 0.88$  &  $64.60 \pm 1.14$  &  $67.31 \pm 0.96$  &  $68.77 \pm 0.71$  \\
			BALD  &  $42.79 \pm 1.81$  &  $54.98 \pm 1.31$  &  $60.81 \pm 0.96$  &  $65.23 \pm 0.93$  &  $67.97 \pm 0.72$  &  $69.78 \pm 0.69$  \\
			Coreset &  $45.89 \pm 2.26$  &  $53.70 \pm 1.70$  &  $59.00 \pm 1.34$  &  $63.23 \pm 0.84$  &  $66.23 \pm 0.62$  &  $68.44 \pm 0.88$  \\
			Entropy &  $44.37 \pm 3.23$  &  $55.50 \pm 2.06$  &  $60.70 \pm 1.80$  &  $64.75 \pm 1.58$  &  $68.10 \pm 1.36$  &  $69.72 \pm 0.96$  \\
			VAAL & $49.75 \pm 0.62$ & $56.36 \pm 1.10$ & $61.22 \pm 0.56$ & $63.42 \pm 0.77$ & $66.25 \pm 0.38$ & $67.82 \pm 0.64$ \\
			\textbf{DeAn} &  $\bm{49.38 \pm 1.40}$  &  $\bm{57.45 \pm 1.29}$  &  $\bm{61.47 \pm 1.31}$  &  $\bm{65.45 \pm 0.92}$  &  $\bm{68.08 \pm 1.13}$  &  $\bm{69.37 \pm 0.60}$  \\
			\makecell[l]{\textbf{DeAn}\\+\textbf{denoise}} &  $\bm{49.03 \pm 1.01}$  &  $\bm{58.85 \pm 1.32}$  &  $\bm{64.10 \pm 1.09}$  &  $\bm{68.07 \pm 1.48}$  &  $\bm{71.18 \pm 0.96}$  &  $\bm{72.65 \pm 0.96}$  \\
			\multicolumn{7}{c}{$\epsilon = 0.3$}\\
			\midrule
			Random &  $44.82 \pm 2.02$  &  $52.44 \pm 0.91$  &  $56.58 \pm 1.52$  &  $60.33 \pm 1.09$  &  $62.63 \pm 0.87$  &  $64.82 \pm 1.07$  \\
			BALD  &  $38.70 \pm 1.83$  &  $49.70 \pm 1.64$  &  $55.38 \pm 1.83$  &  $59.76 \pm 1.34$  &  $62.85 \pm 0.77$  &  $65.10 \pm 0.99$  \\
			Coreset &  $41.27 \pm 1.75$  &  $49.46 \pm 1.32$  &  $54.74 \pm 1.15$  &  $58.98 \pm 1.19$  &  $61.54 \pm 0.84$  &  $64.03 \pm 0.82$  \\
			Entropy &  $39.20 \pm 2.62$  &  $50.23 \pm 2.35$  &  $56.11 \pm 1.49$  &  $60.66 \pm 1.53$  &  $63.88 \pm 0.97$  &  $65.31 \pm 1.24$  \\
			VAAL &  $45.00 \pm 1.36$  &  $50.41 \pm 1.52$  &  $57.02 \pm 0.81$  &  $60.03 \pm 0.58$  &  $61.20 \pm 1.79$  &  $62.56 \pm 1.10$  \\
			\textbf{ DeAn} &  $\bm{44.54 \pm 2.02}$  &  $\bm{51.49 \pm 1.57}$  &  $\bm{56.86 \pm 1.34}$  &  $\bm{60.52 \pm 1.38}$  &  $\bm{63.15 \pm 0.85}$  &  $\bm{64.84 \pm 0.83}$  \\
			\makecell[l]{\textbf{DeAn}\\+\textbf{denoise}} &  $\bm{45.14 \pm 1.58}$  &  $\bm{53.72 \pm 1.55}$  &  $\bm{59.64 \pm 1.30}$  &  $\bm{63.35 \pm 1.63}$  &  $\bm{66.44 \pm 1.21}$  &  $\bm{68.80 \pm 1.11}$  \\
			\multicolumn{6}{c}{$\epsilon = 0.4$}\\
			\midrule
			Random &  $40.62 \pm 1.81$  &  $46.61 \pm 1.90$  &  $51.87 \pm 1.18$  &  $55.21 \pm 1.35$  &  $57.28 \pm 1.11$  &  $59.79 \pm 1.32$  \\
			BALD  &  $34.52 \pm 1.59$  &  $43.55 \pm 2.09$  &  $49.15 \pm 1.66$  &  $53.85 \pm 1.82$  &  $57.19 \pm 0.87$  &  $59.53 \pm 0.98$  \\
			Coreset &  $36.52 \pm 1.83$  &  $44.52 \pm 1.50$  &  $49.97 \pm 1.58$  &  $53.32 \pm 1.02$  &  $56.86 \pm 0.94$  &  $59.46 \pm 0.98$  \\
			Entropy &  $35.08 \pm 2.76$  &  $44.21 \pm 1.82$  &  $50.61 \pm 1.20$  &  $54.65 \pm 1.78$  &  $56.66 \pm 2.16$  &  $59.74 \pm 1.84$  \\
			VAAL & $38.62 \pm 0.97$ & $46.68 \pm 1.99$ & $50.88 \pm 0.57$ & $54.57 \pm 1.70$ & $57.04 \pm 0.22$ & $58.51 \pm 1.11$ \\
			\textbf{DeAn} &  $\bm{39.69 \pm 1.40}$  &  $\bm{46.67 \pm 1.43}$  &  $\bm{51.47 \pm 1.38}$  &  $\bm{54.63 \pm 1.66}$  &  $\bm{58.16 \pm 0.80}$  &  $\bm{59.87 \pm 1.38}$  \\
			\makecell[l]{\textbf{DeAn}\\+\textbf{denoise}} &  $\bm{39.68 \pm 1.34}$  &  $\bm{47.64 \pm 1.56}$  &  $\bm{52.35 \pm 1.89}$  &  $\bm{56.56 \pm 2.89}$  &  $\bm{59.28 \pm 1.88}$  &  $\bm{62.12 \pm 1.19}$  \\
        \bottomrule
		\end{tabular}
		}
\end{table*}

\begin{table*}[t!]
	\centering
	\caption{Active learning results for SVHN dataset.}
	\label{tab:results_SVHN}
	\begin{tabular}{llllll}
		\toprule
		\multirow{2}*{Algorithm} & \multicolumn{5}{c}{Number of acquired samples}\\
		\cmidrule{2-6}
		& 5000 & 10000 & 15000 & 20000 & 25000 \\
		\midrule
		\multicolumn{6}{c}{noise free}\\
		\midrule
		Random & $86.92 \pm 0.81$ & $90.04 \pm 0.75$ & $91.46 \pm 0.40$ & $92.29 \pm 0.44$ & $92.58 \pm 0.09$ \\
		BALD  & $82.46 \pm 0.89$ & $91.18 \pm 0.42$ & $92.81 \pm 0.43$ & $94.21 \pm 0.31$ & $94.47 \pm 0.39$ \\
		Coreset & $85.64 \pm 1.62$ & $90.05 \pm 0.52$ & $90.76 \pm 0.30$ & $91.92 \pm 0.23$ & $92.43 \pm 0.31$\\
		Entropy & $83.28 \pm 1.11$ & $91.23 \pm 0.50$ & $93.12 \pm 0.31$ & $93.88 \pm 0.12$ & $94.25 \pm 0.14$\\
		VAAL & $86.43 \pm 0.53$ & $89.99 \pm 0.37$ & $91.35 \pm 0.28$ & $92.04 \pm 0.56$ & $92.78 \pm 0.43$ \\
		\textbf{ DeAn} & $\bm{86.30 \pm 0.90}$ & $\bm{91.88 \pm 0.35}$ & $\bm{93.46 \pm 0.40}$ & $\bm{94.05 \pm 0.48}$ & $\bm{94.18 \pm 0.28}$\\
		\multicolumn{6}{c}{$\epsilon = 0.1$}\\
		\midrule
		Random & $82.34 \pm 1.16$ & $86.46 \pm 0.75$ & $88.63 \pm 0.74$ & $89.65 \pm 0.62$ & $90.05 \pm 0.70$ \\
		BALD  & $76.26 \pm 2.78$ & $87.15 \pm 0.82$ & $90.37 \pm 0.55$ & $91.73 \pm 0.49$ & $92.54 \pm 0.52$ \\
		Coreset & $80.99 \pm 2.27$ & $86.08 \pm 0.86$ & $88.08 \pm 0.69$ & $89.45 \pm 0.63$ & $90.19 \pm 0.76$\\
		Entropy & $78.83 \pm 2.12$ & $88.12 \pm 0.80$ & $90.32 \pm 0.49$ & $91.67 \pm 0.34$ & $92.54 \pm 0.45$\\
		VAAL & $81.99 \pm 0.49$ & $86.64 \pm 0.42$ & $88.76 \pm 0.61$ & $88.35 \pm 0.89$ & $89.74 \pm 0.74$ \\
		\textbf{ DeAn} & $\bm{81.61 \pm 1.30}$ & $\bm{88.54 \pm 0.91}$ & $\bm{90.37 \pm 0.64}$ & $\bm{91.60 \pm 0.59}$ & $\bm{92.24 \pm 0.54}$\\
		\textbf{ DeAn}\newline+\textbf{denoise} & $\bm{83.01 \pm 1.31}$ & $\bm{89.76 \pm 0.96}$ & $\bm{91.98 \pm 0.41}$ & $\bm{93.15 \pm 0.37}$ & $\bm{93.78 \pm 0.21}$\\
		\multicolumn{6}{c}{$\epsilon = 0.2$}\\
		\midrule
		Random & $78.46 \pm 1.73$ & $83.87 \pm 1.11$ & $85.57 \pm 1.24$ & $87.60 \pm 0.82$ & $88.27 \pm 0.92$ \\
		BALD  & $70.82 \pm 4.09$ & $83.97 \pm 1.59$ & $87.08 \pm 1.24$ & $89.42 \pm 0.82$ & $90.53 \pm 0.76$ \\
		Coreset & $76.75 \pm 2.32$ & $82.65 \pm 1.75$ & $85.43 \pm 1.08$ & $86.60 \pm 1.19$ & $88.23 \pm 1.03$\\
		Entropy & $74.00 \pm 1.30$ & $84.29 \pm 1.33$ & $87.46 \pm 0.99$ & $89.30 \pm 0.89$ & $90.12 \pm 0.87$\\
		\textbf{ DeAn} & $\bm{77.98 \pm 2.27}$ & $\bm{84.82 \pm 1.43}$ & $\bm{87.86 \pm 1.38}$ & $\bm{88.86 \pm 1.10}$ & $\bm{90.33 \pm 0.67}$\\
		\textbf{ DeAn}\newline+\textbf{denoise} & $\bm{81.09 \pm 1.75}$ & $\bm{88.47 \pm 1.39}$ & $\bm{91.19 \pm 0.58}$ & $\bm{92.47 \pm 0.47}$ & $\bm{93.16 \pm 0.31}$\\
		\multicolumn{6}{c}{$\epsilon = 0.3$}\\
		\midrule
		Random & $73.75 \pm 2.33$ & $79.07 \pm 1.79$ & $82.53 \pm 1.31$ & $83.74 \pm 1.49$ & $85.49 \pm 1.14$ \\
		BALD  & $65.72 \pm 4.43$ & $79.72 \pm 1.52$ & $83.38 \pm 1.72$ & $86.15 \pm 1.33$ & $88.14 \pm 0.95$ \\
		Coreset & $71.84 \pm 2.52$ & $78.45 \pm 1.86$ & $81.87 \pm 1.55$ & $83.99 \pm 1.38$ & $85.78 \pm 1.28$\\
		Entropy & $68.52 \pm 2.97$ & $79.63 \pm 1.53$ & $83.92 \pm 1.25$ & $85.98 \pm 1.48$ & $87.84 \pm 1.02$\\
		VAAL & $72.69 \pm 1.11$ & $78.17 \pm 1.95$ & $81.51 \pm 1.84$ & $84.18 \pm 1.03$ & $85.77 \pm 1.22$ \\
		\textbf{ DeAn} & $\bm{73.04 \pm 2.60}$ & $\bm{81.14 \pm 1.23}$ & $\bm{83.14 \pm 1.62}$ & $\bm{85.87 \pm 1.17}$ & $\bm{87.71 \pm 1.37}$\\
		\textbf{ DeAn}\newline+\textbf{denoise} & $\bm{78.80 \pm 1.47}$ & $\bm{87.19 \pm 1.59}$ & $\bm{90.04 \pm 0.76}$ & $\bm{91.56 \pm 0.38}$ & $\bm{92.26 \pm 0.48}$\\
		\multicolumn{6}{c}{$\epsilon = 0.4$}\\
		\midrule
		Random & $68.23 \pm 2.01$ & $73.07 \pm 2.47$ & $77.46 \pm 2.27$ & $79.79 \pm 2.37$ & $81.52 \pm 1.93$ \\
		BALD  & $58.76 \pm 4.05$ & $73.83 \pm 2.25$ & $78.98 \pm 2.12$ & $81.78 \pm 1.65$ & $83.59 \pm 1.68$ \\
		Coreset & $65.26 \pm 3.40$ & $73.23 \pm 2.51$ & $77.18 \pm 1.80$ & $80.44 \pm 1.40$ & $81.78 \pm 1.87$\\
		Entropy & $61.36 \pm 4.37$ & $72.97 \pm 1.76$ & $78.61 \pm 2.21$ & $81.40 \pm 1.03$ & $83.88 \pm 0.59$\\
		\textbf{ DeAn} & $\bm{67.23 \pm 3.37}$ & $\bm{74.08 \pm 2.73}$ & $\bm{78.15 \pm 2.15}$ & $\bm{80.48 \pm 1.56}$ & $\bm{82.78 \pm 1.68}$\\
		\textbf{ DeAn}\newline+\textbf{denoise} & $\bm{76.07 \pm 2.58}$ & $\bm{83.89 \pm 1.00}$ & $\bm{88.46 \pm 0.66}$ & $\bm{90.53 \pm 0.51}$ & $\bm{91.58 \pm 0.32}$\\
        \bottomrule
	\end{tabular}
\end{table*}

\begin{table*}[t!]
    \caption{Active learning results for EMNIST dataset (b=500).}
	\label{tab:results_EMNIST_b_500}
	\centering
	\begin{tabular}{llllll}
		\toprule
		\multirow{2}*{Algorithm} & \multicolumn{5}{c}{Number of acquired samples}\\
		\cmidrule{2-6}
		& 200 & 400 & 600 & 800 & 1000 \\
		\midrule
		\multicolumn{6}{c}{noise free}\\
		\midrule
		Random & $61.67 \pm 1.27$ & $69.54 \pm 0.72$ & $73.68 \pm 0.60$ & $75.88 \pm 0.58$ & $77.66 \pm 0.61$ \\
        BALD  & $55.27 \pm 1.63$ & $66.97 \pm 1.21$ & $72.80 \pm 1.18$ & $75.91 \pm 1.02$ & $78.06 \pm 0.90$ \\
        Coreset & $61.10 \pm 1.13$ & $69.53 \pm 1.00$ & $73.46 \pm 0.33$ & $76.34 \pm 0.79$ & $77.96 \pm 0.89$ \\
        Entropy & $56.56 \pm 1.14$ & $67.50 \pm 1.06$ & $74.04 \pm 1.16$ & $77.03 \pm 1.16$ & $78.74 \pm 0.80$ \\
        VAAL & $61.12 \pm 0.85$ & $69.30 \pm 0.94$ & $73.28 \pm 0.72$ & $75.88 \pm 0.52$ & $77.18 \pm 0.44$ \\
        \textbf{DeAn} & $\bm{61.86 \pm 1.13}$ & $\bm{72.45 \pm 0.67}$ & $\bm{76.50 \pm 0.43}$ & $\bm{79.06 \pm 0.44}$ & $\bm{80.54 \pm 0.34}$ \\
		\multicolumn{6}{c}{$\epsilon = 0.1$}\\
		\midrule
		Random & $57.21 \pm 1.18$ & $64.47 \pm 0.97$ & $68.51 \pm 1.07$ & $71.08 \pm 0.60$ & $73.15 \pm 0.58$ \\
        BALD  & $52.08 \pm 1.12$ & $62.76 \pm 1.56$ & $68.49 \pm 1.39$ & $72.45 \pm 0.99$ & $75.03 \pm 0.78$ \\
        Coreset & $56.43 \pm 1.12$ & $63.89 \pm 1.08$ & $68.47 \pm 0.95$ & $71.98 \pm 1.05$ & $73.85 \pm 1.06$ \\
        Entropy & $52.71 \pm 1.74$ & $62.60 \pm 1.06$ & $69.29 \pm 0.75$ & $73.02 \pm 0.82$ & $75.88 \pm 1.06$ \\
        VAAL & $57.08 \pm 1.28$ & $64.35 \pm 1.04$ & $68.26 \pm 0.91$ & $70.79 \pm 0.74$ & $72.85 \pm 0.69$ \\
        \textbf{DeAn} & $\bm{57.54 \pm 1.45}$ & $\bm{67.33 \pm 0.92}$ & $\bm{72.10 \pm 0.56}$ & $\bm{75.46 \pm 0.50}$ & $\bm{77.37 \pm 0.41}$ \\
        \textbf{DeAn}\newline+\textbf{denoise} & $\bm{57.77 \pm 1.61}$ & $\bm{67.92 \pm 2.08}$ & $\bm{72.65 \pm 1.36}$ & $\bm{76.65 \pm 1.61}$ & $\bm{78.17 \pm 1.22}$ \\ 
		\multicolumn{6}{c}{$\epsilon = 0.2$}\\
		\midrule
		Random & $52.69 \pm 1.73$ & $59.53 \pm 0.61$ & $63.63 \pm 0.97$ & $66.22 \pm 0.87$ & $68.18 \pm 0.66$ \\
        BALD  & $48.23 \pm 1.63$ & $57.95 \pm 1.73$ & $64.14 \pm 1.49$ & $68.45 \pm 1.29$ & $70.68 \pm 0.84$ \\
        Coreset & $52.83 \pm 1.52$ & $59.44 \pm 1.22$ & $64.03 \pm 0.94$ & $66.18 \pm 1.25$ & $68.80 \pm 1.17$ \\
        Entropy & $49.44 \pm 1.54$ & $59.04 \pm 1.83$ & $65.52 \pm 1.17$ & $68.74 \pm 1.14$ & $71.79 \pm 1.45$ \\
        VAAL & $52.37 \pm 1.14$ & $59.98 \pm 0.82$ & $64.11 \pm 0.93$ & $66.55 \pm 0.72$ & $68.63 \pm 0.86$ \\
        \textbf{DeAn} & $\bm{52.62 \pm 1.43}$ & $\bm{61.93 \pm 1.33}$ & $\bm{67.45 \pm 0.94}$ & $\bm{71.08 \pm 0.59}$ & $\bm{73.52 \pm 0.71}$ \\
        \textbf{DeAn}\newline+\textbf{denoise} & $\bm{54.33 \pm 1.33}$ & $\bm{63.16 \pm 2.44}$ & $\bm{70.32 \pm 2.01}$ & $\bm{73.63 \pm 2.08}$ & $\bm{76.00 \pm 1.11}$ \\
		\multicolumn{6}{c}{$\epsilon = 0.3$}\\
		\midrule
		Random & $47.27 \pm 1.93$ & $54.06 \pm 1.79$ & $58.09 \pm 0.90$ & $61.06 \pm 0.95$ & $62.85 \pm 0.75$ \\
        BALD  & $45.48 \pm 1.24$ & $53.70 \pm 1.34$ & $59.68 \pm 1.11$ & $63.22 \pm 0.96$ & $66.05 \pm 0.69$ \\
        Coreset & $48.06 \pm 1.16$ & $54.10 \pm 1.59$ & $57.62 \pm 1.95$ & $60.81 \pm 1.60$ & $62.85 \pm 0.97$ \\
        Entropy & $45.34 \pm 1.83$ & $53.64 \pm 2.45$ & $59.20 \pm 2.15$ & $63.82 \pm 1.44$ & $66.86 \pm 1.13$ \\
        VAAL & $46.86 \pm 2.08$ & $53.14 \pm 1.53$ & $57.66 \pm 0.65$ & $60.40 \pm 0.83$ & $62.63 \pm 0.75$ \\
        \textbf{DeAn} & $\bm{48.33 \pm 1.57}$ & $\bm{55.85 \pm 1.55}$ & $\bm{61.15 \pm 1.30}$ & $\bm{64.43 \pm 0.90}$ & $\bm{66.93 \pm 0.92}$ \\
        \textbf{DeAn}\newline+\textbf{denoise} & $\bm{51.27 \pm 2.70}$ & $\bm{58.98 \pm 1.82}$ & $\bm{64.99 \pm 3.63}$ & $\bm{69.71 \pm 2.30}$ & $\bm{71.01 \pm 2.96}$ \\
		\multicolumn{6}{c}{$\epsilon = 0.4$}\\
		\midrule
		Random & $41.90 \pm 2.38$ & $47.68 \pm 2.01$ & $51.71 \pm 1.45$ & $54.01 \pm 1.12$ & $56.36 \pm 0.90$ \\
        BALD  & $41.16 \pm 2.38$ & $48.32 \pm 2.56$ & $53.49 \pm 2.23$ & $57.37 \pm 1.45$ & $59.93 \pm 1.71$ \\
        Coreset & $43.71 \pm 2.10$ & $48.50 \pm 1.52$ & $52.60 \pm 1.45$ & $54.62 \pm 1.45$ & $56.52 \pm 1.63$ \\
        Entropy & $42.65 \pm 1.56$ & $50.12 \pm 2.43$ & $55.64 \pm 2.14$ & $58.69 \pm 1.27$ & $62.32 \pm 1.46$ \\
        VAAL & $42.62 \pm 1.48$ & $48.58 \pm 1.23$ & $52.36 \pm 0.67$ & $54.45 \pm 0.87$ & $56.80 \pm 1.00$ \\
        \textbf{DeAn} & $\bm{43.55 \pm 1.65}$ & $\bm{49.95 \pm 1.34}$ & $\bm{53.39 \pm 1.58}$ & $\bm{56.61 \pm 0.98}$ & $\bm{58.01 \pm 1.27}$ \\
        \textbf{DeAn}\newline+\textbf{denoise} & $\bm{46.14 \pm 3.61}$ & $\bm{55.16 \pm 4.00}$ & $\bm{58.68 \pm 2.98}$ & $\bm{62.07 \pm 4.08}$ & $\bm{64.50 \pm 7.20}$ \\
 		\bottomrule
	\end{tabular}
\end{table*}

\begin{table*}[t!]
    \caption{Active learning results for EMNIST dataset (b=100).}
	\label{tab:results_EMNIST_b_100}
	\centering
	\begin{tabular}{llllll}
		\toprule
		\multirow{2}*{Algorithm} & \multicolumn{5}{c}{Number of acquired samples}\\
		\cmidrule{2-6}
		& 200 & 400 & 600 & 800 & 1000 \\
		\midrule
		\multicolumn{6}{c}{noise free}\\
		\midrule
		Random & $43.13 \pm 1.37$ & $50.75 \pm 0.86$ & $55.72 \pm 1.07$ & $59.17 \pm 1.11$ & $62.63 \pm 1.29$ \\
        BALD  & $41.18 \pm 1.48$ & $48.52 \pm 1.34$ & $53.66 \pm 1.73$ & $57.91 \pm 1.18$ & $61.08 \pm 0.79$ \\
        Coreset & $43.98 \pm 0.41$ & $50.73 \pm 0.71$ & $54.98 \pm 0.78$ & $59.04 \pm 0.70$ & $61.45 \pm 0.61$ \\
        Entropy & $38.92 \pm 1.83$ & $46.71 \pm 1.00$ & $51.91 \pm 1.64$ & $56.58 \pm 1.17$ & $60.14 \pm 1.03$ \\
        VAAL & $43.20 \pm 2.32$ & $50.98 \pm 1.18$ & $55.26 \pm 1.28$ & $59.39 \pm 0.86$ & $61.41 \pm 0.62$ \\
        BatchBALD & $40.51 \pm 1.15$ & $44.40 \pm 1.20$ & $46.81 \pm 1.72$ & $50.05 \pm 1.53$ & $51.93 \pm 1.01$ \\
        \textbf{DeAn} & $\bm{43.85 \pm 1.80}$ & $\bm{51.76 \pm 1.06}$ & $\bm{57.28 \pm 1.29}$ & $\bm{61.98 \pm 1.09}$ & $\bm{64.80 \pm 0.97}$ \\
		\multicolumn{6}{c}{$\epsilon = 0.1$}\\
		\midrule
		Random & $41.46 \pm 1.64$ & $47.14 \pm 1.81$ & $51.60 \pm 1.86$ & $55.44 \pm 1.17$ & $57.87 \pm 1.45$ \\
        BALD  & $39.30 \pm 1.32$ & $45.73 \pm 1.56$ & $49.69 \pm 1.75$ & $54.13 \pm 1.19$ & $56.79 \pm 1.29$ \\
        Coreset & $42.18 \pm 1.22$ & $47.53 \pm 1.09$ & $51.66 \pm 1.68$ & $54.68 \pm 1.14$ & $57.06 \pm 1.38$ \\
        Entropy & $38.64 \pm 1.61$ & $44.51 \pm 1.59$ & $48.95 \pm 1.72$ & $53.11 \pm 1.68$ & $57.01 \pm 1.70$ \\
        VAAL & $41.00 \pm 1.49$ & $47.30 \pm 1.61$ & $51.74 \pm 1.51$ & $55.07 \pm 1.24$ & $58.08 \pm 1.23$ \\
        BatchBALD & $37.89 \pm 0.94$ & $41.57 \pm 1.24$ & $43.83 \pm 1.67$ & $46.58 \pm 1.16$ & $48.18 \pm 0.93$ \\
        \textbf{DeAn} & $\bm{41.42 \pm 2.00}$ & $\bm{48.10 \pm 1.31}$ & $\bm{52.97 \pm 1.40}$ & $\bm{57.55 \pm 1.18}$ & $\bm{60.66 \pm 1.07}$ \\
		\multicolumn{6}{c}{$\epsilon = 0.2$}\\
		\midrule
		Random & $38.93 \pm 1.62$ & $44.32 \pm 1.61$ & $48.20 \pm 1.50$ & $51.46 \pm 1.21$ & $53.60 \pm 1.31$ \\
        BALD  & $37.16 \pm 1.72$ & $42.11 \pm 1.56$ & $45.99 \pm 1.44$ & $49.77 \pm 1.15$ & $52.90 \pm 1.28$ \\
        Coreset & $39.89 \pm 1.36$ & $44.95 \pm 1.48$ & $48.17 \pm 1.38$ & $51.23 \pm 1.47$ & $53.26 \pm 1.30$ \\
        Entropy & $36.71 \pm 1.23$ & $41.69 \pm 1.87$ & $44.84 \pm 2.02$ & $49.23 \pm 1.90$ & $51.71 \pm 2.02$ \\
        VAAL & $39.35 \pm 1.75$ & $43.82 \pm 1.66$ & $47.64 \pm 1.35$ & $50.99 \pm 1.54$ & $52.93 \pm 1.36$ \\
        BatchBALD & $38.12 \pm 1.63$ & $40.58 \pm 1.56$ & $41.99 \pm 1.92$ & $44.96 \pm 2.24$ & $46.05 \pm 1.59$ \\
        \textbf{DeAn} & $\bm{38.89 \pm 1.36}$ & $\bm{44.13 \pm 1.88}$ & $\bm{47.60 \pm 1.86}$ & $\bm{50.88 \pm 1.48}$ & $\bm{53.68 \pm 1.47}$ \\
		\multicolumn{6}{c}{$\epsilon = 0.3$}\\
		\midrule
		Random & $36.53 \pm 1.26$ & $40.91 \pm 1.62$ & $43.98 \pm 1.38$ & $46.88 \pm 1.31$ & $49.33 \pm 0.63$ \\
        BALD  & $36.13 \pm 1.75$ & $39.47 \pm 1.75$ & $42.84 \pm 1.82$ & $45.56 \pm 1.79$ & $48.16 \pm 1.50$ \\
        Coreset & $37.04 \pm 1.46$ & $40.88 \pm 1.70$ & $44.18 \pm 1.55$ & $47.23 \pm 1.52$ & $48.74 \pm 1.30$ \\
        Entropy & $34.84 \pm 1.58$ & $38.58 \pm 1.81$ & $41.72 \pm 2.21$ & $44.67 \pm 1.93$ & $47.59 \pm 2.18$ \\
        VAAL & $36.62 \pm 1.61$ & $41.11 \pm 1.72$ & $44.51 \pm 1.48$ & $47.10 \pm 1.67$ & $49.45 \pm 1.57$ \\
        BatchBALD & $34.55 \pm 0.76$ & $38.20 \pm 1.10$ & $39.55 \pm 0.80$ & $41.50 \pm 1.13$ & $42.85 \pm 0.94$ \\
        \textbf{DeAn} & $\bm{37.05 \pm 1.40}$ & $\bm{41.55 \pm 1.65}$ & $\bm{44.85 \pm 1.84}$ & $\bm{47.90 \pm 1.69}$ & $\bm{50.43 \pm 1.74}$ \\
		\multicolumn{6}{c}{$\epsilon = 0.4$}\\
		\midrule
		Random & $33.32 \pm 2.21$ & $36.49 \pm 2.17$ & $38.97 \pm 1.65$ & $42.51 \pm 1.79$ & $45.19 \pm 1.80$ \\
        BALD  & $33.93 \pm 1.32$ & $36.50 \pm 1.38$ & $38.69 \pm 1.45$ & $41.76 \pm 1.41$ & $43.87 \pm 1.37$ \\
        Coreset & $35.77 \pm 1.56$ & $38.07 \pm 2.05$ & $40.55 \pm 2.18$ & $42.64 \pm 1.34$ & $44.40 \pm 1.69$ \\
        Entropy & $35.17 \pm 1.66$ & $37.49 \pm 1.38$ & $39.66 \pm 1.88$ & $42.42 \pm 1.70$ & $44.74 \pm 1.99$ \\
        VAAL & $34.36 \pm 2.03$ & $37.05 \pm 1.79$ & $40.04 \pm 1.91$ & $42.48 \pm 2.23$ & $45.09 \pm 1.43$ \\
        BatchBALD & $32.18 \pm 1.42$ & $33.88 \pm 1.32$ & $35.55 \pm 1.73$ & $37.77 \pm 2.08$ & $38.87 \pm 2.55$ \\
        \textbf{DeAn} & $\bm{34.38 \pm 1.38}$ & $\bm{37.66 \pm 1.71}$ & $\bm{40.34 \pm 1.93}$ & $\bm{43.45 \pm 1.58}$ & $\bm{46.09 \pm 1.59}$ \\
 		\bottomrule
	\end{tabular}
\end{table*}

\begin{table*}[t!]
\caption{Active learning results for CIFAR100 dataset.}
	\label{tab:results_CIFAR100}
	\centering
	\begin{tabular}{lllll}
		\toprule
		\multirow{2}*{Algorithm} & \multicolumn{4}{c}{Number of acquired samples}\\
		\cmidrule{2-5}
		& 8000 & 16000 & 24000 & 32000 \\
		\midrule
		\multicolumn{5}{c}{noise free}\\
		\midrule
		Random & $21.40 \pm 0.79$ & $29.02 \pm 1.04$ & $34.14 \pm 1.00$ & $38.46 \pm 1.00$ \\
		BALD  & $16.83 \pm 1.35$ & $23.46 \pm 0.60$ & $27.26 \pm 0.93$ & $32.69 \pm 0.84$ \\
		Coreset & $19.85 \pm 0.86$ & $27.61 \pm 0.50$ & $33.32 \pm 0.78$ & $38.13 \pm 0.79$\\
		Entropy & $16.73 \pm 0.74$ & $23.03 \pm 0.65$ & $26.79 \pm 1.08$ & $33.83 \pm 1.34$\\
		VAAL & $21.17 \pm 0.94$ & $26.93 \pm 0.71$ & $31.31 \pm 0.59$ & $36.78 \pm 0.69$ \\
		\textbf{DeAn} & $\bm{21.68 \pm 0.86}$ & $\bm{28.82 \pm 0.87}$ & $\bm{34.16 \pm 0.88}$ & $\bm{38.27 \pm 0.69}$\\
		\multicolumn{5}{c}{$\epsilon = 0.1$}\\
		\midrule
		Random & $19.18 \pm 0.68$ & $25.54 \pm 0.96$ & $30.16 \pm 1.03$ & $33.28 \pm 0.96$ \\
		BALD  & $15.80 \pm 0.45$ & $19.61 \pm 0.78$ & $24.75 \pm 0.51$ & $29.04 \pm 1.41$ \\
		Coreset & $17.04 \pm 1.02$ & $24.14 \pm 0.90$ & $28.96 \pm 0.83$ & $33.30 \pm 0.58$\\
		Entropy & $15.44 \pm 0.68$ & $18.84 \pm 0.56$ & $24.43 \pm 0.94$ & $28.31 \pm 0.39$\\
		VAAL & $18.45 \pm 0.54$ & $24.89 \pm 0.69$ & $28.89 \pm 1.32$ & $31.60 \pm 0.77$ \\
		\textbf{DeAn} & $\bm{18.67 \pm 0.88}$ & $\bm{25.41 \pm 0.72}$ & $\bm{29.70 \pm 0.62}$ & $\bm{33.84 \pm 1.17}$\\
		\textbf{DeAn}+\textbf{denoise} & $\bm{19.30 \pm 0.79}$ & $\bm{26.17 \pm 1.02}$ & $\bm{30.84 \pm 1.03}$ & $\bm{34.51 \pm 1.11}$\\
		\multicolumn{5}{c}{$\epsilon = 0.3$}\\
		\midrule
		Random & $14.50 \pm 0.52$ & $18.99 \pm 0.77$ & $22.84 \pm 0.92$ & $25.94 \pm 0.99$ \\
		BALD  & $10.41 \pm 0.89$ & $15.93 \pm 0.46$ & $18.96 \pm 0.75$ & $23.15 \pm 1.15$ \\
		Coreset & $13.34 \pm 1.01$ & $18.17 \pm 0.75$ & $22.26 \pm 1.03$ & $25.32 \pm 0.73$\\
		Entropy & $11.32 \pm 0.72$ & $15.54 \pm 0.40$ & $18.72 \pm 0.97$ & $22.02 \pm 0.85$\\
		VAAL & $14.00 \pm 0.69$ & $17.74 \pm 0.67$ & $21.64 \pm 0.71$ & $24.86 \pm 0.86$ \\
		\textbf{DeAn} & $\bm{14.81 \pm 0.48}$ & $\bm{19.19 \pm 0.90}$ & $\bm{22.78 \pm 1.09}$ & $\bm{26.01 \pm 0.85}$\\
		\textbf{DeAn}+\textbf{denoise} & $\bm{15.35 \pm 0.51}$ & $\bm{21.60 \pm 0.80}$ & $\bm{25.56 \pm 0.76}$ & $\bm{28.20 \pm 1.00}$\\
        \bottomrule
	\end{tabular}
\end{table*}

\end{document}